\pdfoutput=1 
\documentclass[review]{elsarticle}
\usepackage{times}
\usepackage{amsmath}
\usepackage{graphicx}

\usepackage{geometry}
\usepackage[numbers]{natbib}

\usepackage{subfig}
\usepackage{algorithm}
\usepackage{algorithmic}

\usepackage{wrapfig}

\newcommand{\changelog}[1]{}

\newcommand{\norm}[1]{\left\lVert#1\right\rVert}
\newcommand{\isVec}[1]{\boldsymbol{#1}}
\newcommand{\isMat}[1]{\mathbf{#1}}
\newcommand{\refFig}[1]{Fig.~\ref{#1}}
\newcommand{\refEqu}[1]{(\ref{#1})}

\newcommand{\ignoreText}[1]{}

\newcommand{\red}[1]{#1}
\newcommand{\blue}[1]{#1}

\begin{document}

\begin{frontmatter}

%\title{Sparsity Cognizant loop closure for appearance based SLAM // Loop Closure detection as a convex optimization problem}
%\title{WORKING TITLE : A Generalized Framework for Closing Loops via Sparse and Redundant Representation}
%\title{An Online Sparsity-Cognizant Loop-Closure Algorithm for Visual Navigation}
\title{Sparse Optimization for Robust and Efficient Loop Closing}
%\title{Loop closing via convex optimization}
%\title{Online loop closing via sparse optimization}
% You will get a Paper-ID when submitting a pdf file to the conference system
%\author{Author Names Omitted for Anonymous Review. Paper-ID [112]}

%\author{
%\authorblockN{Yasir Latif\authorrefmark{1},
%Guoquan Huang\authorrefmark{2},
%John Leonard\authorrefmark{2}, 
%and Jos\'{e} Neira\authorrefmark{1}}
%\authorblockA{\authorrefmark{1}Instituto de Investigaci\'{o}n en Ingenier\'{i}a de Arag\'{o}n (I3A)\\
%Universidad de Zaragoza, Zaragoza, Spain \\
%Email: \{ylatif, jneira\}@unizar.es}
%\authorblockA{\authorrefmark{2}Computer Science and Artificial Intelligence Laboratory\\
%Massachusetts Institute of Technology, Cambridge, MA 02139, USA\\
%Email: \{gqhuang, jleonard\}@mit.edu}
%}

\author[UoA]{Yasir Latif}
%\address{ARC Center for Robotic Vision, University of Adelaide, Adelaide SA, Australia}
%
\author[UoD]{Guoquan Huang}
\author[MIT]{John Leonard}
%\address{Computer Science and Artificial Intelligence Laboratory Massachusetts Institute of Technology, Cambridge, MA 02139, USA}
%
\author[Unizar]{Jos\'{e} Neira}
%\address{Instituto de Investigaci\'{o}n en Ingenier\'{i}a de Arag\'{o}n (I3A) Universidad de Zaragoza, Zaragoza, Spain}

\address[UoA]{ARC Center for Robotic Vision, University of Adelaide, Adelaide SA 5005, Australia\footnote{Corresponding author, email: \{{\tt yasir.latif@adelaide.edu.au}\}}}
\address[UoD]{Department of Mechanical Engineering, University of Delaware, Newark, DE 19716, USA}
\address[MIT]{Computer Science and Artificial Intelligence Laboratory Massachusetts Institute of Technology, Cambridge, MA 02139, USA}
\address[Unizar]{Instituto de Investigaci\'{o}n en Ingenier\'{i}a de Arag\'{o}n (I3A) Universidad de Zaragoza, Zaragoza, Spain}
%\author{\authorblockN{Michael Shell}
%\authorblockA{School of Electrical and\\Computer Engineering\\
%Georgia Institute of Technology\\
%Atlanta, Georgia 30332--0250\\
%Email: mshell@ece.gatech.edu}
%\and
%\authorblockA{Twentieth Century Fox\\
%\and
%\authorblockN{James Kirk\\ and Montgomery Scott}
%\authorblockA{Starfleet Academy\\
%San Francisco, California 96678-2391\\

% avoiding spaces at the end of the author lines is not a problem with
% conference papers because we don't use \thanks or \IEEEmembership

%Homer Simpson\authorrefmark{2},
%Montgomery Scott\authorrefmark{3} and
%Eldon Tyrell\authorrefmark{4}}
%Telephone: (800) 555--1212, Fax: (888) 555--1212}
%\authorblockA{\authorrefmark{4}Tyrell Inc., 123 Replicant Street, Los Angeles, California 90210--4321}}

%\maketitle

\begin{abstract}

It is essential for a robot to be able to detect revisits or {\em loop closures} for long-term visual navigation.  
A key insight \red{explored in this work} is that the loop-closing event inherently occurs {sparsely}, i.e.,
the image currently being taken matches with only a small subset (if any) of previous images. 
Based on this observation, we formulate the problem of loop-closure detection as a 
{\em sparse, convex} $\ell_1$-minimization problem.
By leveraging fast convex optimization techniques, 
we are able to efficiently find loop closures, thus enabling real-time robot navigation.
This novel formulation requires no offline dictionary learning, as required
by most existing approaches, and thus allows {\em online incremental} operation.  
Our approach ensures a {\em unique} hypothesis
by choosing only a single globally optimal match when making a loop-closure decision.  
Furthermore, the proposed formulation enjoys a {\em flexible} representation
with {\em no} restriction imposed on how images should be represented, while
requiring only that the representations are 
\red{``close''} 
to each other when the corresponding images are visually similar.
The proposed algorithm is validated extensively using real-world datasets. 

\end{abstract}

\end{frontmatter}
%\IEEEpeerreviewmaketitle

\changelog{
\section*{CHANGE LOG} 
\begin{itemize}
\item Changed Fig. \ref{fig:KITTI} to plot Precision and Recall instead of Percentage TP and FP.
\item Added Fig. \ref{fig:KITTI-Multi} and the corresponding Section \ref{sec:DeepMultiExpr} showing experiments for ``Stacked Descriptors''.
\item Updated Table. \ref{tab:timing} to show timing for the C++ implementation instead of Matlab and updated the corresponding paragraph. The C++ timing are one-third of the previous timings.
\item Added some references to the original RSS paper and another paper which solved the optimization in an incremental way.

\end{itemize}

}

%%%%%%%%%%%%%%%%%%%%%%%%%%%%%%%%%%%%%%
% BEGIN \input{intro_relwork.tex}
%!TEX root = ras_2015.tex
\section{Introduction} \label{sec:intro}

With a growing demand for autonomous robots in a range of applications, 
such as search and rescue~\cite{Sugiyama2005ICCC,Capezio2009JCIT}, and space and underwater exploration~\cite{Cannell2005OCEANS}, 
it is essential for the robots to be able to navigate accurately for an extended period of time in order to accomplish the assigned tasks.
To this end, the ability to detect revisits (i.e., {\em loop closure} or place recognition) becomes necessary, 
since it allows the robots to bound the errors and uncertainty in the estimates of their positions and orientations (poses).
In this work, we particularly focus on loop closure during visual navigation, 
i.e., given a camera stream we aim to efficiently determine whether the robot has previously seen the current place or not.

Even though the problem of loop closure has been extensively studied in the visual-SLAM literature 
(e.g., see~\cite{lowry2016visual,CumminsNewmanIJRR08,GalvezTRO12}),
a vast majority of existing algorithms typically require the {\em offline} training of visual words (dictionary) 
from {\em a priori} images that are acquired previously in visually similar environments.
Clearly, this is not always the case when a robot operates in an unknown, drastically different environment.
%
%R%Moreover, these approaches often use high-dimensional descriptors, 
%R%such as SIFT~\cite{lowe2004distinctive} and BRIEF~\cite{calonder2010brief},
%R%which are extracted {\em sparsely} from the images in order to construct the visual Bag of Words (BoW).
%R%As a result, much valuable information available in the images gets lost.
%
%
In general, it is difficult to reliably find loops in (visual) appearance space.
One particular challenge is the perceptual aliasing -- that is, while images may be  
similar in appearance, they might be coming from different places. 
To mitigate this issue, both temporal 
(i.e., loops will only be considered closed if there are other loops closed nearby)
and geometric constraints 
(i.e., if a loop has to be considered closed,  a valid transformation must exist between the matched images)
can be employed~\cite{GalvezTRO12}.
\red{It is important to point out that 
the approach of~\cite{GalvezTRO12}} decides on the quality of a match {\em locally} --
If the match with the highest score (in some distance measure) is away from the second highest, it is considered a valid candidate. 
\red{
However, the local information may lead to incorrect loop-closure decisions
%even if both temporal and geometric consistency conditions hold,
%This is due to the fact that 
because both temporal and geometric conditions can easily fail in highly self-similar environments such as corridors in a hotel.}
%R%For example, all the corridors on all the floors of a hotel look alike,
%R%and a robot travelling on any one of them would be able to find ``loops'' 
%R%that are consistent over time and satisfy geometric transformation between the corresponding images,
%R%however, which can be catastrophic for robot navigation.

To address the aforementioned issues, 
in this paper we introduce a {\em general}, {\em online} loop-closure approach for vision-based robot navigation.
In particular, 
by realizing that loops typically occur intermittently in a navigation scenario,
we, for the first time ever, formulate loop-closure detection as a sparse $\ell_1$-minimization problem  that is convex. 
This is opposed to the current methods that cast loop closure detection as an image retrieval problem \cite{GalvezTRO12,CumminsNewmanIJRR08}. 
By leveraging the fast convex optimization techniques, we subsequently solve the problem efficiently
and achieve real-time frame-rate generation of loop-closure hypotheses.
Furthermore, the proposed formulation enjoys {\em flexible} representations
and can produce loop-closure hypotheses regardless of what the extracted features represent -- 
that is, any discriminative information, 
such as descriptors, Bag of Words (BoW), or even whole images,
%that can represent images, 
can be used for detecting loops.
%
%Furthermore, based on the photo-consistency assumption 
%(i.e., a same place observed by two cameras is assumed to have the similar intensity values in both images),
%we employ the whole (down-sampled) images, instead of sparse descriptors, to detect loops in the images.
%
%It is important to point out that our formulation is general, 
%and regardless of information loss, can also use the descriptors extracted from the images to detect loops. 
%
%{\color{blue}
Lastly, we shall stress that our proposed approach declares a loop that is valid only when it is {\em globally unique},
which ensures that if perceptual aliasing is being caused by more than one previous image, {\em no} loop closure will be declared. 
Although this is conservative in some cases,
since a false loop closing can be catastrophic while missing a loop closure generally is not, 
ensuring such global uniqueness is necessary and important, 
in particular, in highly self-similar environments.
%}

The rest of the paper is organized as follows: 
After reviewing the related work, %in the next section, 
we formulate loop-closure detection as a sparse  $\ell_1$-minimization problem in Section \ref{sec:SparseAndRedunant}.
In Section \ref{sec:DetectionLoops} we present in detail the application of this formulation  to visual navigation,
which is validated via the real-world experiments in Section \ref{sec:Expr}.
Finally, Section \ref{sec:conclusions}  concludes this work as well as outlines the possible directions for future research.

\section{Related Work} \label{sec:relwork}

The problem of loop-closure detection has been extensively studied in the SLAM literature 
and many different solutions have been proposed over the years (e.g., see~\cite{LatifIJRR,lowry2016visual} and references therein).
In what follows, we briefly overview the work that closely relates to the proposed approach.
%For instance, the recent work~\cite{LatifIJRR} introduces a consensus-based approach for robust loop closure over time,
%which is based on a series of Mahalanobis distance tests and thus is not appearance-based.
%
%Therefore, 
%In this work, we focus only on visual-appearance-based methods.
%

In particular, 
\red{
the FAB-MAP~\cite{CumminsNewmanIJRR08} is
a probabilistic appearance-based approach using visual BoW for place recognition,
and was shown to work robustly over trajectories up to $1000$~km.
%
%However, in~\cite{CumminsNewmanIJRR08} the dictionary consisting of visual BoW for the particular environment is learned beforehand,
%which is later used for detecting loop closures when the robot actually operates in that area. 
%%
Similarly, the Binary-BoW (BBoW)-based method~\cite{GalvezTRO12}
detects the FAST keypoints~\cite{rosten2005fusing} 
and employs a variation of the BRIEF  descriptors \cite{calonder2010brief} to construct the BoW. 
}
A verification step is further enforced to geometrically check the features extracted from the matched images. 
It should be pointed out that 
both methods~\cite{CumminsNewmanIJRR08,GalvezTRO12} are based on the similar ideas of text-retrieval \cite{hbow,Sivic03}:
These methods learn the BoW dictionaries beforehand,
which are used later for detecting loop closures when the robots actually operates in the field. 
This restricts the expressive power of the dictionary in cases where it has to operate in environments %not seen before.
drastically different from where the dictionary was constructed.
In contrast, the proposed approach builds the dictionary {\em online} 
as the robot explores an unknown environment,
while at the same time efficiently detecting loops (if any).
Moreover, rather than solely relying on the descriptors-based BoW,
our method is flexible and can utilize 
{\em all} pixel information to discriminate places even in presence of dynamic objects (encoded as sparse errors),
any descriptor that can represent similar places,
or any combination of such descriptors.

\red{
Some recent work has focused on loop closure under extreme changes in the environment
such as different weather and/or lighting conditions at different times of the day. 
For example, \citet{milford-daynight} proposed the SeqSLAM that is able to localize with drastic lighting and weather changes 
by matching sequences of images with each other as opposed to single images. 
%
%They showed that the system is able to correctly localize even with great changes in lighting and weather conditions.
%
\citet{churchill2013experience} introduced the experience-based maps
that learn the different appearances of the same place as it gradually changes in order to perform long-term localization.
Building upon \cite{churchill2013experience}, \citet{paul2013self} also discovered  new images to attain better localization.
%
%In effect, the findings of these works motivate us to investigate how much raw information is needed for closing loops
%even in the case without {\em a priori} experience (see Section~\ref{sec:Expr}).
In addition, \citet{lee2013place,lee2014place} have  explored geometric features such as lines for the task of loop closure detection in both indoor and outdoor scenarios.
Note that
if the information invariant to such changes can be extracted as in~\cite{CumminsNewmanIJRR08,GalvezTRO12,milford-daynight,churchill2013experience,paul2013self,lee2013place,lee2014place},
the proposed formulation can also be used to obtain loop-closure hypotheses.
Essentially, in this work we focus on finding loop closures given some discriminative descriptions such as descriptors and whole images,
assuming {\em no} specific type of image representations.
}

More recently, with the rediscovery of efficient machine learning techniques, 
Convolutional Neural Networks (CNNs)\cite{bengio2012deep,lecun1995convolutional} have been exploited to address loop closure detection \cite{sunderhaufIROS15,sunderhaufRSS15}. 
These networks are multi-layered architectures that are typically trained on millions of images for tasks such as object detection and scene classification. 
The internal representations at each layer are learned from the data itself and therefore can be used as features to replace hand-crafted features. %such as GIST and SURF.
Based on this approach, \citet{sunderhaufIROS15} extract features from different layers in the network and identify the layers that are useful for view-point 
and illumination invariant place recognition. 
Moreover, in \cite{sunderhaufRSS15} landmarks are treated as objects by finding object proposals in the images and features are extracted for them using deep networks. These features then allow for view-point invariant place categorization by matching different objects from varied viewpoints. 
In these CNN-based place categorization techniques, the networks are used as feature extractors followed by some form of matching. 
In this paper, we show that these deep features can also be utilized in the proposed framework of loop-closure detection.

It should be noted that in our previous conference publication \cite{LatifRSS14}, 
we have preliminarily shown that the proposed loop-closing framework is general and can employ most hand-crafted features. 
\red{
Recently, \citet{ShakeriFSR14} extended this sparse-optimization based framework to an incremental formulation allowing for the use of the previous solution
of the sparse optimization to jump start the next one,
while \citet{Zhang-RSS-16} further extended it to a multi-step delayed detection of loops (instead of single-step detection as in our prior work~\cite{LatifRSS14})
in order to exploit the  structured sparsity of the problem.}
In this paper, we present more detailed analysis and thorough performance evaluations, 
 including new experiments using deep features and validations in challenging multiple-revisit scenarios, 
 as well as new comparisons against the well-known nearest neighbour (NN) search.

%A previous version of this work appeared in \cite{LatifRSS14}. \citet{ShakeriFSR14} extended the sparse optimization framework presented in \cite{LatifRSS14} to an incremental solving approach, that allows to use the previous solution of the sparse optimization to jump start the next one. 

%\begin{itemize}
%\item Effect of image size
%\item Complexity with growing number of images (large dataset?)
%\item Time complexity?
%\item View point dependence ?
%\item Seasonal changes 
%\item long term loop closing
%\end{itemize}  
% END \input{intro_relwork.tex}
% BEGIN \input{problem.tex}
%!TEX root = ras_2015.tex
\section{Sparse Optimization for Loop Closure} \label{sec:SparseAndRedunant}

In this section, we  formulate loop-closure detection as a sparse optimization problem
based on a sparse and redundant representation.
Such representations have been widely used in computer vision for problems 
such as denoising~\cite{elad2010role}, deblurring~\cite{elad2006image}, 
and face recognition~\cite{cheng2010learning}. 
Similarly, \citet{Casafranca13icraWS,Casafranca13iros} formulated the back-end of graph SLAM as an $\ell_1$-minimization problem.
%To the best of our knowledge, 
However, {\em no} prior work has yet investigated this powerful technique for loop closure \red{detection} in robot navigation.
The key idea of this approach is to represent the problem {\em redundantly}, 
from which a {\em sparse} solution is sought for a given observation. 

%%The approach of the current method can be best described as: ``Given the current image, 
%%which image among the previously seen images is the best explanation for it?'' Before diving into the details of the method, 
%%it is necessary to review the mathematical foundations behind the proposed method. 

Suppose that we have the current image represented by a vector $\isVec{b} \in \mathcal R^n$, 
which can be either the vectorized full raw image or descriptors extracted from the image.
Assume that we also have a dictionary denoted by 
$\isMat{B} = \left[ \isVec b_1~\cdots ~\isVec b_m \right]\in \mathcal R^{n \times m}$,
which consists of $m$ basis vectors of the same type as $\isVec b$.
Thus, solving the linear system $\isMat{B}\isVec{x}=\isVec{b}$ yields
the representation of $\isVec{b}$ in the base $\isMat{B}$ in the form of the vector $\isVec{x} \in \mathcal R^{m}$. 
Elements of $\isVec{x}$ indicate which basis vectors, $\isVec b_i$, best explain $\isVec{b}$ 
and how much the corresponding contributions are.
\red{
A zero contribution ($x_i= 0$) simply implies that the corresponding basis vector $\isVec b_i$ is irrelevant to~$\isVec{b}$.}
One trivial example of a dictionary is the $n \times n$ identity matrix, $\isMat B = \isMat{I}_n$, 
which gives us the same representation, i.e., 
$\isMat{I}_n  \isVec{x}=\isVec{b} \Rightarrow \isVec{x}= \isVec{b}$. 
It is important to note that 
we have made {\em no} assumption about what the dictionary contains,
and in general, any arbitrary bases including random matrices or wavelets can be used as a dictionary. 
%as long as it fulfils some conditions. 

%One aspect of the problem is to ensure that a vector can be represented by a given dictionary. 
We know that a general vector $\isVec b$ can be represented by a basis matrix $\isMat B$ 
if and only if it belongs to the range space of $\isMat B$, i.e., 
$\exists \isVec x \neq \isVec 0, ~{\rm s.t.}~ \isMat{B}\isVec{x}=\isVec{b}$.
Carefully choosing these bases may result in a sparse representation, i.e., $\isVec x$ is sparse.
Very often this is the case in practice, because the signal is naturally sparse when represented in a certain, often overcomplete, basis \cite{elad2010role}. 
For instance, when representing an image using the wavelet basis, there are only a few coefficients that are nonzero. 
Note also that a vector may not be representable by the bases, for example,  
if the basis matrix (dictionary) is rank-deficient and the vector is in \blue{the right nullspace of the matrix}. 
To exclude such singular cases, 
in this work, we assume that the image vector $\isVec b$ is always representable by the basis matrix 
which is of full row rank (i.e., ${\rm rank} (\isMat B) = n$).
Moreover, we allow the bases to be redundant, 
that is, we may have more (not necessarily orthogonal) basis vectors than the dimension of the image vector (i.e., $m > n$).
Note that this assumption can be satisfied by carefully designing the dictionary (see Section~\ref{sec:DetectionLoops}).
In general, a redundant dictionary leads to the sparse representation of the current image, 
which is what we seek and describes the sparse nature of the loop-closure events (i.e., occurring sparsely).
%in contrast to a non-redundant (minimum) orthogonal basis matrix that results in a non-sparse (dense) representation.
%
%In what follows, we look at how sparse representation can be achieved using redundant dictionaries 
%and how it can be applied to the problem of loop closure detection in Visual SLAM.
%
%
%
%Consider the case when we have a set of vectors $\isVec{d_i} \in R^n $ where $i \in \{1,2,3 \hdots m\}$ 
%which form the the basis vectors or ``atoms'' of our dictionary. At the moment, we do not impose any 
%restriction on what these vectors represent. As before, we stack them in a matrix $\mathbf{D} \in R^{n \times m}$. 
%Each basis vector lies in $\mathbf{R}^n$ and we have $m$ such basis. As already mentioned, any vector in $R^n$ can 
%be represented by an a set of $n$ orthogonal vectors $\in R^n$ but this representation is non-sparse in general. 
%That is for a non-trivial signal, all the orthogonal basis have non-zero contribution towards reconstructing the given vector. 
%
%

Consider that we have $m > n$ basis vectors,
then $\isMat{B}\isVec{x}=\isVec{b}$ becomes an under-determined linear system and 
has infinitely many solutions. 
%(assuming $\isMat{D}$ has full row rank). 
%
Therefore, we have to regularize it 
in order to attain a unique solution by specifying a desired criterion that the solution should satisfy. 
Typically, this regularization takes the form of looking for a solution, $\isVec{x}^\star$, that leads to the minimum 
reconstruction error in the $\ell_2$-norm sense, which corresponds to the least-squares formulation:
\begin{align}
 \underset{\isVec{x}}{\text{min}} ~~ \norm{\isMat{B}\isVec{x}-\isVec{b}}_2^2  
~~\Rightarrow~~ \isVec{x}^\star = \isMat{B}^T(\isMat{B}\isMat{B}^T)^{-1}\isVec{b}
 \label{equ:l2}  
\end{align}
%%
%where $\norm{.}_2^2$ is the Euclidean or $l_2$-norm. The problem in (\ref{equ:l2}) has a closed form solution i.e
%\[
%\isVec{x}^\star = \isMat{D}^T(\isMat{D}\isMat{D}^T)^{-1}\isVec{b}
%\]
%%
%where $\isMat{D}\isMat{D}^T$ is invertible due to the full row rank assumption.
%
%
Note that $\ell_2$-norm is widely used in practice in part because of the closed-form unique solution, 
while leading to a {\em dense} representation, i.e., almost all of the elements of $\isVec{x}^\star$ are non-zero 
and thus all the basis vectors are involved in representing the current image vector. 
This goes against our prior knowledge of loop closures occurring sparsely. 
Theoretically, the least-squares solution can lie very far from the sparse solution for an underdetermined system \cite{donoho2006most}, 
which motivates the ensuing sparse formulation.

Due to the fact that loop-closure events often occur sparsely,
we instead aim to find the {\em sparsest} possible solution under the condition that it best explains the current image.
%
%Another regularization scheme is to look for a solution that can reconstruct the observed vector $\isVec{b}$ 
%using as few of the basis vectors as possible. 
%%This means that we want our solution $\isVec{x}$ to have as few non-zero entries as possible. 
%In other words, we are looking for the sparsest possible solution under the condition that it can explain our current observation $\isVec{b}$. 
%%
Intuitively, by assuming that the basis dictionary consists of all the previous observed images and thus is redundant, 
we are looking for the smallest possible subset of previous images that can best explain the current image. 
The smallest such subset would contain just a single image 
which is ``closest'' to the current image (in appearance or descriptor space) under the assumption that 
there exists a unique image from the past which matches the current image. 
%
%For this reason, this framework can be used to detect loop closures since redundant representation allows us 
%to consider each previous image as a basis vector in our dictionary and solving for the sparsest 
%solution gives us the subset of previous images that best explains that current image. 
%
%
To that end, we  employ the $\ell_0$-norm to quantify the sparsity of a vector,
i.e.,  $\norm{\isVec{x}}_0 = {\rm Card} (\isVec{x}: \forall i, x_i \neq 0 )$, 
the total number of non-zero elements in $\isVec{x}$.
Note that a vector with $d$ non-zero elements is called $d$-sparse.
Thus, the problem of loop closure  can be formulated as follows:
\begin{equation}
\begin{aligned}
& \underset{\isVec{x}}{\text{min}}
& & \norm{\isVec{x}}_0 
& \text{subject to}
& & \mathbf{\isMat{B}\isVec{x}}=\isVec{b}
\end{aligned}
\label{equ:l0}
\end{equation}
%
%where $\isVec{b}$ is the observation at the current point in time.
The above problem is a combinatorial optimization problem which in general is NP-hard~\cite{Amaldi1998237},
because all of the possible $d$-sparse vectors have to be enumerated to check whether they fulfill the constraint. 
To address this computational-intractability issue, we relax the $\ell_0$-norm in~\eqref{equ:l0} to $\ell_1$-norm,
which is defined as the summation of absolute values of the elements of $\isVec{x}$, 
$\norm{\isVec{x}}_1 = \sum_{i=1}^n |x_i|$,
\red{since it is known that $\ell_1$-norm 
is the closest convex approximation to $\ell_0$-norm
and also results in a sparse solution~\cite{Donoho2006TSP}, i.e.,}
%We can now state the relaxed version of the problem in (\ref{equ:l0}) as:
%
\begin{equation}
\begin{aligned}
& \underset{\isVec{x}}{\text{min}}
& & \norm{\isVec{x}}_1 
& \text{subject to}
& & \isMat{B}\isVec{x}=\isVec{b}
\end{aligned}
\label{equ:l1}
\end{equation}

The problem~\eqref{equ:l1} assumes the perfect reconstruction without noise, which clearly is not the case in practice.
Hence, we introduce a sparse noise (error) term along with the basis vectors to explain the current image 
\red{and reformulate \eqref{equ:l1} as follows:}
\begin{align}
& \underset{\isVec{x, e}}{\text{min}}~~ \norm{\isVec{x}}_1 + \norm{\isVec{e}}_1  ~~ \text{subject to}~~ \isMat{B}\isVec{x} + \isVec{e}=\isVec{b} \label{equ:l1-noisy} \\
\Rightarrow~ & \underset{\isVec{\alpha}}{\text{min}} ~~ \norm{\isVec{\alpha}}_1  ~~ \text{subject to} ~~ \isMat{D}\isVec{\alpha}=\isVec{b} \label{equ:l1-noisy-compact}
\end{align}
where 
$\isMat{D} :=\left[ \begin{array}{cc}	\isMat{I}_n & \isMat{B}	\end{array} \right]$ and 
$\isVec{\alpha} := 	\left[ \begin{array}{c} 	\isVec{e} \\ \isVec{x} \end{array}\right]$.
%
%(\ref{equ:l1-noisy}) can be rewritten as 
%
%\begin{equation}
%\begin{aligned}
%& \underset{\isVec{\alpha}}{\text{min}}
%& & \norm{\isVec{\alpha}}_1 
%& \text{subject to}
%& & \isMat{B}\isVec{\alpha}=\isVec{b}
%\end{aligned}
%\label{equ:l1-noisy-compact}
%\end{equation}
%\noindent
%which is the modified version of (\ref{equ:l1-noisy}) that caters for noise in the current 
%observation while still being an $l_1$-norm minimization problem similar to (\ref{equ:l1}).
%
%%%%%%%%%%%%%%
%%%%%%%%%%%%%%
%{\color{blue}
Note that we have normalized the basis vectors when building the dictionary $\isMat D$,
and thus $\isMat I_n$ can be considered as the noise bases along each of the $n$ directions of the data space,
and $\isVec e$ becomes an indication variable for which noise components dominate the reconstruction error.
This allows us to normalize $\isVec x$ and $\isVec e$ together when computing contributions of 
data and noise bases (see Section~\ref{sec:build-dict}).
%}
%
We stress that this new formulation of loop closure~\eqref{equ:l1-noisy-compact} 
takes advantage of the fact that $\ell_1$-norm automatically promotes sparsity,
\red{as opposed to the more commonly used $\ell_2$-based least-squares formulation~\cite{bach2011convex}. 
In solving \eqref{equ:l1-noisy-compact} for a minimum $\ell_1$-norm solution,
we are in effect seeking an explanation of the current 
image with the fewest basis vectors from the redundant dictionary.}
This problem is also known as atomic decomposition~\cite{elad2010role},
since $\isVec b$ is decomposed into its constituent  atoms in the dictionary. 

% END \input{problem.tex}
% BEGIN \input{closingLoops.tex}
%!TEX root = ras_2015.tex
\section{Closing Loops via $\ell_1$-Minimization} \label{sec:DetectionLoops}

\red{
We have formulated loop closure as a sparse convex $\ell_1$-minimization problem~\eqref{equ:l1-noisy-compact} in the preceding section.
We now   present in detail how this novel formulation can be utilized in vision-based robot navigation.
In what follows, we first explain how a dictionary can be constructed {\em incrementally online},
and then how such a dictionary can be used at {\em each time step} to generate loop-closure hypotheses. 
}

\subsection{Building Dictionary} \label{sec:build-dict}

The first step in the proposed process of detecting loops is to build a set of basis vectors that make up the dictionary. 
Unlike the state-of-the-art loop-closure detection methods (e.g.,~\cite{GalvezTRO12}), 
the proposed approach does {\em not} learn the dictionary {\em offline} before the start of the experiment. 
Instead, the dictionary 
%(a set of basis vectors which are not a ``dictionary'' in the sense of BoW) 
is built exclusively for the current experiment,
as the robot moves and collects images  -- that is, incrementally {\em online} as images arrive.

As a new image becomes available, a mapping function, 
$\isVec{f}: \mathcal R^{(r,c)} \to \mathcal R^n$, 
transforms the image of resolution $r \times c$ to a {unit} vector of dimension $n$.
Due to the flexibility of representation enjoyed by our proposed approach,
this function is general and can be either a whole image or descriptors 
such as HOG \cite{dalal2005histograms} and GIST \cite{oliva2001modeling} %, or local descriptors 
%such as SIFT \cite{lowe2004distinctive} and SURF \cite{bay2006surf} 
computed over the image. 
That is, the basis vectors can represent any information that helps distinguish between two images. 
The proposed method can be considered as data association in a high-dimensional space 
carried out by (approximately) reconstructing a given vector from a subset of unit vectors in that space. 
As such, this approach is agnostic to what these vectors physically represent. 
For this reason, versatility of representation is inherent to our algorithm, 
allowing representations ranging from whole images to descriptors, BoW, or even mean and variance normalized images over time 
for matching sequences across different times of day or changing weather conditions. 
%
%
%When the basis vectors are in some descriptor space, $\isMat{D}\isVec{\alpha}$ reconstructs a vector 
%in that space which is identical to the observed descriptor $\isVec{b}$ [see~\eqref{equ:l1-noisy-compact}]. 
%
Also, since we use the $\ell_2$-norm of the error 
\red{(i.e., Euclidean distance)} to measure how good the reconstruction is, 
any descriptor whose distance between two vectors is measured in term of $\ell_2$-norm,
can be naturally incorporated in the proposed approach.
% 
%Moreover, it is assumed that these basis can be reshaped to form a vector in a fixed dimensional space. 
%In all our experiments, unless otherwise stated, 
%we simply vectorize the image to obtain the vector of  $\mathcal R^{n = r \times c }$.
%%out into a vector by taking all the columns of the image and stacking them up in a long vector $\in \mathcal R^{n = r \times c }$. 

In order to ensure full row rank of the dictionary matrix~$\isMat D$,
we initialize the dictionary with an identity matrix $\isMat{I}_{n}$,
%%%
which also accounts for the bases of the noise $\isVec e$ [see~\eqref{equ:l1-noisy-compact}].
%%%
%
When the first image denoted by $\isVec{i}_1$ arrives, $\isVec b_1 = \isVec{f}(\isVec{i}_1)$ is added to the dictionary. 
In general, updating the dictionary at the $i$-th time step is simply appending $\isVec b_i =\isVec{f}(\isVec{i}_i)$ to 
the end of the current dictionary%
\footnote{
Although this simple augmentation would make the dictionary grow unbounded, 
more sophisticated update policies (e.g., replacing or merging basis vectors for the same locations) 
can be designed in order to control the size of the dictionary when a robot repeatedly operates in the same environment.
}.
However, before augmenting the dictionary, we need to determine whether or not there are some 
previous images that explain the current one, 
i.e., we need to detect any loop that can be closed based on the current image.

%{\color{blue}
\subsection{Solving $\ell_1$-Minimization} \label{sec:solve-ell1-min}
%}

Once the dictionary $\mathbf D$ is available and when an image arrives at every time step,
we are now ready to solve the convex, sparse $\ell_1$-minimization problem~\eqref{equ:l1-noisy-compact} in order to find (if any) loop closures.
\red{While various approaches such as the primal-dual method are available for solving a convex optimization problem~\cite{Boyd2004},
we here leverage the homotopy approach because of its efficiency~\cite{malioutov2005homotopy,donoho2006fast},
which is specifically designed to take advantage of the properties of $\ell_1$-minimization.}

In particular, relaxing the equality constraint in~\eqref{equ:l1-noisy-compact} yields  the following {\em constrained} minimization:
\begin{equation}
\begin{aligned}
& \underset{\isVec{\alpha}}{\text{min}}
& & \norm{\isVec{\alpha}}_1 
& \text{subject to}
& & \norm{\isMat{D}\isVec{\alpha} - \isVec{b} }_2 \leq \epsilon
\end{aligned}
\label{equ:l1-bpdn}
\end{equation}
where $\epsilon > 0 $ is a pre-determined noise level.
This  is termed the basis pursuit denoising  problem in compressive sensing~\cite{Donoho2006TSP}.
\red{
An equivalent variant} of~\eqref{equ:l1-bpdn} is the following {\em unconstrained} minimization problem that is actually solved by the homotopy approach: 
\begin{equation}
\begin{aligned}
& \underset{\boldsymbol\alpha}{\text{min}}
& & \lambda \norm{\isVec{\alpha}}_1 + \frac{1}{2}\norm{\isMat{D}\isVec{\alpha} - \isVec{b} }_2
\end{aligned}
\label{equ:l1-unconstrained}
\end{equation}
where $\lambda$ is a scalar weighting parameter. 
%This converts the original constrained problem into an unconstrained problem. 
%
To solve~\eqref{equ:l1-unconstrained}, the homotopy method uses the fact that the objective function undergoes 
a homotopy continuation 
\red{from the $\ell_2$ cost (the second term) to the $\ell_1$ cost (the first term) as $\lambda$ increases from zero to one~\cite{donoho2006fast}.} 
The computational complexity of this approach is $O(dn^2+dnm)$ for recovering a $d$-sparse signal in $d$ steps using an $n \times m$ dictionary,
although in the worst case when recovering a non-sparse solution in a high-dimensional observational 
space and large number of basis vectors, it can perform as worse as $O(m^3)$, 
which fortunately is not the case in this work.

The above homotopy solver is employed to determine loop closure for the $i$-th image represented by $\isVec{i_i}$,
by solving~\eqref{equ:l1-unconstrained} for $\isVec{ f (i_i)}$ using the up-to-date dictionary 
$\isMat D_{i-1} = \left[
		\begin{array}{rrrrr}
		 \isMat{I}_{n} &  \isVec{f (i_1)} & \hdots &\isVec{f (i_{i-1})}
		\end{array} 
\right]$.\footnote{The subscript $i-1$ is hereafter used to denote the time index,
thus revealing the online incremental process of the proposed approach.}
The solution $\isVec{\alpha_i} = \begin{bmatrix} \alpha_{i,1} & \cdots & \alpha_{i,n+i-1}\end{bmatrix}^T$ 
at the $i$-th time step contains the contribution of all previous bases in constructing 
the current image. To find a unique image to close a loop with, we are interested in which basis vector
has the greatest relative contribution, which can be found by calculating the unit vector 
${\isVec{\hat\alpha_i}} = \isVec{\alpha_i}/\norm{\isVec{\alpha_i}}_2$. 
Any entry greater than a predefined threshold, $\tau$, is considered a loop-closure candidate.
%(by noting $\sum_{j=1}^{n} {\hat\alpha_{i,j}}^2 = 1$).
%
%
In addition, due to the fact that in a visual navigation scenario, 
the neighbouring images are typically overlapped with the current image and thus have great ``spurious'' contributions,
we explicitly ignore a time window, $t_g$, around the current image, 
during which loop-closure decisions are not taken. 
This is a design parameter and can be chosen based on the camera frequency (fps) and robot motion.
Once the decision is made, the dictionary is updated by appending $\isVec{f (i_i)}$ to it, i.e., 
$\isMat D_{i} = \left[
		\begin{array}{rrrrr}
		 \isMat D_{i-1} & \isVec{f (i_{i})}
		\end{array} \right]$.
The main steps of this process are summarized in Algorithm~\ref{alg:l1-loops}.

%%%%%%%%
\begin{algorithm}[t!]
\caption{Closing Loops via $\ell_1$-Minimization}
\label{alg:l1-loops}

\small

\begin{algorithmic}[1]

\REQUIRE Dictionary $\isMat{D}_{i-1}$, Current image $\isVec{i}_i$, Threshold $\mathbf{\tau}$, Weight $\mathbf{\lambda}$, Ignoring-time window $t_g$

\ENSURE Loop-closure hypotheses $\mathbf{H}$, Updated dictionary $\isMat{D}_{i}$

\STATE $\isVec{b}_i := \isVec {f(i_i)}$ 

\STATE \textit{Hypothesis generation:}

\STATE ~Solve $\underset{\isVec{\alpha}_i}{\min}~~\lambda \norm{\isVec{\alpha}_i}_1 + \frac{1}{2}
\norm{\isMat{D}_{i-1}\isVec{\alpha}_i - \isVec{b}_i}_2$ using the homotopy approach (see Section~\ref{sec:solve-ell1-min})

\STATE ~~Normalize ${\isVec{\hat\alpha_i}} := \frac{\isVec{\alpha_i}}{\norm{\isVec{\alpha_i}}_2} $ 

\STATE ~~Find hypotheses $ \mathbf H := \{ j ~|~ {{\hat\alpha_{i,j}}} > \tau, ~\norm{i-j}_1 > t_{g} \}$

\STATE \textit{Dictionary update:}

\STATE ~~$\isMat{D}_{i} := \left[ 
			\begin{array}{cc}
			\isMat{D}_{i-1}& \isVec{b}_i
		\end{array} 
		\right]$

\end{algorithmic}

\end{algorithm}

%%%%%%%%%%%%

\begin{figure*}[t!]
        \centering
        \includegraphics[width=0.20\textwidth]{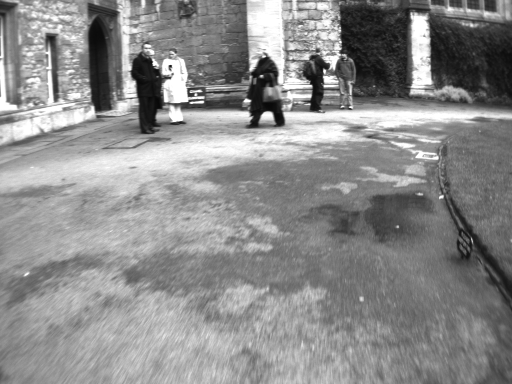}
        \includegraphics[width=0.20\textwidth]{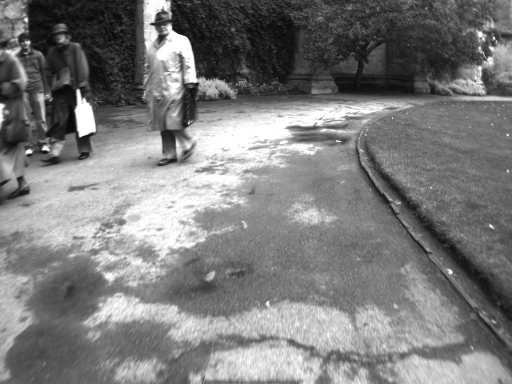}
        \includegraphics[width=0.20\textwidth]{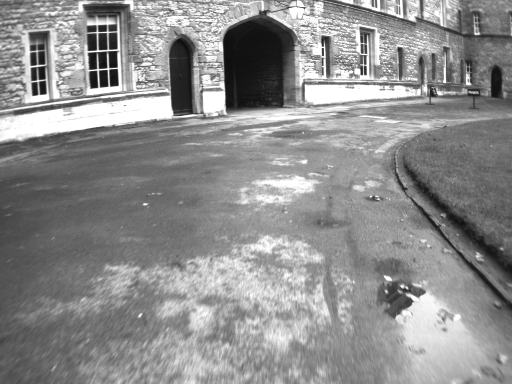}
        \includegraphics[width=0.20\textwidth]{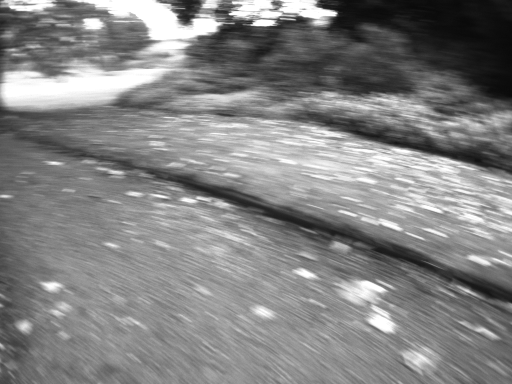}
		\\ \vspace{1.30mm}
		\includegraphics[width=0.20\textwidth]{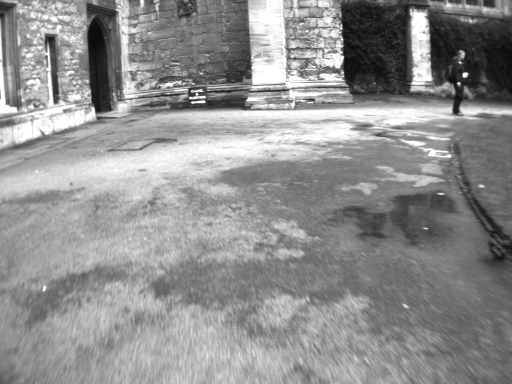}
		\includegraphics[width=0.20\textwidth]{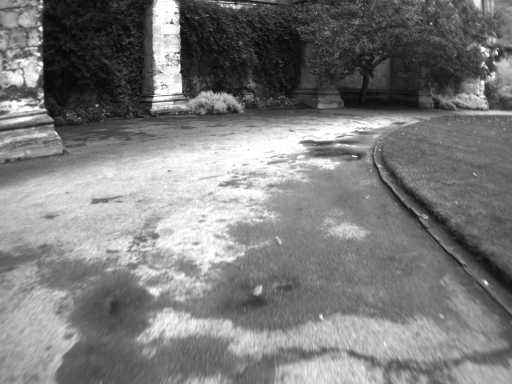}
		\includegraphics[width=0.20\textwidth]{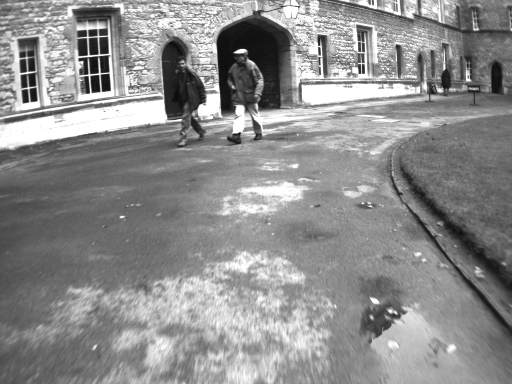}
		\includegraphics[width=0.20\textwidth]{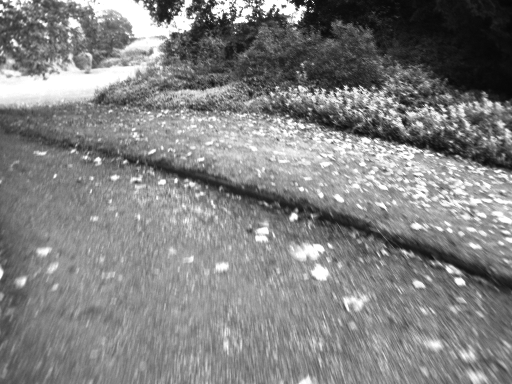}
        \caption{Sample images from the New College dataset: Query images (top) and the corresponding match images (bottom). 
        The images are down-sampled to $8 \times 6$ pixels. %($\tau = 0.99$). 
        Note that in spite of dynamic changes and motion blurs occurring in these images which deteriorate the loop-closure problem,
        the proposed approach still provides reliable results.}
        \label{fig:newCollege-pics}
        \vspace{-3mm}
\end{figure*}

%{\color{blue}

\subsection{Remarks}

\subsubsection{Global uniqueness}
\label{sec:uniqueness}

It is important to note that 
the solution to~\eqref{equ:l1-unconstrained}, by construction, is guaranteed to be sparse. 
\red{
In the ideal case of no perceptual aliasing, the solution is expected to be $1$-sparse, 
because {\em ideally} there  exists only one image
in the dictionary that  matches the current image when a revisit occurs.} 
In the case of exploration where there is no actual loop-closure match in the dictionary, 
the current image is best explained by the last observed and the solution hence is still $1$-sparse,
\red{which however will not generate a valid loop-closure detection because of the temporal constraint $t_g$ enforced in our implementation.} 
\blue{
%This is under the assumption that the current image is sufficiently similar to one  image in the dictionary. 
Note that if there are significant differences such as illumination or dynamic objects, the solution may no longer be $1$-sparse (see Section~\ref{sec:Norland}).
}

In a general case where $k>1$ images that have been previously observed and that are visually similar to the current image,
a naive thresholding based method --
which simply compares the current image to each of the previous ones based on some similarity measure --
would likely produce $k$ loop-closure hypotheses corresponding to the $k$ images in the dictionary.
It is very important to note that such an approach independently calculates the contribution of each previous image, 
{\em without} taking into account the effects of other images or data noise, 
%This can be considered as {\em decoupled} computation of contributions -- 
despite the fact that due to noise they may be correlated 
and thus is {\em suboptimal}.
%
%while an erroneous loop closure may be catastrophic for the navigation (estimation) algorithm.
% 
In contrast, the proposed  $\ell_1$-minimization-based approach simultaneously computes the optimal contribution
of all the previous images and noise by finding the global optimal solution $\hat\alpha_{i}$ of the convex problem~\eqref{equ:l1-unconstrained},
and guarantees the unique hypothesis by selecting the $j$-th image with the greatest $\hat\alpha_{i,j}$.
%

%{\color{blue}
In the case of multiple revisits to the same location, 
the proposed approach, as presented here, is {\em conservative} (i.e, only one, but the best one, revisit would be selected). 
Including the corresponding images from earlier visits in the dictionary would lead to a non-unique solution, when the same location is revisited again.
%
%, by closing only a single, instead of multiple, loops, 
%it at least ensures the globally best one to close.
%
%
However, the proposed method can be easily extended to detect loops on multiple revisits. 
Instead of considering the contribution of all the previous basis separately, 
if a loop exists between previous locations $k$ and $l$, 
we consider their joint contribution ($\hat{\alpha}_{i,k} + \hat{\alpha}_{i,l}$) when making the decision. 
This ensures that even though these places are not individually unique enough to explain the current image, 
together (and since they are visually similar as we already have a loop closure between them), 
they best explain the current observation, allowing us to detect loop closures in case of multiple revisits.
%}
%
%{\color{red}
%In the worst case where $k>1$ images {\em perfectly} match the current image due to perceptual aliasing, 
%we would have recovered a $k$-sparse solution (where $k << m$). 
%%
%However, this case can occur with almost zero probability in practice since it requires to have {\em identical} $k$ camera poses at different times;
%even if it happened, the proposed approach can avoid this scenario
%due to its built-in {globally unique} mechanism.
%%
%Specifically, in this case each one of the $k$ images has a contribution of $1/k$ in the normalized vector ${\isVec{\hat\alpha_i}}$. 
%For any $\tau > 0.5$ and $k>1$, it is trivial to see that $1/k < \tau$. 
%This implies that if more than one image exactly match the current image, 
%the maximum that it can contribute would still be less than any threshold greater than $0.5$. 
%In the conservative side, this results in {\em no} valid loop closure after thresholding.
%}%
%%%%%%%%????
%There can only be a single {\em unique} value ${\hat\alpha_{i,j}} >0.5$.  %\hat\alpha_{i,j}^2???
%Consider if the $j$-th element has a value $\beta$, then the sum of all the other elements 
%is equal to $1-\beta$. If $\beta$ is greater than $\tau$, then $(1-\beta) < \tau $ holds.

\subsubsection{Flexible basis representation}

We stress that the dictionary representation used by the proposed approach is {general} and {flexible}.
Although we have focused on the simplest basis representation using the down-sampled whole images (see Section~\ref{sec:Expr}),
this does not restrict our method only to work with this representation.
In fact, any discriminative feature that can be extracted from the image (e.g., GIST, HOG, etc.) 
can be used as dictionary bases for finding loops, 
thus permitting the desired properties such as view and illumination invariance. 
To show that, particular experiments have been performed in Section~\ref{sec:Expr},
using different types of bases and their combinations.
Moreover, it is not limited to a single representation at a time. 
If we have $k$ descriptors $\isVec{f}_i : \mathcal R^{(r,c)} \to \mathcal R^{k_i}$, 
a multi-modal descriptor can be easily formed by stacking them up in a vector in $\mathcal R^K$ ($K = \sum_{i=1}^{k}k_i$).
\red{This idea has recently been exploited in \cite{Zhang-RSS-16}.}
Therefore, our proposed method can be considered as a {\em generalized} loop-closing approach 
that can use any basis vectors as long as a metric exists to provide the distance between them.

\subsubsection{Robustness}

\begin{figure*}[t!]
\centering
\subfloat[Basis used as raw $64 \times 48$ images]
{\includegraphics[width=0.3\textwidth]{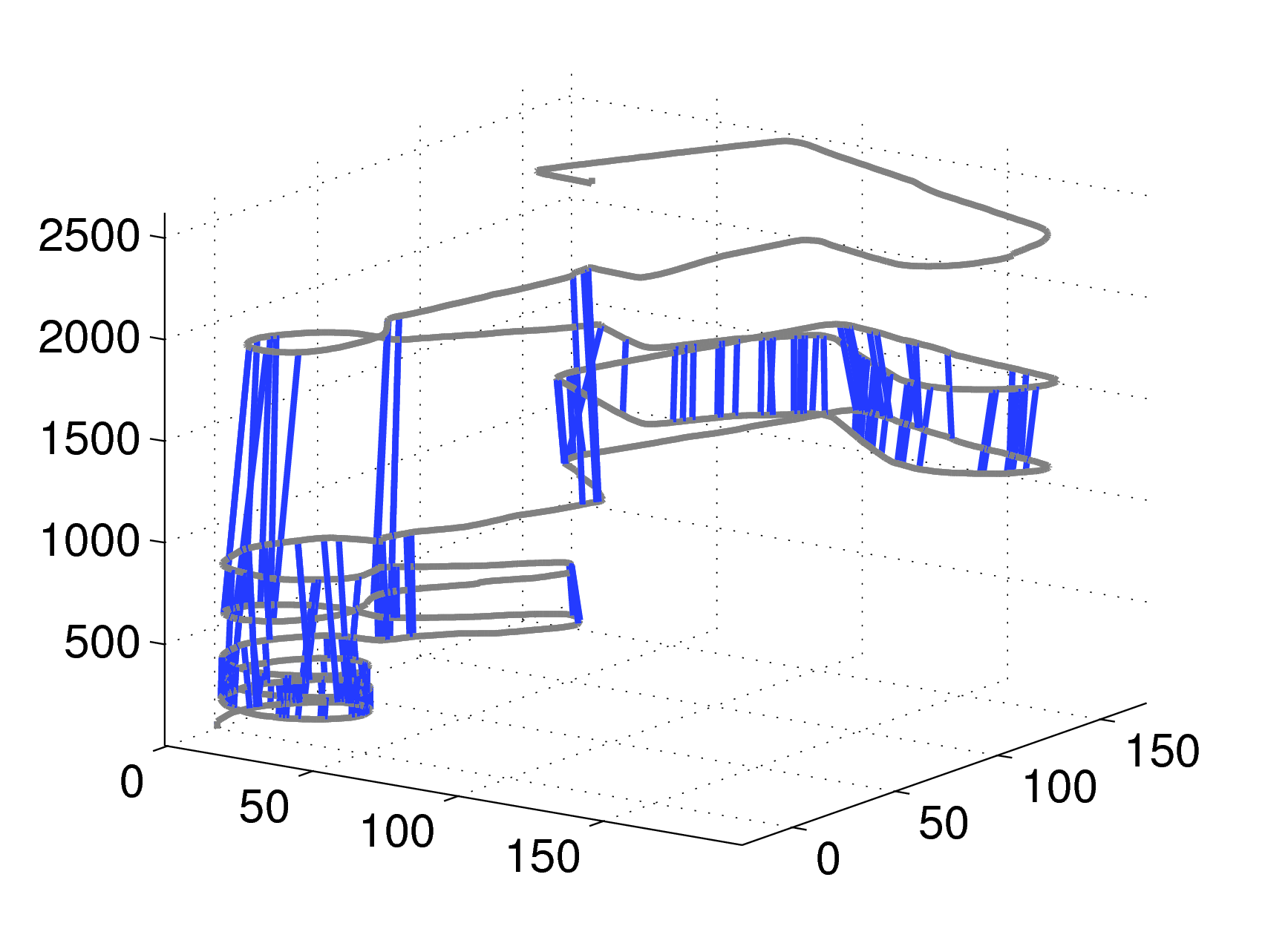} } ~~
\subfloat[Basis used as raw $8 \times 6$ images] %, $\tau=0.99$
{\includegraphics[width=0.3\textwidth]{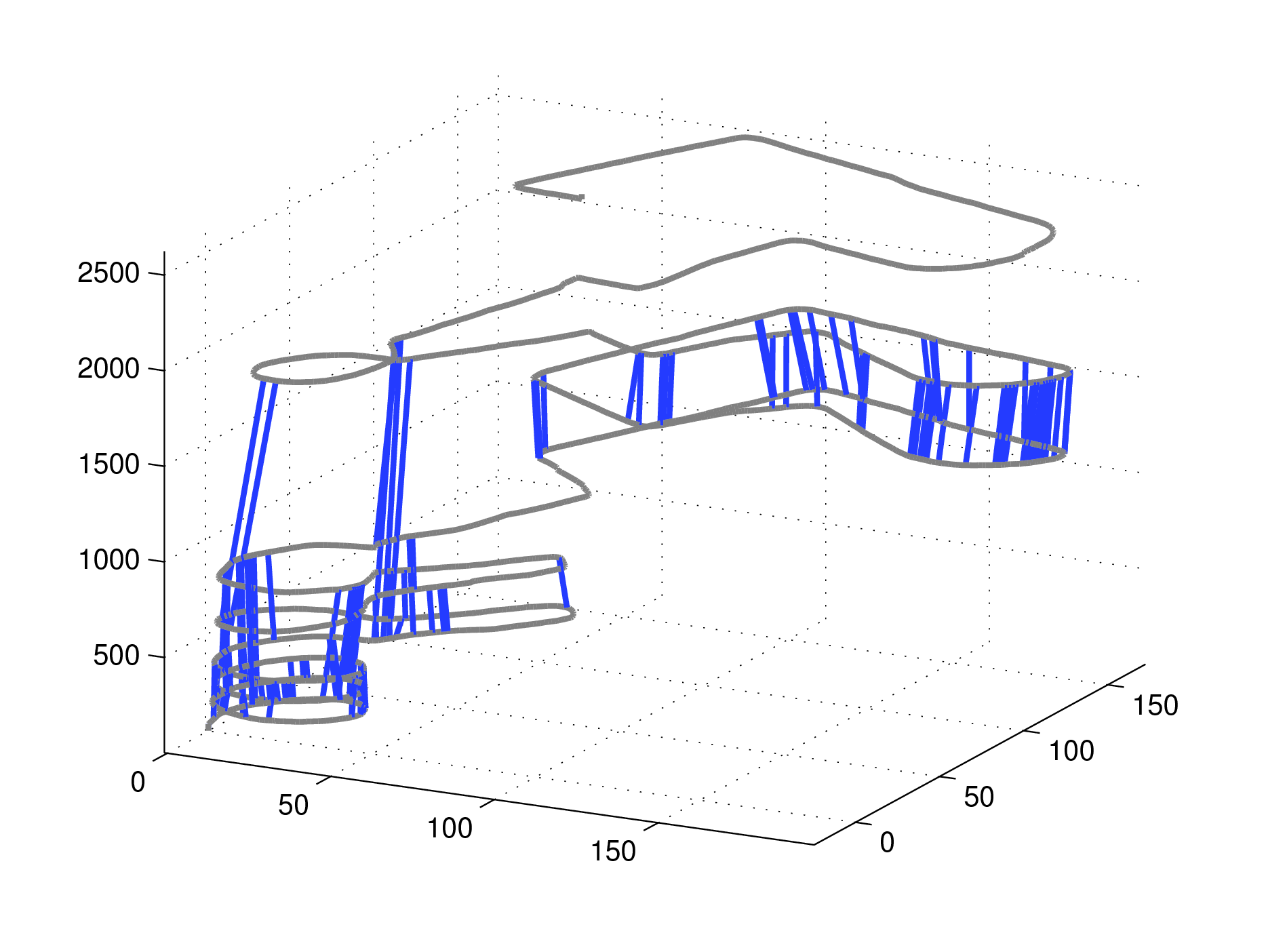}  %, height=4cm, trim=8.31mm 10.58mm 10.6mm 14.7mm, clip
\label{fig:Newcollege-8x6} }  ~~
\subfloat[Basis used as GIST descriptors] %, $\tau=0.9$
{\includegraphics[width=0.3\textwidth]{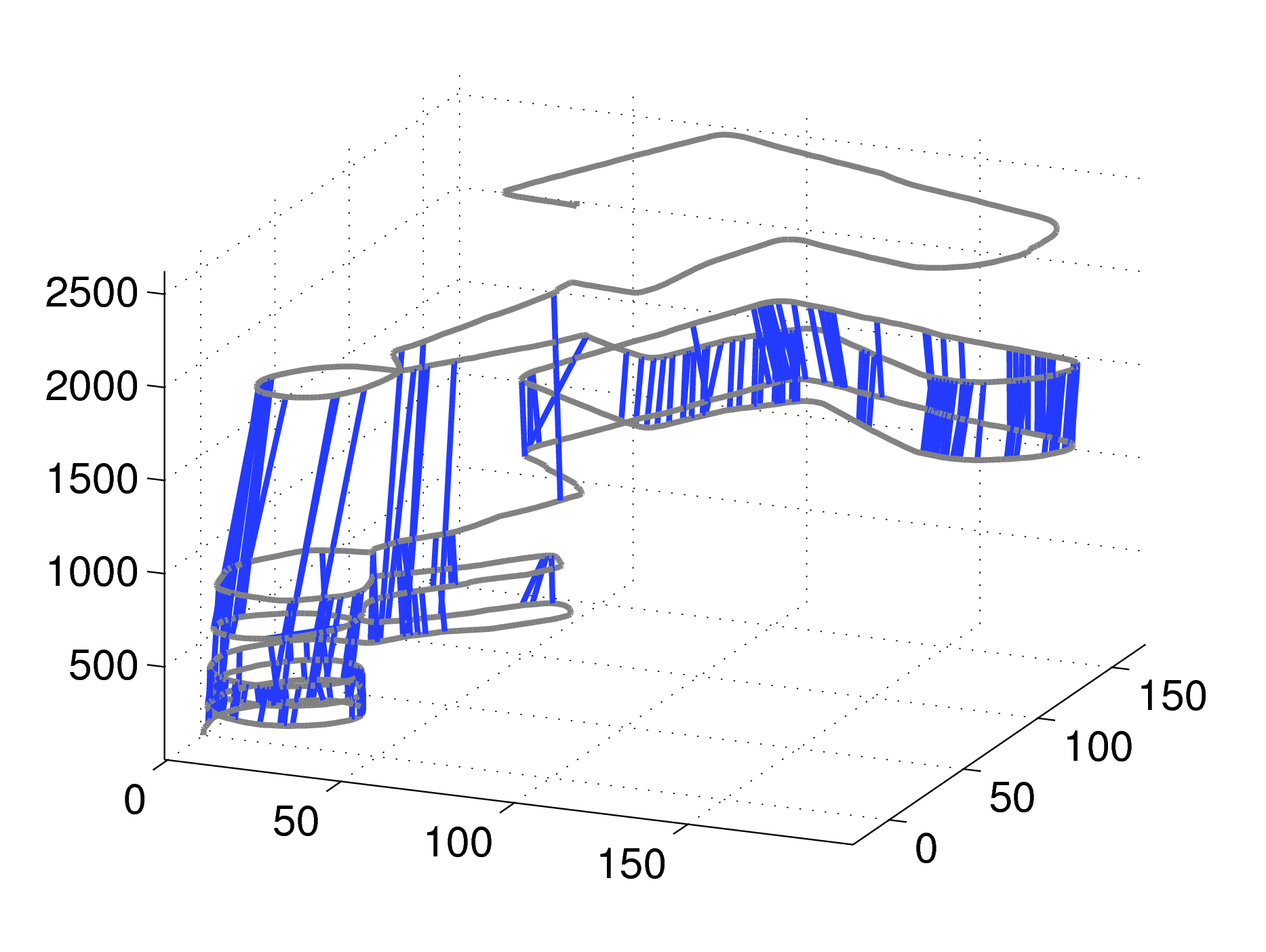}  %, height=4cm, trim=6.01mm 8.58mm 10.6mm 14.7mm, clip
\label{fig:Newcollege-8x6-gist}	} 
\caption{Loop closures detected by the proposed approach using two
different bases for the New College dataset. In these plots, visual
odometry provided with the dataset is shown in gray while the loop closures are shown in blue. 
The $z$-axis represents time (in seconds) and
the $x$- and $y$-axes represent horizontal position (in meters).}
\label{fig:Newcollege}
\end{figure*}

It is interesting to point out that sparse $\ell_1$-minimization inherently is robust to {\em data noise},
which is widely appreciated in computer vision (e.g., see~\cite{cheng2010learning}).
In particular, the sparse noise (error) term in~\eqref{equ:l1-noisy-compact} can account for the presence of dynamic changes or motion blurs.
For example, in~\refFig{fig:newCollege-pics} the dominant background basis explains most of the image, 
while the dynamic elements (which have not been observed before) can be represented by the sparse noise,
and Fig.~\ref{fig:Newcollege} shows that the proposed approach robustly finds these loops.
Such robust performance becomes necessary particularly for long-term mapping 
where the environment often gradually changes over time and thus reliable loop closure in presence of such changes is essential.

As a final remark, the proposed $\ell_1$-minimization-based loop-closure algorithm is also robust to {\em information loss},
which is closely related to the question  raised by \citet{milford2013vision}: 
How much information is needed to successfully close loops?
In this work, we have empirically investigated this problem 
by down-sampling the raw images (which are used as the bases of the dictionary) without any undistortion 
and then evaluating the performance of the proposed approach under such an adverse circumstance.
As shown in Section~\ref{sec:Expr}, 
truly small raw images, even with size as low as $48$ pixels, can be used to reliably identify loops,
which agrees with the findings of~\cite{milford2013vision}.

%}% END \input{closingLoops.tex}
% BEGIN \input{experiments.tex}
%!TEX root = ras_2015.tex
\section{Experimental Results}  \label{sec:Expr}

%This section provides a set of experiments that evaluate the performance of the proposed method. 
%First we provide qualitative results using the New College \cite{smith2009new} dataset using
%both the whole image as the basis vector as well as extracting GIST \cite{oliva2001modeling} 
%descriptors from the images and finding loops based on them. 
%
%In the second set of experiments we investigate various aspects of the proposed framework by 
%looking at different images sizes and how the performance is effected by various parameters of the framework.

To validate the proposed $\ell_1$-minimization-based loop-closure algorithm,
we perform a set of real-world experiments on the publicly-available datasets.
In particular, a qualitative test is conducted on the New College dataset~\cite{smith2009new}, 
where we examine the different types of bases (raw images and descriptors) in order to show 
the flexibility of basis representation of our approach as well as the robustness to dynamics in the scene. 
Subsequently, we evaluate the proposed method on the RAWSEEDS dataset~\cite{rawseeds}
and focus on the effects of the design parameters used in the algorithm.
\red{
Finally, we perform experiments on the KITTI Visual Odometry Benchmark \cite{KITTI}, 
by highlighting the ability of the proposed approach to use different types of deeply learned features as representations,  
as well as the superior performance against a nearest neighbour (NN)-based approach.
}

%which include: (1) image size, (2) threshold ($\tau$), and (3) weight ($\lambda$).

\begin{figure*}[t!]
\centering
\subfloat[Query image]{\includegraphics[height=0.21\textwidth]{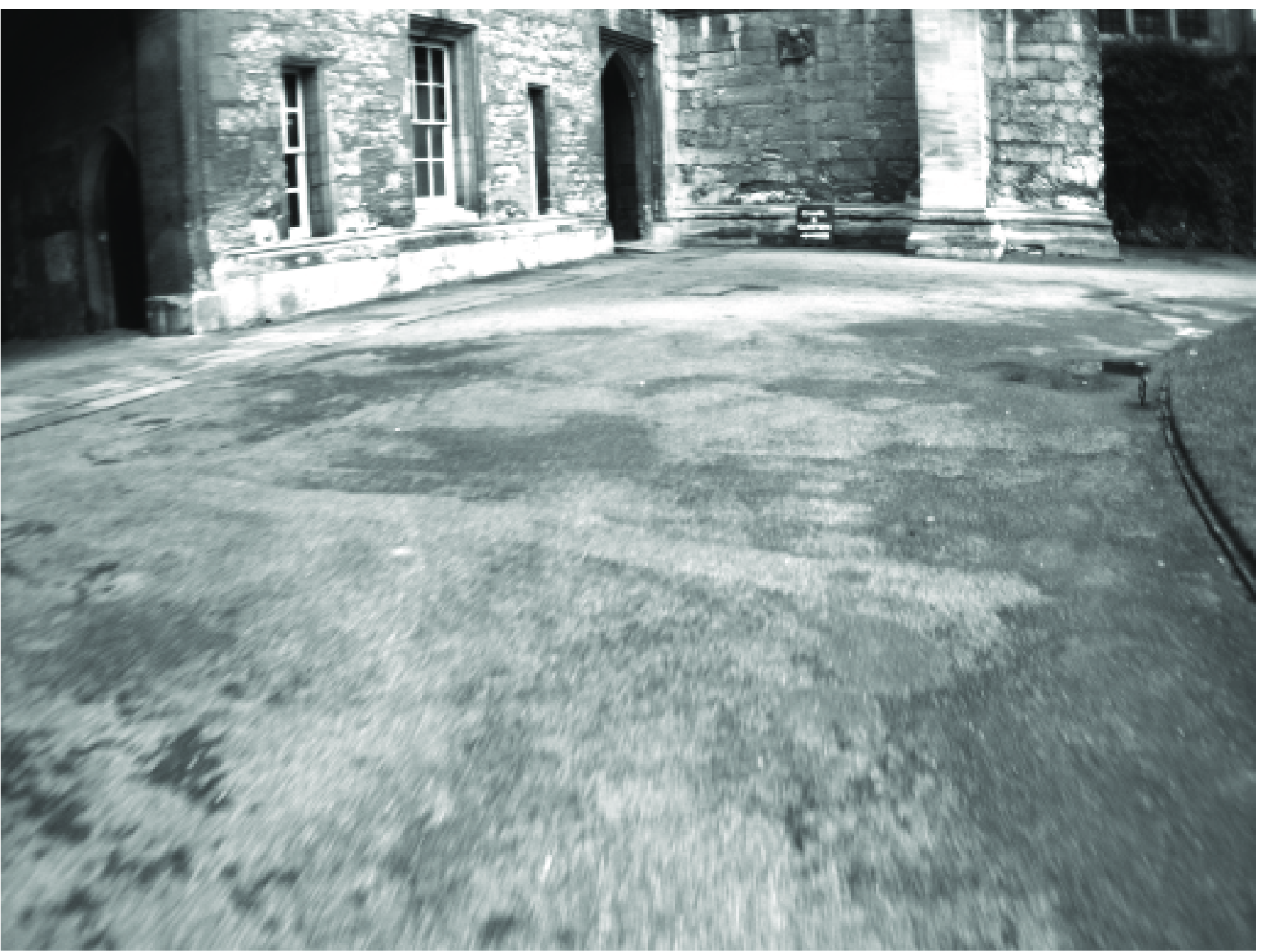} }~~
\subfloat[Match image]{\includegraphics[height=0.21\textwidth]{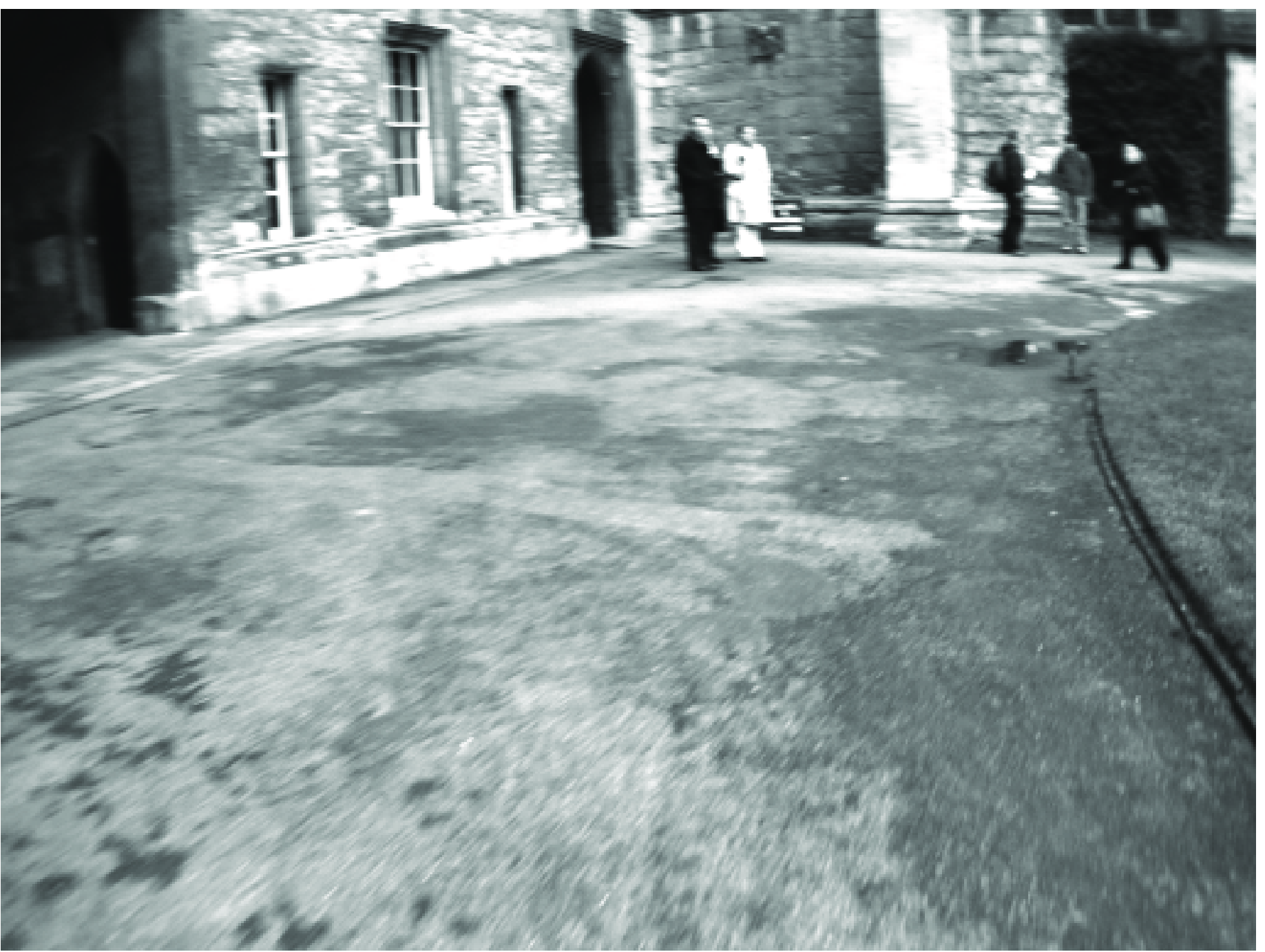} }~~
\subfloat[Noise contributions]{\includegraphics[height=0.215\textwidth]{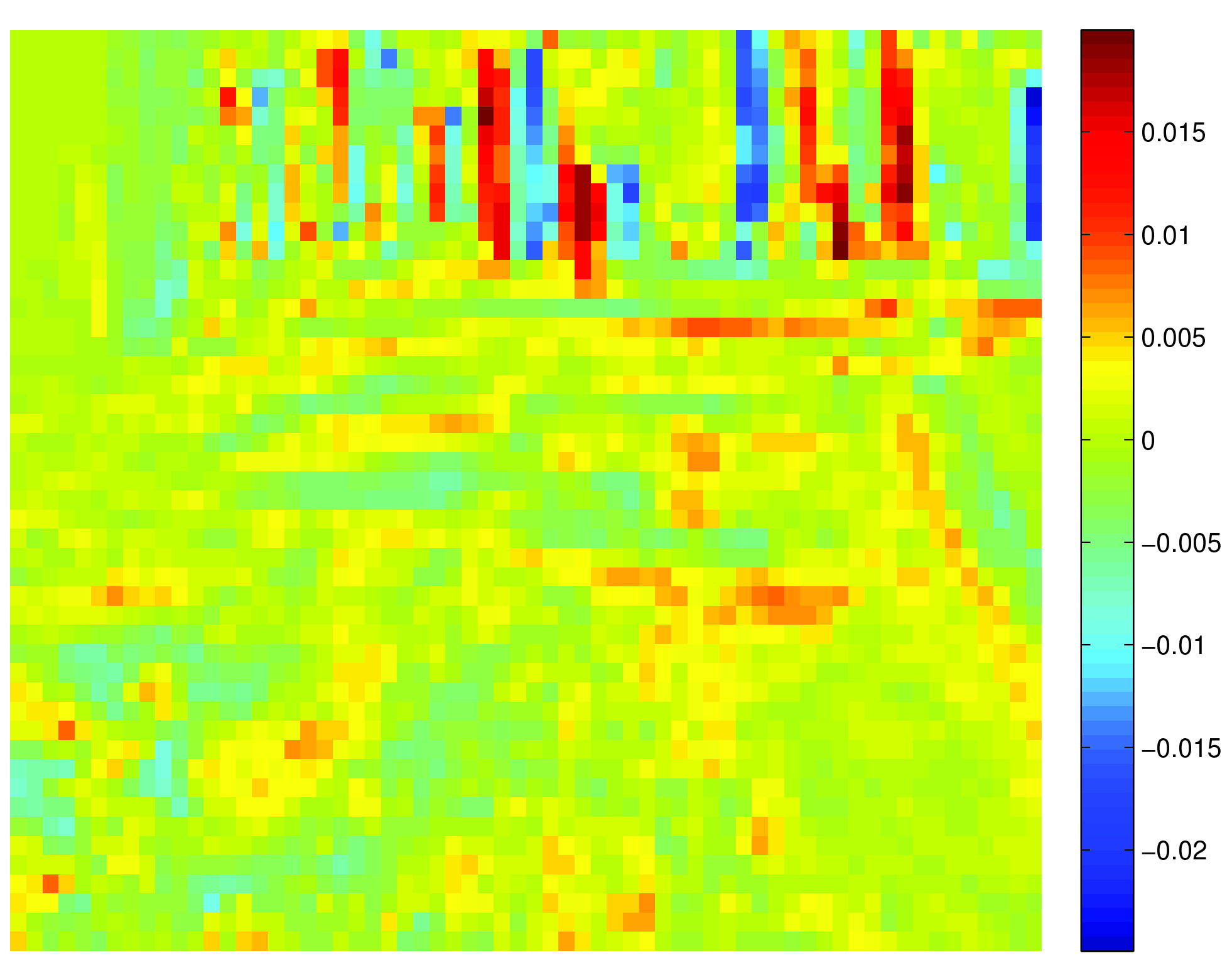} }
\caption{A typical dynamic scenario in the New College dataset: 
When querying a current image to the dictionary that uses $64\times 48$ raw images as its bases, 
the proposed approach robustly finds the correct match -- which however is contaminated with moving people --
by modelling the dynamics as noise.}
\label{fig:robust_NG}
\end{figure*}

\subsection{New College: Different Types of Basis}

The New College dataset~\cite{smith2009new} provides stereo images at $20$ Hz along a $2.2$~km trajectory,
while in this test we only use every $20$th frame giving an effective frame rate of $1$ Hz and in total $2624$ images.
Each image originally has a resolution of $512 \times 384$, but here is down-sampled to either $64\times 48$ or $8 \times 6$ pixels.
%with a $64 \times 64$ reduction in the number of pixels. 
We show below that even under such adverse circumstance, the proposed approach can reliably find the loops. 
The image is scaled so that its gray levels are between zero and one, 
and then is vectorized and normalized as a unit column vector.
For the results presented in this test, 
we use the threshold $\tau = 0.99$ and the weighting parameter $\lambda=0.5$.
Due to the fact that neighbouring images typically are similar to the current one and thus generate false loop closures,
we ignore the hypotheses within a certain time window from the current image and set $t_g=10$ sec,
which effectively excludes the spurious loops when reasoning about possible closures. 
%(i.e., loops within the last $10$ seconds from the current image are ignored). 
Note that $t_g$ can be chosen according to speed of the robot as well as the frame-rate at which the loop closing algorithm is working. 
We also eliminate random matches %can also be seen in \refFig{fig:Newcollege-sparsity}, 
by enforcing a temporal consistency check, 
requiring at least one more loop closure within a time window from the current match. 
We ran all the experiments in Matlab on a Laptop with Core-i5 CPU of 2.5GHz and 16 GB RAM,
and use the homotopy-based method~\cite{asif2008primal} for solving the optimization problem~\eqref{equ:l1-unconstrained}.
%For solving the optimization problem, we use a homotopy based method \cite{asif2008primal} 
%using their publicly-available code\footnote{http://users.ece.gatech.edu/~sasif/homotopy/}. 

%{\color{blue}
The qualitative results  are shown in \refFig{fig:Newcollege}
where we have used {\em three} different bases, 
i.e, down-sampled $64\times 48$ and $8\times 6$ raw images,
%\footnote{In part due to the unoptimized implementation in Matlab,
%it is too costly to use the vectorized full-sized raw images that would have dimension of $512\times 384 = 196608$ 
%as the basis for our proposed approach.
%Thus, we have employed down-sampled raw images of different sizes as the simplest possible basis 
%to show the effectiveness as well as the robustness of our method.}
and GIST descriptors.
In these plots, the odometry-based trajectory provided by the dataset is superimposed by the loop closures detected by the proposed approach,
which are shown as vertical lines connecting two locations where a loop is found. 
All the lines parallel to the $z$-axis represent loop closures that connect the same places at different times.
Any false loops would appear as non-vertical lines, and clearly do not appear in \refFig{fig:Newcollege}, 
which validates the effectiveness of the proposed method in finding correct loops.
These results clearly show the flexibility of bases representation of the proposed method. 
In particular, instead of using the different down-sampled raw images as bases,
our approach can use the GIST descriptors, $\mathbf{GIST(b)} \in \mathcal R^{512} $,  which are computed over the whole image,
and is able to detect the same loop closures as with the raw images. 
%(see \refFig{fig:Newcollege-8x6} vs. \refFig{fig:Newcollege-8x6-gist}).
%}

\begin{figure}[t!]
\centering
 \includegraphics[width=0.75\textwidth]{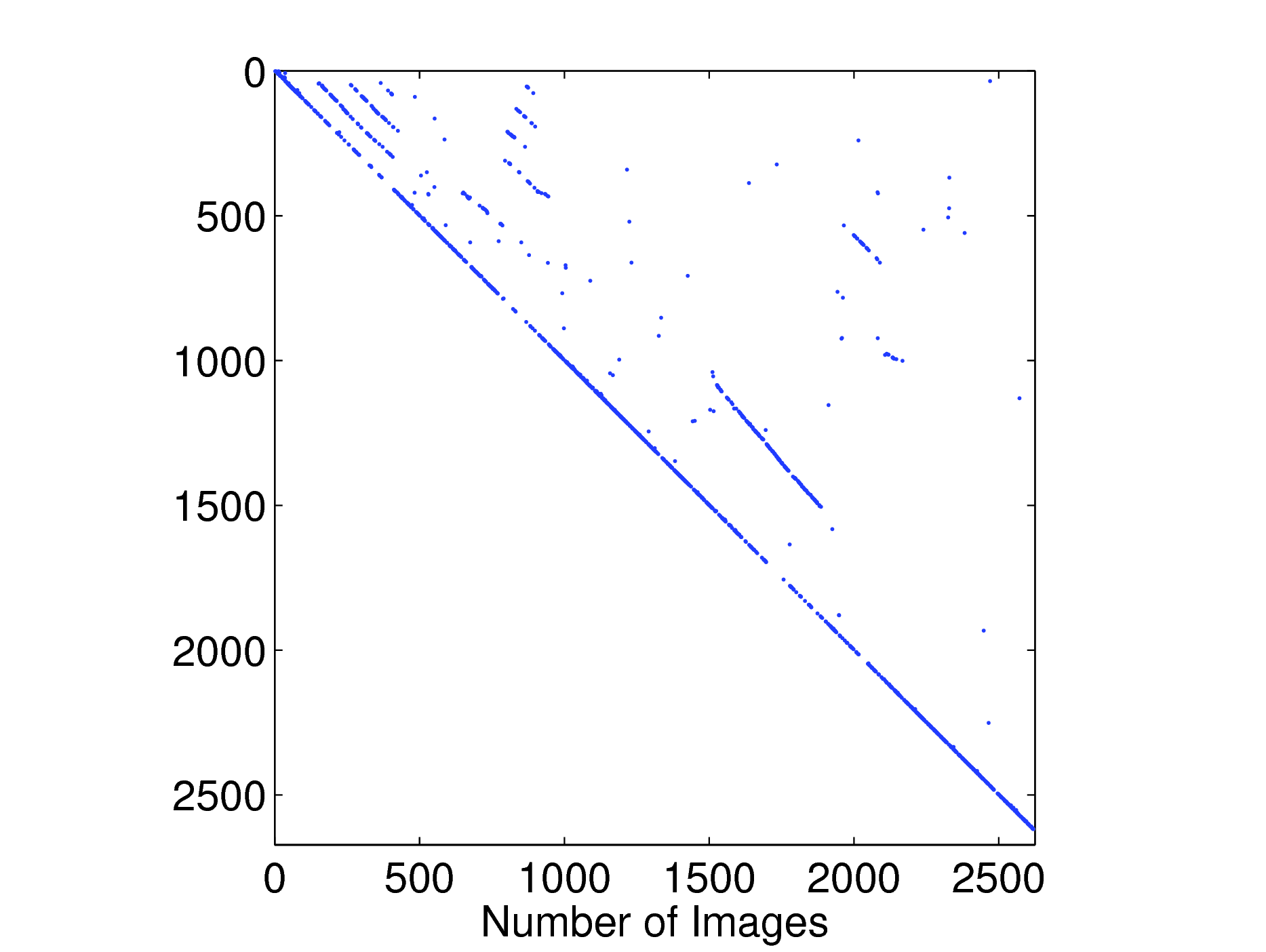}
\caption{Sparsity pattern induced by solving~\eqref{equ:l1-unconstrained} for {\em all} the images in the New College dataset. 
The $i$-th  column corresponds to the solution for the $i$-th image,
and the non-zeros are the values in each column that are greater than $\tau=0.99$. 
%after the vector has been normalized to unit length. 
Note that the main diagonal occurs due to the current image being best explained by its neighboring image,
while the off-diagonal non-zero elements indicate the loop closures. 
%The first three runs in the same circular area in the New College dataset can be seen as the smaller off diagonal lines in the top-left. 
%Further details can be found in the text.
}
\label{fig:Newcollege-sparsity}
\end{figure}

An interesting way of visualizing the locations 
where loop closures occur is to examine the sparsity pattern of the solution matrix,
which is obtained by stacking all the solutions, ${\isVec{\hat\alpha}_i}$, 
for all the queried images in a matrix.
\refFig{fig:Newcollege-sparsity}   shows such a matrix 
that contains non-zero values in each column corresponding to the elements greater than the threshold $\tau$. 
In the case of no loop closure, 
each image can be best explained by its immediate neighbour in the past,
which gives rise to non-zeros along the main diagonal. 
Most importantly, the off-diagonal non-zeros indicate the locations where loops are closed. 
It is interesting to see that there are a few sequences of loop closures 
appearing as off-diagonal lines in \refFig{fig:Newcollege-sparsity}. 
This is due to the fact that the first three runs in the circular area at the beginning of the dataset,
correspond to the three off-diagonal lines in the top-left of the matrix;
while a sequence of loop closures detected in the lower part of New College, 
correspond to the longest line parallel to the main diagonal. 

%{\color{blue}
It is important to note that although both dynamic changes and motion blurs occur in the images,
the proposed approach is able to reliably identify the loops (e.g., see \refFig{fig:newCollege-pics}),
which is attributed to the sparse error used in the $\ell_1$-minimization [see~\eqref{equ:l1-noisy-compact}].
To further validate this robustness  to dynamics,
Fig.~\ref{fig:robust_NG} shows a typical scenario in the New College 
where we query a current image with no dynamics to the dictionary that uses $64\times 48$ down-sampled raw images as its bases,
and the correct match is robustly found, which however contains moving people.
Interestingly, the dominant noise contributions (blue) as shown in \refFig{fig:robust_NG}(c),
mainly correspond to the locations where the people appear in the match image.
This implies that the sparse error in~\eqref{equ:l1-noisy-compact} correctly models the dynamic changes.
%
%}

%Since the problem is sparse by construction (i.e., there can be only one match for any current image),
%the homotopy-based method efficiently solves the optimization problem. 
%For this experiment, the mean running time for generating a single loop-closure hypothesis is $18.06$ ms,
%with the fastest of $3.6$ ms and the slowest of $66.25$ ms.

\subsection{RAWSEEDS: Effects of Design Parameters}

%\begin{figure}[t!]
%\centering
% \includegraphics[width=0.24\textwidth]{figs/Bicocca_1.png} 
% \includegraphics[width=0.24\textwidth]{figs/Bicocca_2.png}
%\caption{Sample distorted images from the Bicocca dataset.}
%
%\label{fig:bicocca_samples}
%\end{figure}

To further test the proposed algorithm, we use the Bicocca 25b dataset from the RAWSEEDS project \cite{rawseeds}.
The dataset provides the laser and stereo images for a trajectory of $774$~m. 
We use the left image from the stereo pair sampled at $5$~Hz, resulting in a total of $8358$ images. 
Note that we do {\em not} perform any undistortion and work directly with the raw images coming from the camera.
%(e.g., see~\refFig{fig:bicocca_samples} for some sample images).
%
%
In this test, we focus on studying the effects of the most important parameters used in the proposed approach,
and evaluate the performance based on precision and recall.
Specifically, precision is the ratio of correctly detected loop closures over all the detections. 
Thus, ideally we would like our algorithm to work at full precision. 
On the other hand, recall is the percentage of correct loop closures that have been detected over all possible correct detections. 
A high recall implies that we are able to recover most of the loop closures.

\subsubsection{Threshold $\tau$ and weight $\lambda$}

\begin{figure}[t!]
\centering
 \includegraphics[width=0.45\textwidth]{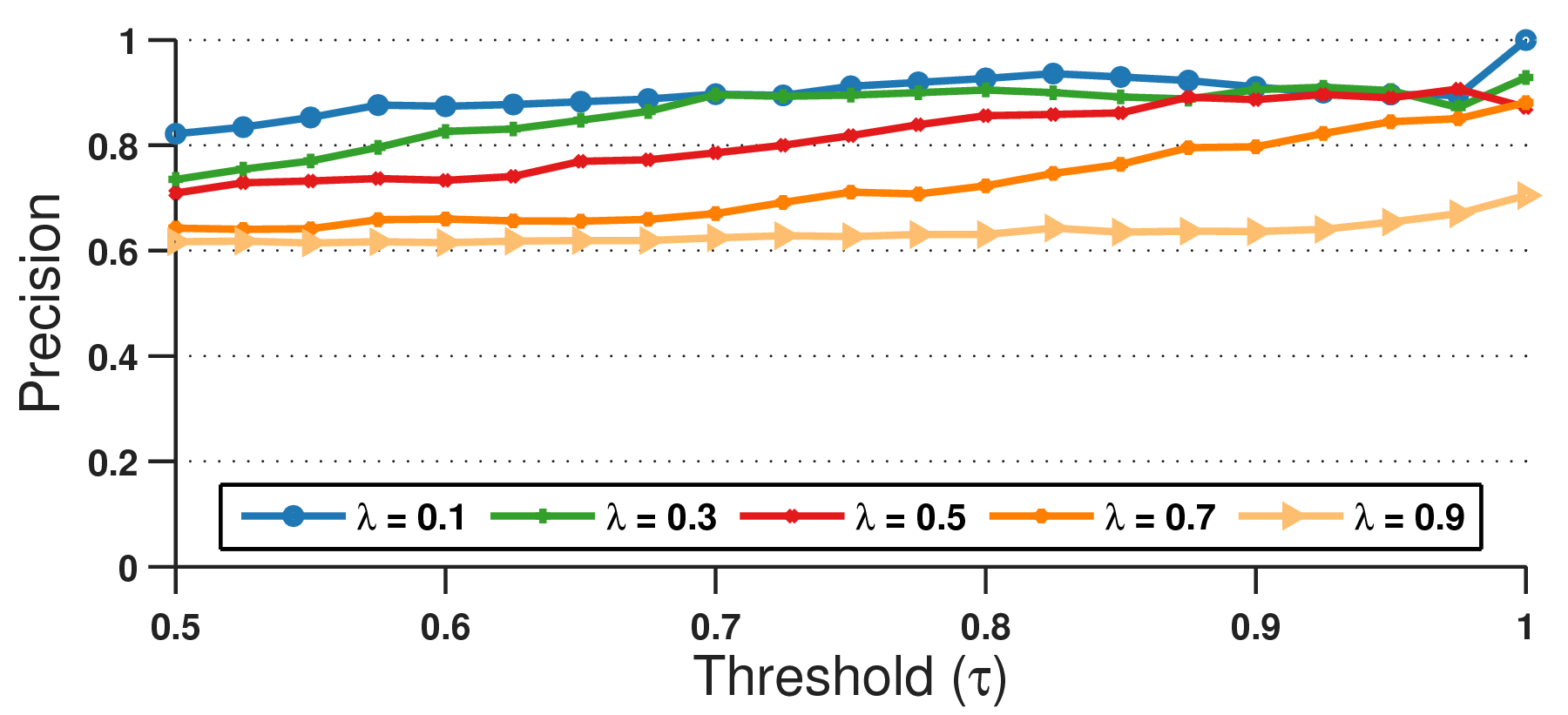}
 \includegraphics[width=0.45\textwidth]{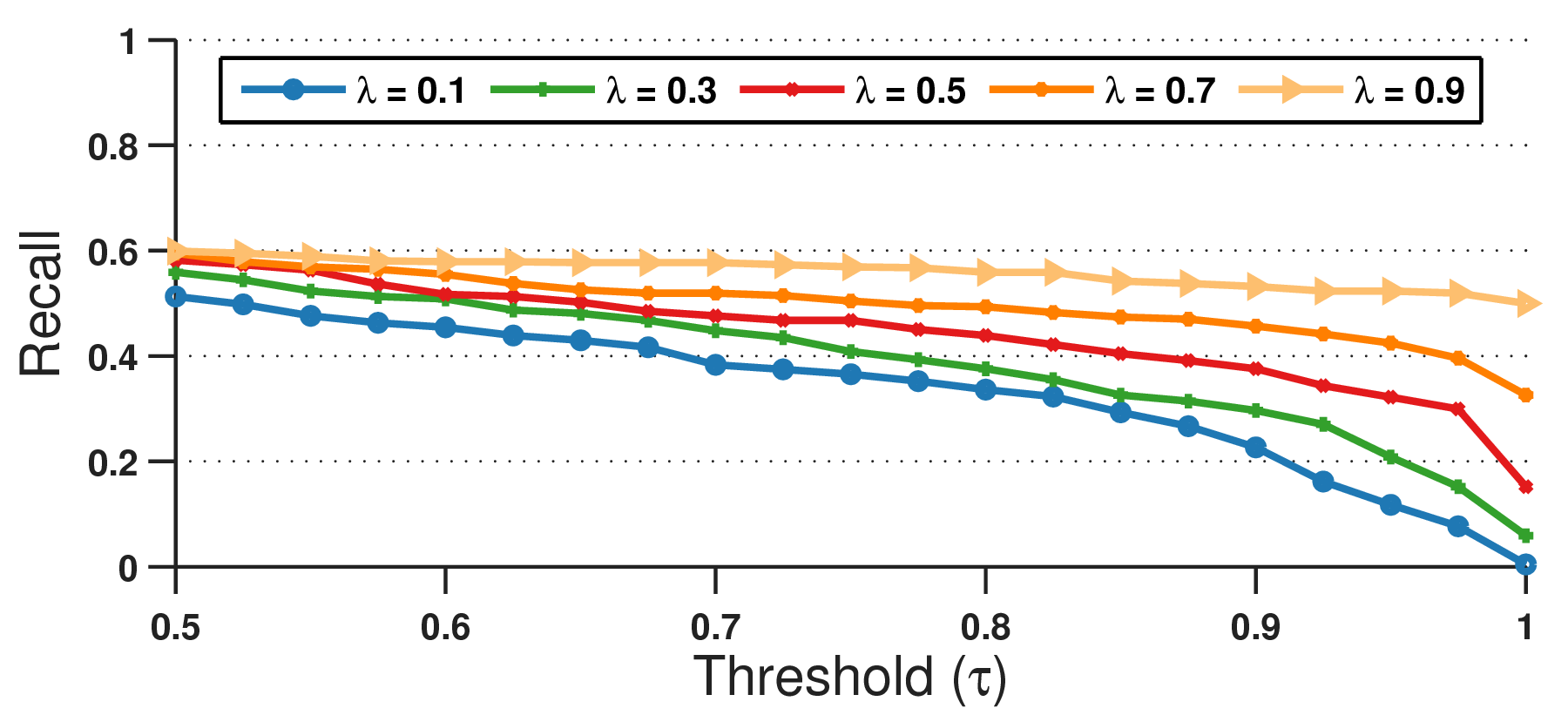}
\caption{Precision and recall curves for the Bicocca  dataset while using the $20 \times 15$ raw images. 
%In these plots, the threshold $\tau$ varies along the $x$-axis, and each curve represents a fixed value of the weight parameter $\lambda$. 
}
\label{fig:varying-tau}
\end{figure}

We first examine the acceptance threshold $\tau$, whose valid values range from 0.5 to 1. 
This parameter can be thought of as the similarity measure between the current 
image and the matched image in the dictionary. In order to study the effect of 
this parameter on the precision and recall, we vary the parameter for a fixed image of $20\times15$ pixels. 
Moreover, we are also interested in 
if and how the weighting parameter $\lambda$ impacts the performance 
and thus vary this parameter as well. 

The results are shown in \refFig{fig:varying-tau}. 
As expected, the general trend is that a stricter threshold (closer to 1) leads to 
higher precision, and as a side effect, a lower recall. 
This is because as the threshold increases, we get fewer loop closing hypotheses but a larger proportion of them is correct. 
Note that this dataset is challenging due to the perceptual aliasing in many parts of the trajectory;
the matched images are visually similar but considered as false positives
since the robot is physically not in the same place. 
%
%

%Regarding the effect of the weighting parameter on the performance of the proposed approach,
\red{Interestingly,  \refFig{fig:varying-tau} also shows that 
the smaller $\lambda$ leads to the higher precision but the lower recall.
This seems to counter the intuition that 
the sparser the solution of \eqref{equ:l1-unconstrained} (by using a larger $\lambda$), the higher fidelity  of the loop-closure detection.
However, it should be noted that this intuition  motivates the proposed sparse formulation 
but does not guarantee that the optimal solution is sparsest (1-sparse) as discussed in Section~\ref{sec:uniqueness},
which heavily  depends on the quality of the data at hand (e.g., signal-to-noise ratio, the similarity level of the revisiting images).
It is important to note that there are two goals to reconcile in our sparse formulation \refEqu{equ:l1-unconstrained}: 
(i) the reconstruction error   represented by the $\ell_2$ term, and (ii) the sparsity level of the solution encoded in the $\ell_1$ term.
A smaller value of this parameter  results in a better data-fitting  solution of smaller reconstruction error,
%which is closer to the corresponding dense least-square solution [see the second term of \refEqu{equ:l1-unconstrained}],
hence requiring the images to be as visually similar as possible but at the same time, lowering the contribution of the greatest basis vector. 
%
%As a rule of thumb, we suggest higher $\tau$ and lower $\lambda$ values. 
}

\subsubsection{Image size}

\begin{figure}[t!]
\centering
 \includegraphics[width=0.45\textwidth]{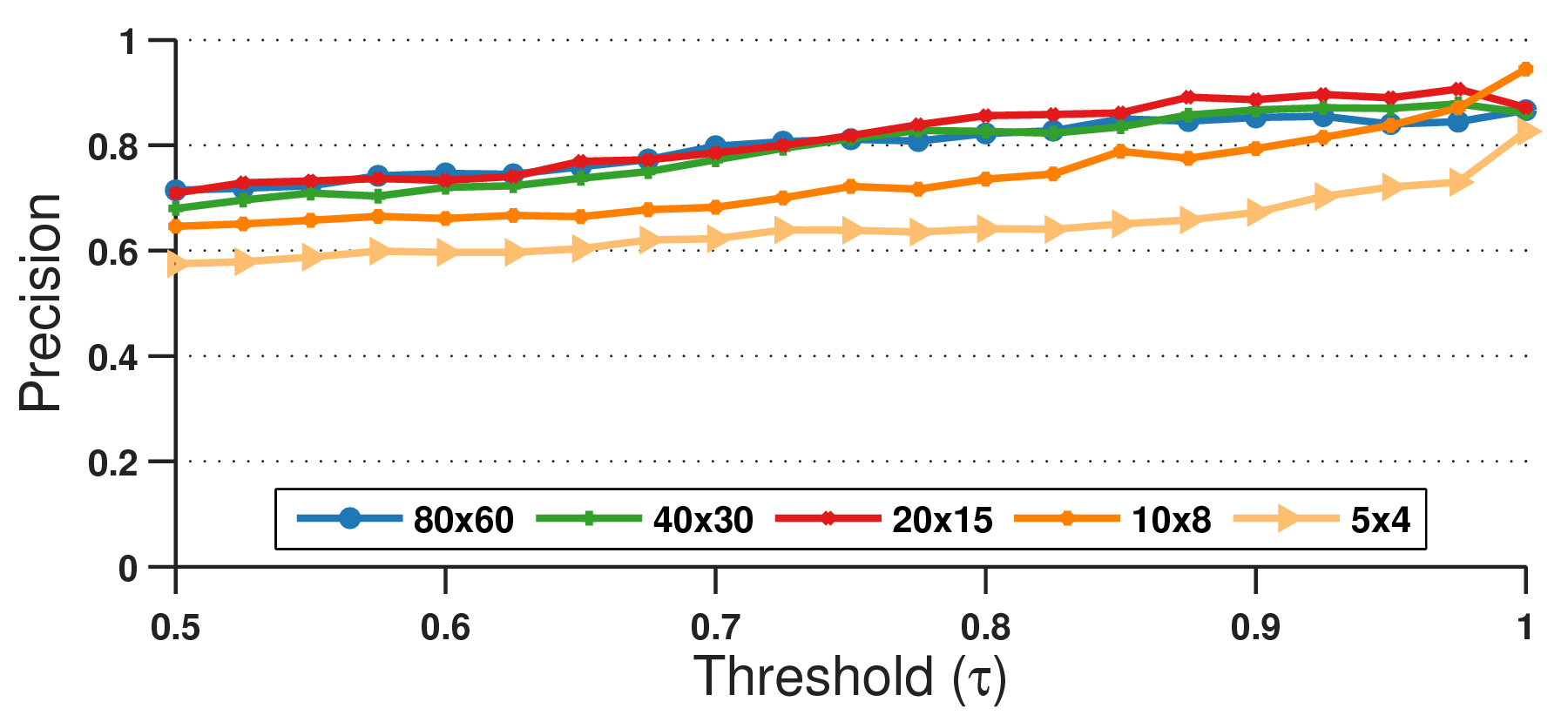}
 \includegraphics[width=0.45\textwidth]{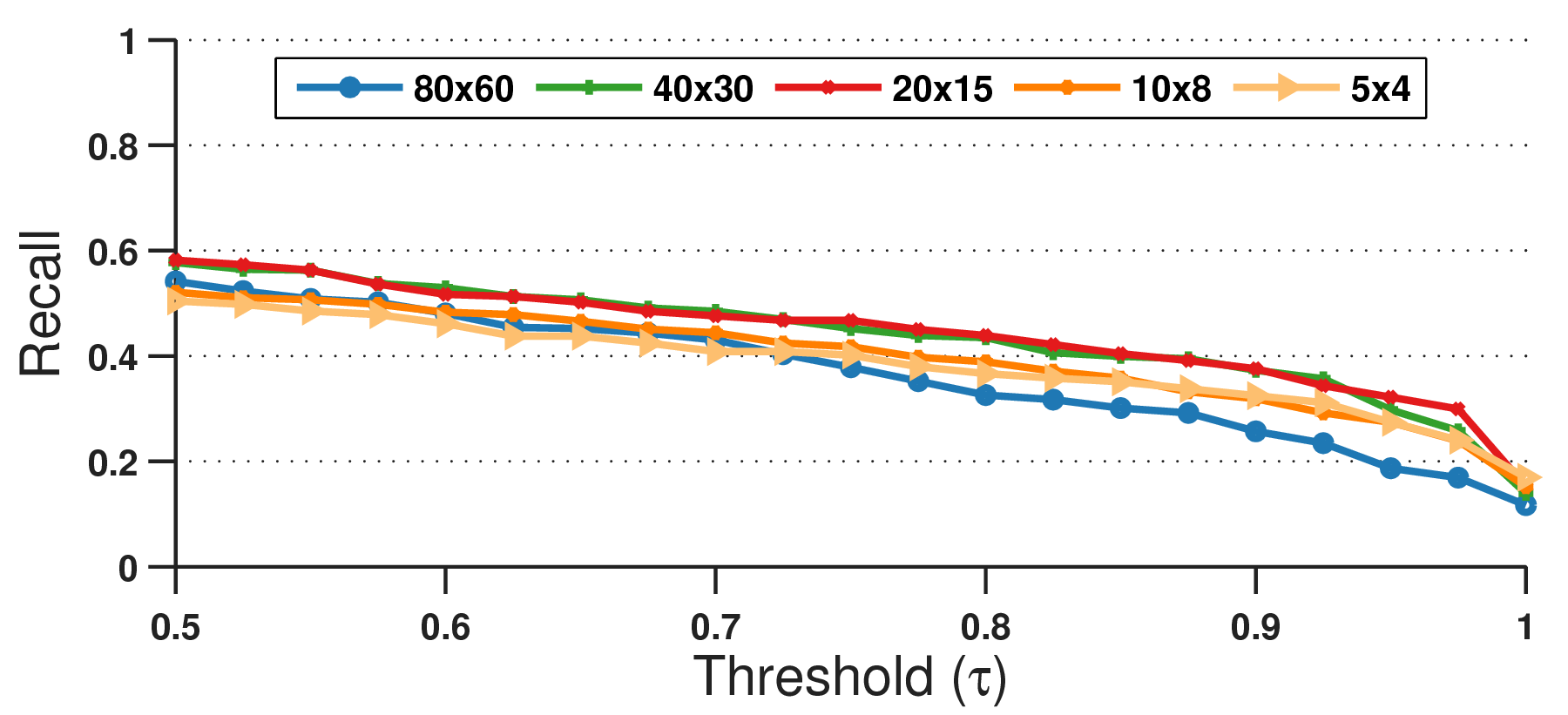}
\caption{Precision and recall curves for the Bicocca  dataset while fixing the weighting parameter $\lambda= 0.5$. 
%In these plots, the threshold $\tau$ varies along the $x$-axis, and each curve represents a fixed image size.
}
\label{fig:varying-imsize}
\end{figure}

Inspired by the recent work~\cite{milford2013vision},
we also examine the performance difference by varying image sizes 
and see if we can obtain meaningful results using small-size images. 
The original image size from the Bicocca dataset is $ 320 \times 240$,
and the first image size we consider is $80 \times 60$ which is a reduction of 
a quarter in each dimension. For each successive experiment, we half the size in each dimension, which results 
in images of size $40 \times 30$, $20 \times 15$, $10 \times 8$, and finally $5 \times 4$. 
The weighting parameter $\lambda$ is fixed to be 0.5. 
Precision and recall curves are generated by varying 
the acceptance threshold $\tau$ are shown in \refFig{fig:varying-imsize}.

%It becomes clear from~\refFig{fig:varying-imsize}
%that even for a very small images of size $10 \times 8$, our method is able to achieve full precision with a reasonable recall. 
It is clear from~\refFig{fig:varying-imsize} that the curves are tightly coupled and undergo the same behaviour 
for each image size. Precision curves for the three largest image sizes overlap each other, 
showing that we can generate the same quality of loop closure hypotheses using any of the image sizes. 
These plots show a graceful degradation as the image size decreases. Considering that the image of 
size $10 \times 8$ is a factor of $960$ times smaller than the original image, our method is able to 
distinguish places based on very little information,
which agrees with the findings of~\cite{milford2013vision}.

\subsubsection{Execution time}

Since the proposed method solves an optimization problem in a high-dimensional space, 
it is important to see how long the method takes to come up with the loop-closing hypotheses. 
Despite that each image is an $r \times c$ vector for an image with $r$ rows and $c$ columns, 
and at the end of the experiment we have nearly $8500$ images, the computation is very efficient 
thanks to the sparsity induced  by the novel formulation. 
Most of our solutions are expected to 
be $1$-sparse (i.e., we expect only one non-zero if the current image matches perfectly 
one of the basis vectors in the dictionary), and thus the homotopy-based solver  
performs efficiently as shown in Table \ref{tab:timing}. 
For the largest image size, 
the mean time is $117$~ms with a maximum less than half a second. 
The proposed method works  
well on small images such as $20 \times 15$, which take on average $3.7$ ms. The runtime 
gradually grows as the number of basis vectors increases. 
The timing information given in Table. \ref{tab:timing} shows that the current method can run fast enough for real time operation at above 5Hz for the largest image size considered.

%In general, even with the current unoptimized Matlab implementation, the execution is fast enough 
%to be used in real-time operations with runtime well below a second in almost all cases. 
%This efficiency is due to the homotopy methods that solve the $\ell_1$-norm minimization problem by taking advantage of its properties. 

Interestingly, we found $\lambda = 0.5$ is a good trade-off between precision/recall and computational cost. 
In general, a higher threshold $\tau$ would lead to fewer high-quality loop closures. 
This parameter can be designed based on the application in question.
Similarly, images of size larger than $20\times 15$ 
do not provide great improvement in terms of precision/recall. 
Thus, the choice of image size should take into account the complexity of the environment being modelled.
In an environment  (e.g., outdoors) where there is rich textural information, smaller images may be used. 
If the environment itself does not contain a lot of distinguishable features, 
larger images can be used in order to be able to differentiate between them.

\begin{table}
\centering
\caption{Execution time for different image sizes. Note that at the end, the dictionary has a size of feature dimension + 8358 (number of basis).}
\begin{tabular}{ccccccc}
\hline
 size (feature dimension) && min (ms) & mean (ms) & max (ms) & std (ms)\\
\hline %\hline
$80 \times 60 ~(4800) $  &&  	45.762 & 116.45 & 417.290 & 43.229 \\ 
\hline
$40 \times 30 ~(1200) $  &&   2.689 & 17.334 & 70.940 & 9.500 \\ 
\hline
$20 \times 15 ~(300)$  &&   0.111 & 3.662 & 20.828 & 2.904 \\ 
\hline
$10 \times  8 ~(80) $  &&   0.029 & 0.707 & 4.448 & 0.552 \\ 
\hline
$ 5 \times  4 ~(20) $  &&   0.021 & 0.410 & 2.541 & 0.3163 \\  
\hline
\end{tabular}
\label{tab:timing}
\end{table}
%

%{\color{blue}
\subsection{Comparison to DBoW}

\begin{figure}[t!]
\centering
\includegraphics[width=0.7\textwidth]{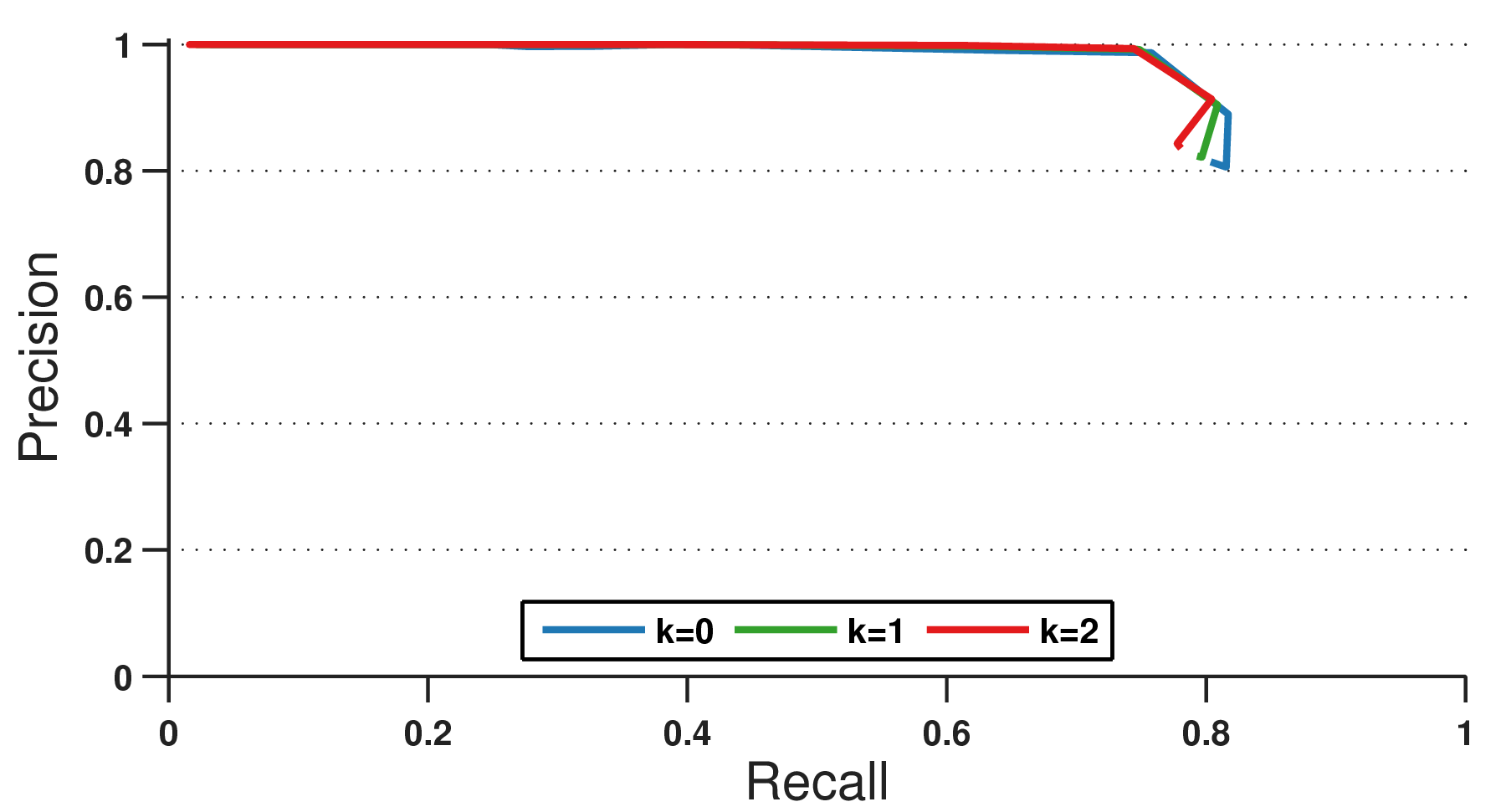} 
\caption{Precision and recall curves of DBoW~\cite{GalvezTRO12} for the Bicocca dataset using the full-sized $320 \times 240$ images. 
In this plot, $k$ denotes the values used for temporal consistency constraint.%, while the symbol $\ast$ indicates no geometric verification.
}
\label{fig:comparision}
\end{figure}

In this section, we compare the performance of the proposed method against the state-of-the-art DBoW algorithm~\cite{GalvezTRO12} on Bicocca 25b dataset. 
%which is publicly available from the author's website\footnote{http://webdiis.unizar.es/~dorian/index.php?p=32}. 
%%
%In order to compare both the methods, we test both methods under as similar conditions as possible.
For the DBoW, we operate on the full-sized $320 \times 240$ images, 
using different temporal constraints ($k=0,1,2$) along with geometric checks enabled.
Its performance is controlled by a so-called confidence parameter $\alpha \in[0,1]$.
We sweep over the values of this parameter and compute the precision and recall for each $\alpha$,
which are shown in~\refFig{fig:comparision}.

%To better visualize the comparison, 
%we show the values of the proposed approach in~\refFig{fig:varying-imsize} 
%as the precision-recall curves in~\refFig{fig:l1-PR}.
%%
%It is important to point out that although this is {\em not} a fair comparison 
%because (i)~the DBoW employs the offline learned dictionary and thus utilizes more information than the proposed online approach,
%and (ii)~the DBoW extracts the BoW from the full-sized images while our method uses the down-sampled distorted raw images as basis.
%%
%Albeit, the proposed approach attains comparable performance in many cases.
%Specifically, as seen from Figs.~\ref{fig:l1-PR} and~\ref{fig:comparision}, 
%the best precision-recall curve of our method competes that in the simple case of the DBoW ($k=0^*$),
%and the proposed algorithm is more conservative and operates with comparatively lower recall. 
%%
%As evident from~\refFig{fig:comparision} that the geometric verification plays an important role in achieving full precision,
%we can also benefit from further enforcing a geometric verification step.
%%

\begin{figure}[t!]
\centering
\includegraphics[width=0.7\textwidth]{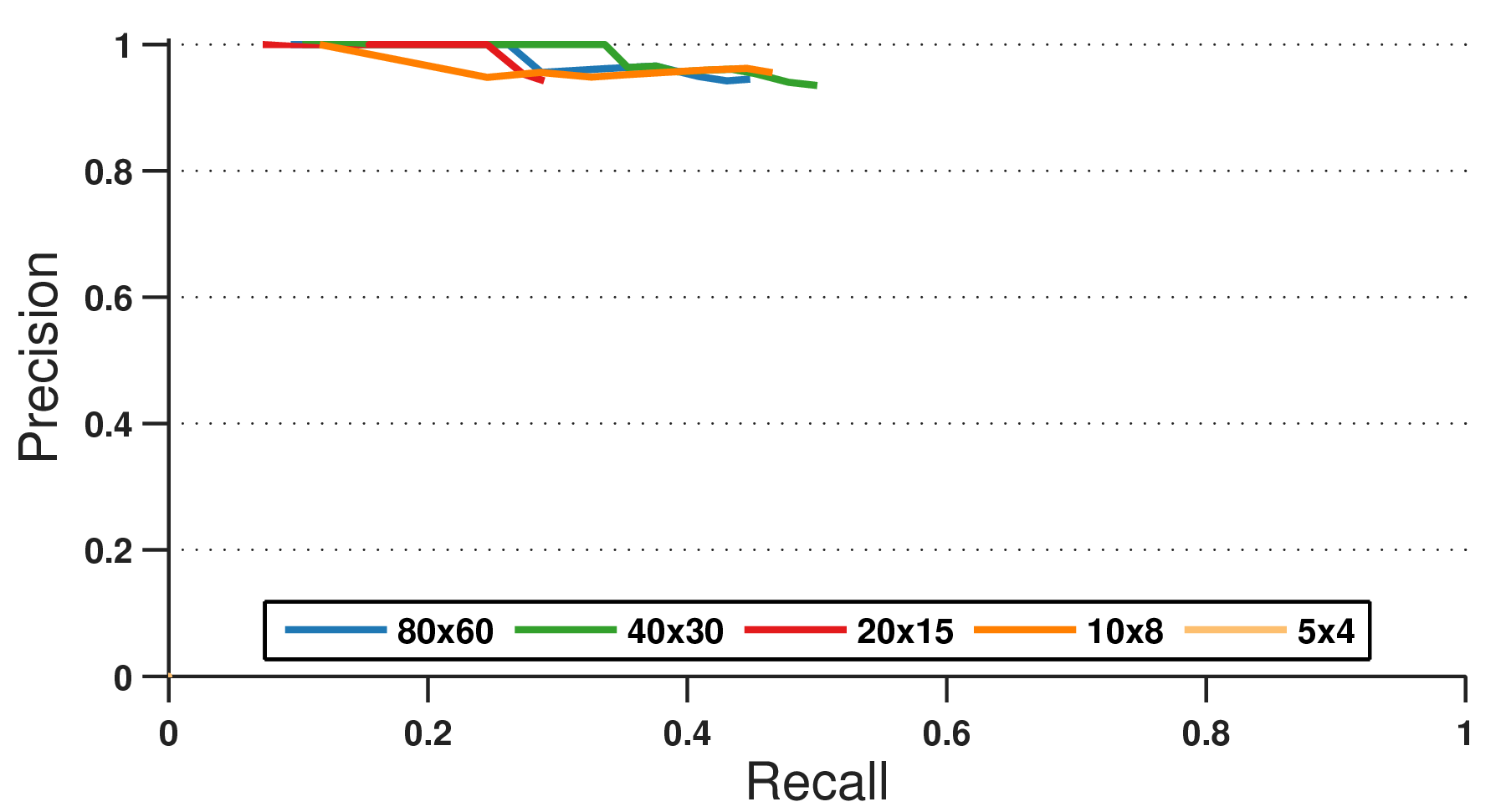} 
\caption{Precision and recall  corresponding to Fig.\ref{fig:varying-imsize} with an additional geometric verification step, same as the one used in DBoW.}
\label{fig:l1-PR}
\end{figure}

%To better visualize the comparison, 
%we show the values of the proposed approach in~\refFig{fig:varying-imsize} as the precision-recall curves in~\refFig{fig:l1-PR}. 
%%
For the purpose of a fair comparison, we carry out the same geometric verification step in DBoW (\refFig{fig:comparision}) and in the proposed method (\refFig{fig:l1-PR}): feature extraction, matching and fitting a fundamental matrix between the matched features. 
If sufficient support is not found for the fundamental matrix, the proposed hypothesis is rejected. 
This geometric verification is carried out on the full resolution images.

%with an additional geometric verification step that involves: feature extraction, matching and fitting a fundamental matrix between the matched features.
%If sufficient supporting matches are not found for the fundamental matrix, the proposed hypothesis is rejected. 
%This geometric verification is carried out on the full resolution images.
%%  

%In comparing our method to the DBoW algorithm, it is important to keep in mind that the
%DBoW employs an {\em offline} learned dictionary, thus utilizing more information than the proposed {\em online} approach. 
%In addition, DBoW extracts visual words from the {\em full-sized} images, 
%while our method uses the {\em down-sampled} distorted raw images as the basis vectors at the hypotheses generation stage.

%Despite this unfavourable comparison, the proposed approach attains comparable performance. 
%In particular, 
As seen from Figs.~\ref{fig:comparision} and~\ref{fig:l1-PR} ,
the best precision-recall curve of our method competes with that of the DBoW in terms of precision;
moreover, the proposed algorithm is more conservative and operates with {\em lower} recalls. 
This low recall is a consequence of requiring the match to be globally unique in order to be considered a loop closure.
Overall, 
%despite performing at a lower precision for some recall than the DBoW algorithm,
%on the same but down-sampled input sequences, 
the proposed approach achieves competitive trade-off between precision and recall.

\subsection{Deep Features}
\label{sec:DeepExpr}

Recently, many computer vision tasks have shown great boost in performance by using features from Convolutional Neural Networks (CNNs).
Instead of using hand-crafted features, these networks learn features from data, which are a more ``natural'' representation for tasks such as image segmentation and object detection
and are shown to outperform hand-crafted features such as SIFT and GIST, 
as shown in the PASCAL Visual Object Classes (VOC) Challenge \cite{pascal-voc-2012}.
To evaluate that these deep features can also be used in the proposed  loop closing framework, 
we further test on the KITTI Visual Odometry benchmark \cite{KITTI}.
The datasets consists of 21 trajectories recorded using multiple sensors,
among which the ground truth is provided for 10 and only 6 contain at least one loop closure.
The ground truth for the loop closures is used from the dataset presented in \cite{latif2014robust},
where loop closures are calculated at 2Hz for the six trajectories with ground truth.
In particular, we use the deep network presented in \cite{VijayNet}. The network learns to represent a given $64 \times 64$ image patch in $\mathcal R^{256}$ for the task of patch matching.
For the experiments presented here, we use this network as a feature extractor and employ different strategies.

\begin{figure}[t!]
\centering
\subfloat[Sequence 00]{ \includegraphics[width=0.5\textwidth]{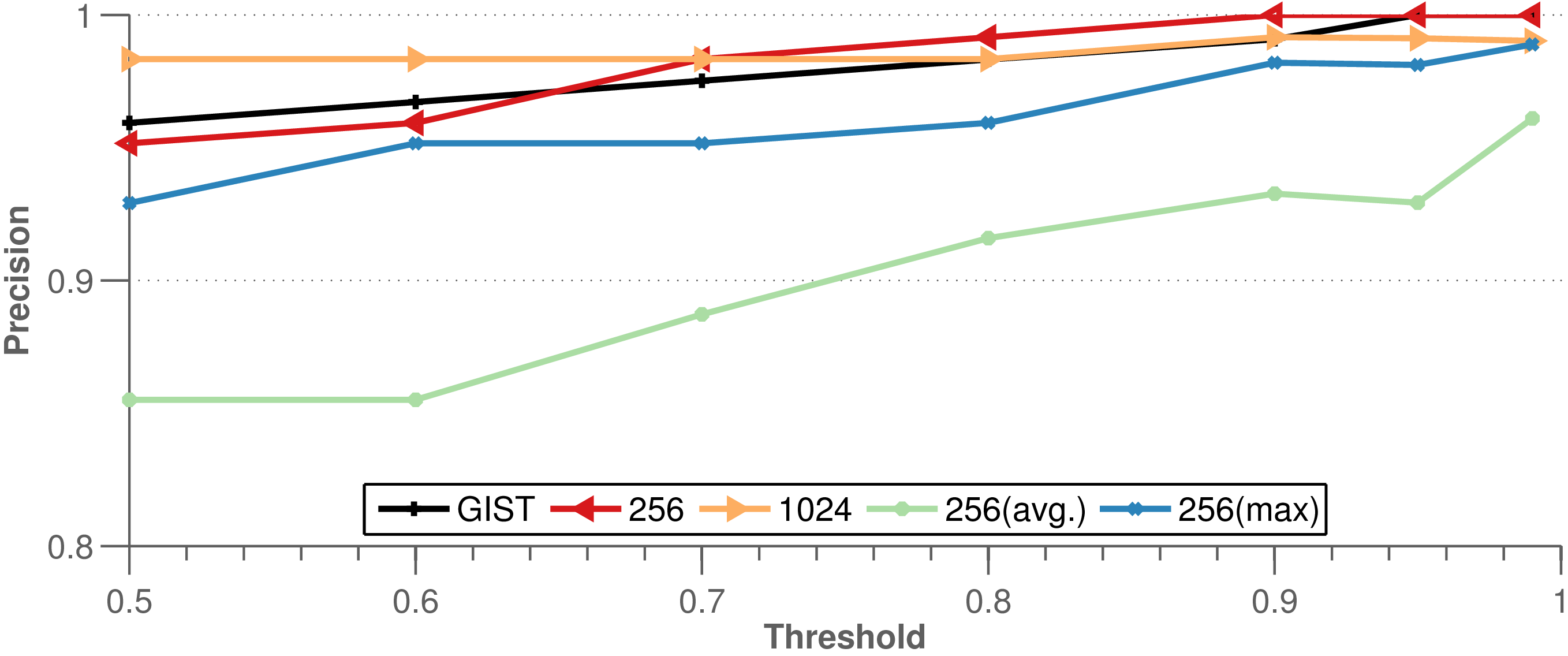} \includegraphics[width=0.5\textwidth]{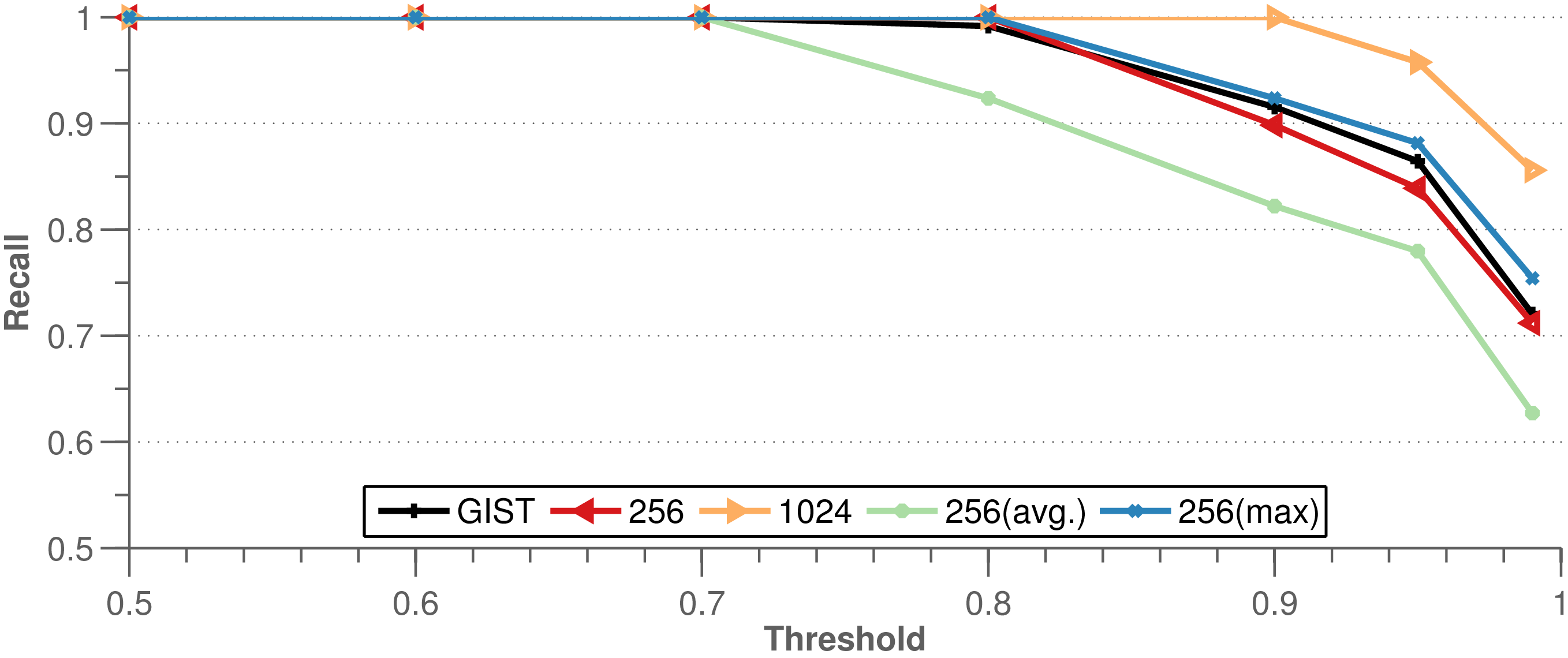} }\\
\subfloat[Sequence 02]{ \includegraphics[width=0.5\textwidth]{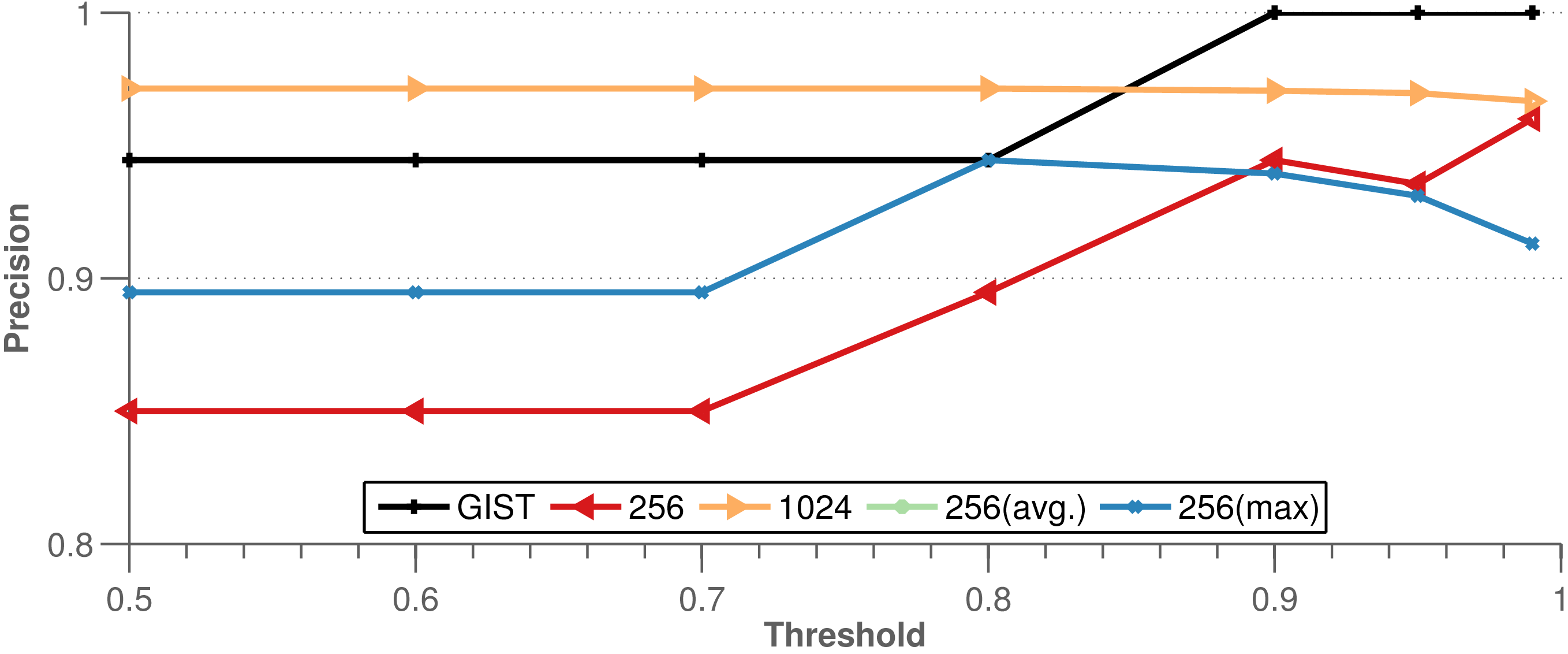} \includegraphics[width=0.5\textwidth]{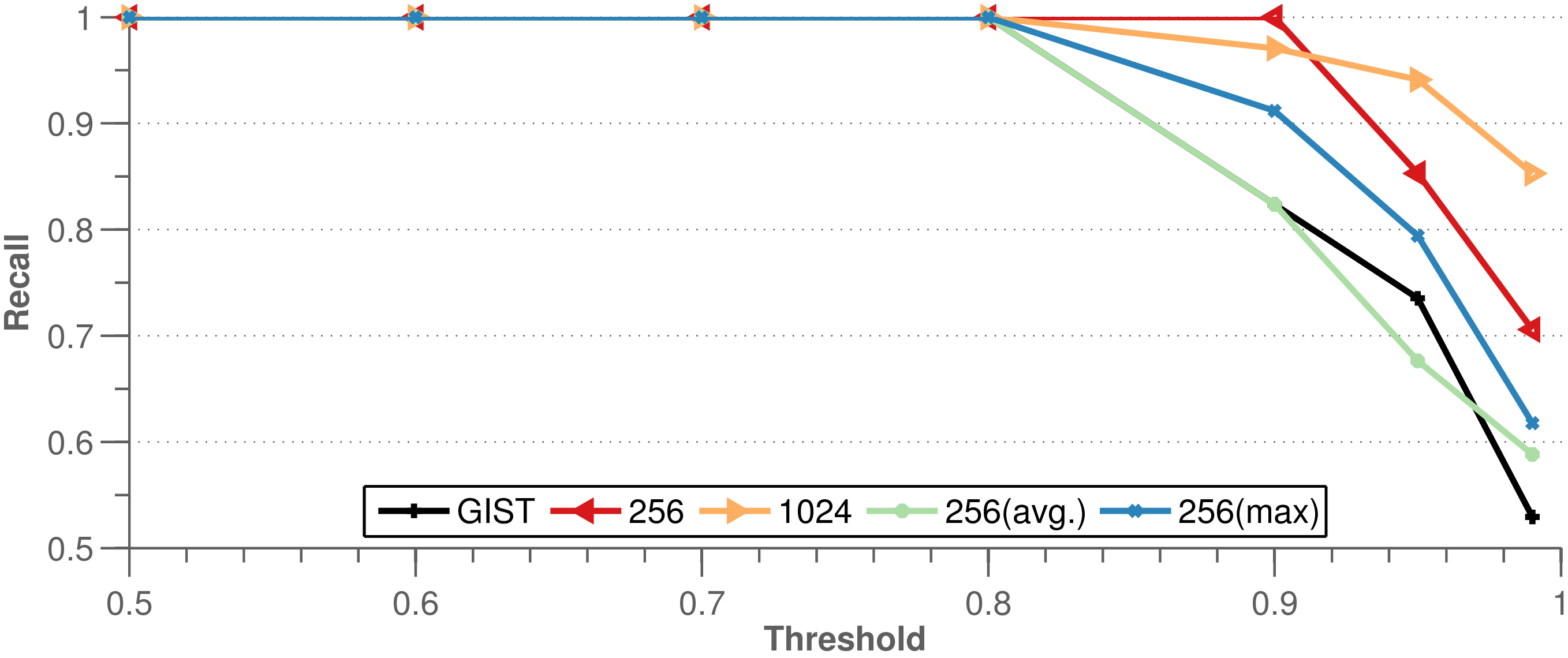} }\\
\subfloat[Sequence 05]{ \includegraphics[width=0.5\textwidth]{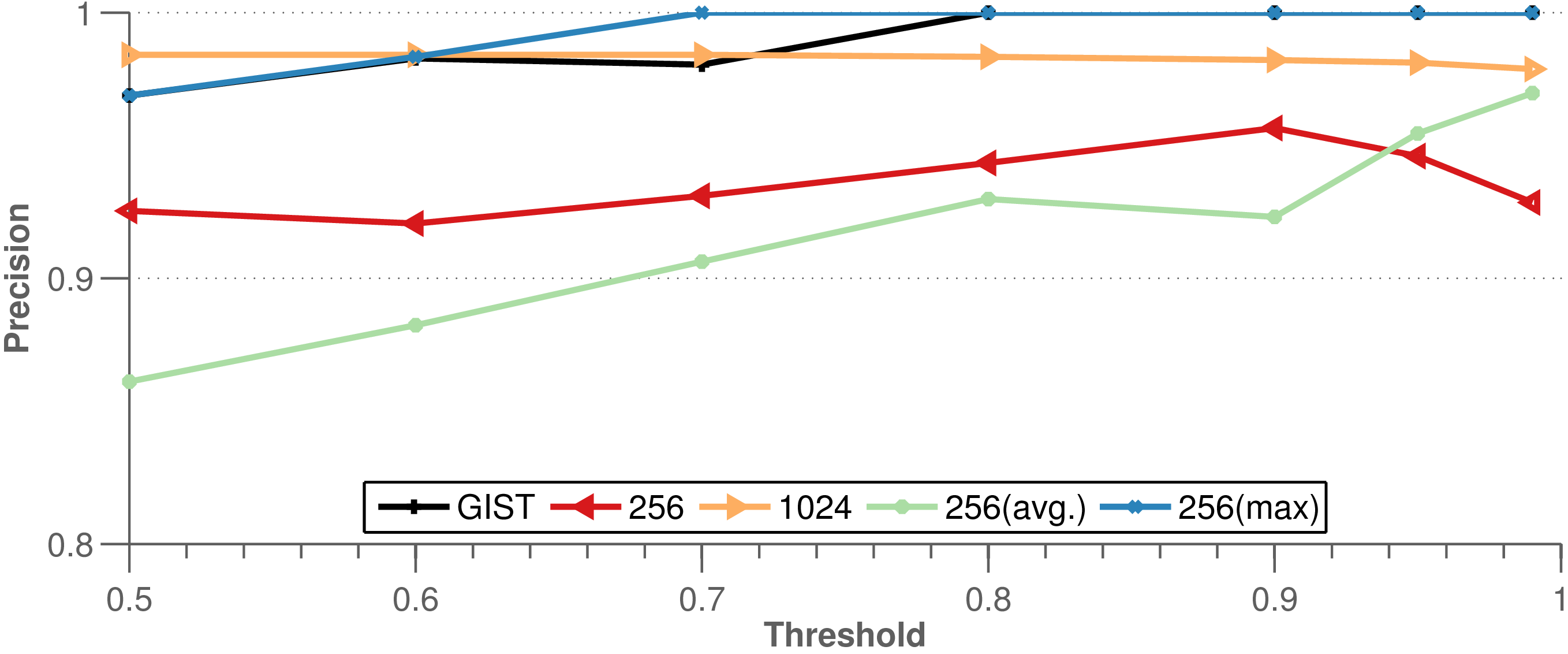} \includegraphics[width=0.5\textwidth]{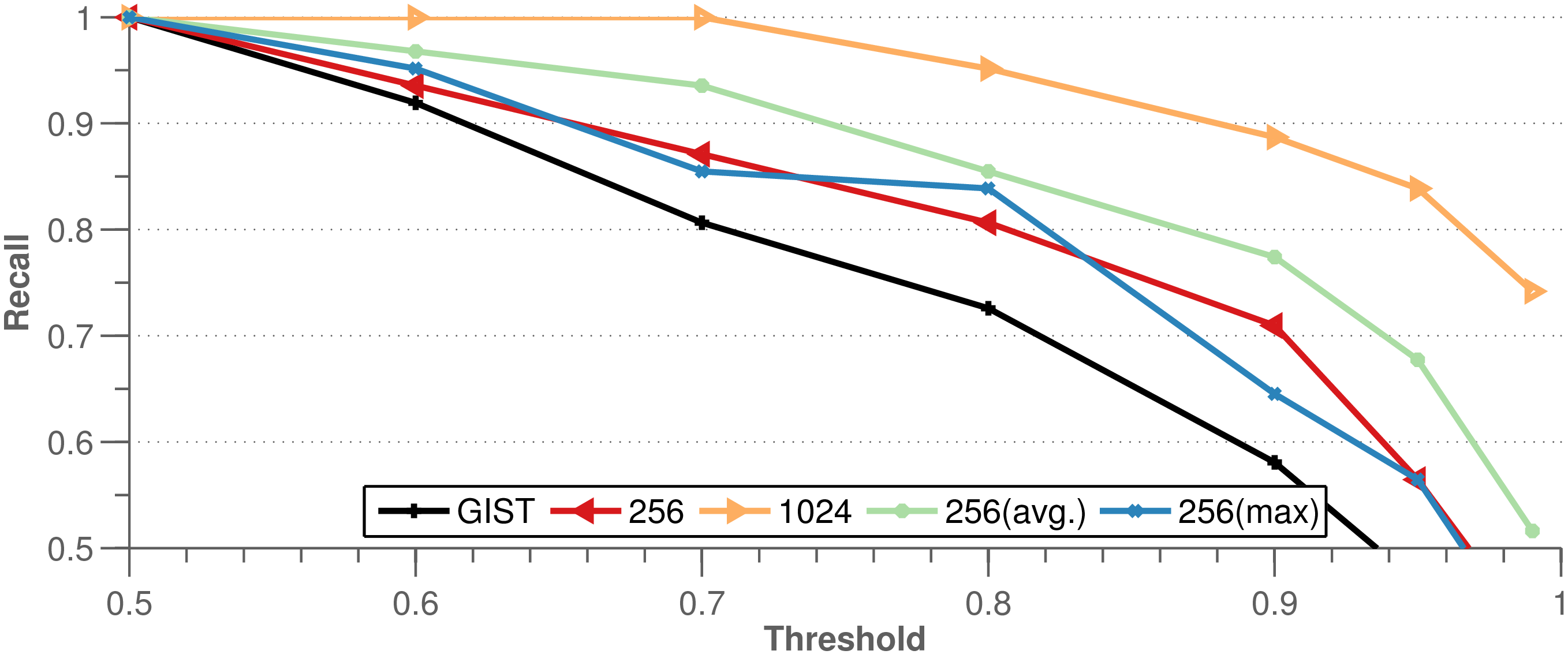} }\\
\subfloat[Sequence 06]{ \includegraphics[width=0.5\textwidth]{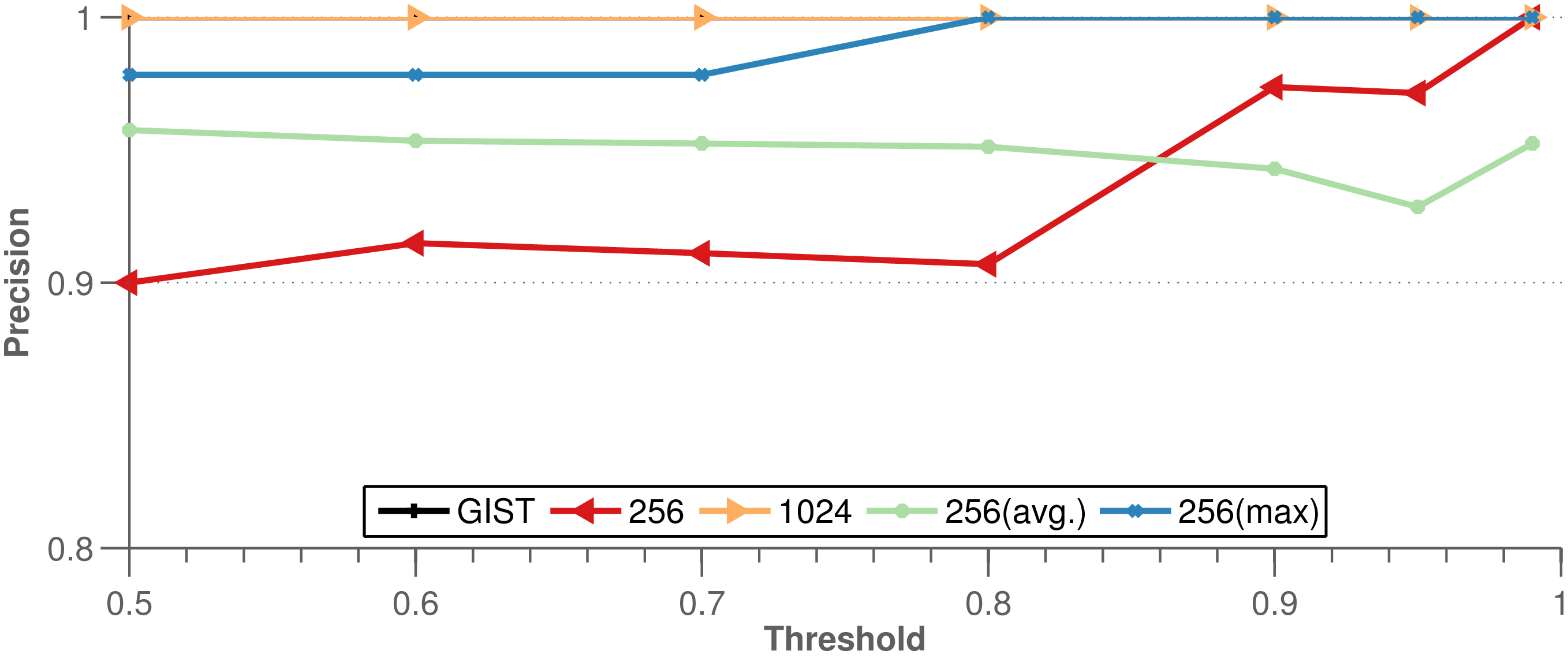} \includegraphics[width=0.5\textwidth]{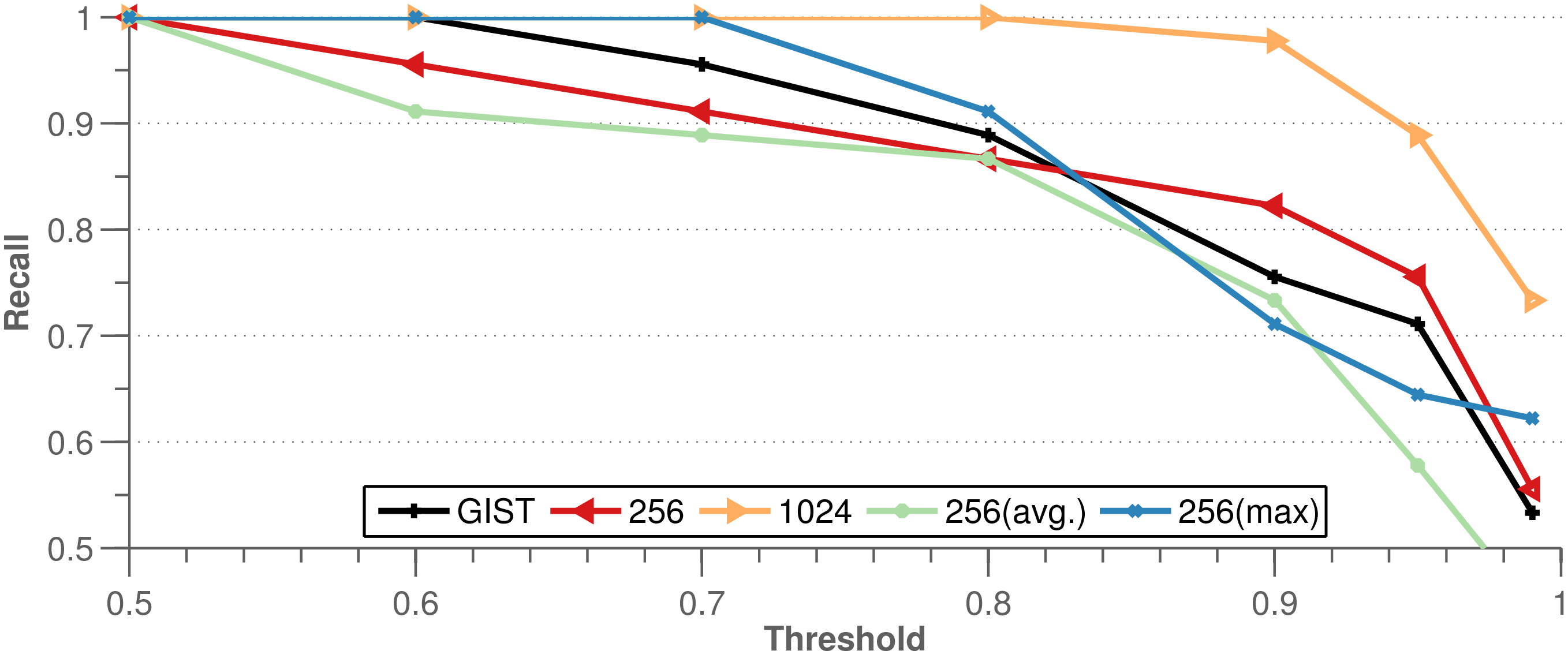} }
\caption{Precision and recall statistics of the proposed loop closing approach using deep features on the KITTI visual odometry benchmark. 
Please note that the Y-axis for the Precision starts at 0.8.}
\label{fig:KITTI}
\end{figure}

In the simplest case, we rescale the input image to $64 \times 64$ giving us a feature in $\mathcal R^{256}$. 
As an alternative strategy, we rescale the image to $128 \times 128$ resulting in four $64 \times 64$ input patches, leading to four vectors in $\mathcal R^{256}$ as output. 
We either stack them to get a vector in $\mathcal R^{1024}$ [1024], average them to get a vector in $\mathcal R^{256}$ [256(avg.)], 
or take the element-wise maximum over them to get a vector in $\mathcal R^{256}$ [256(max)], an operation termed max-pooling. 
We compare the performance of these various configurations derived from deep features with that of GIST for 4 sequences from the KITTI dataset. 
In these experiments, we vary the threshold ($\tau$) to generate different loop closing proposals. The weighing parameter ($\lambda$) is set to 0.5.
The results are presented in \refFig{fig:KITTI}.

It can be seen that the learned features perform better than GIST in all the cases. 
The 1024 configuration is the most discriminative, leading to a smaller number of outliers and greater number of correct loop closures.
Moreover, the performance of 256 is comparable and in most cases better than GIST, which is a 512 dimensional representation. 
This agrees with the insight that learned features have more expressive power compared to hand crafted ones.
Averaging features from the deep network leads to them being less discriminative, leading to the worst performance in Fig. \ref{fig:KITTI}.
It should be pointed out that the original image size for the images in the KITTI sequences is $1241 \times 376$ and the best performance is achieved with a very smaller $128 \times 128 $ representation. Moreover, these experiments also highlight the fact that various representations can be used in the proposed framework for the task of loop closure detection.

\subsection{Multi-Modal Features}
\label{sec:DeepMultiExpr}
A desirable property of the proposed framework is that any unit vector can be used as a feature for carrying out loop closure detection. 
In the previous section, we showed experiments that used single-modal features including GIST and deep features for this purpose. 
However, we are not restricted to using just a single-modal feature. Two distinct features can be stacked to form a new multi-modal feature,
that is, for two feature $\mathbf{f_i} \in \mathcal R^n$ and  $\mathbf{f_j} \in \mathcal R^m $, a new feature $\mathbf{f_{ij}} \in \mathcal R^{(n+m)}$ can be obtained by 
$\mathbf{f_{ij}} = \text{S}(\mathbf{f_i},\mathbf{f_j}) = [\mathbf{f_i}^{T}~\mathbf{f_j}^{T}]^{T} $, followed by projection onto the unit sphere.

\begin{figure}[t!]
\centering
\subfloat[Sequence 00]{ \includegraphics[width=0.5\textwidth]{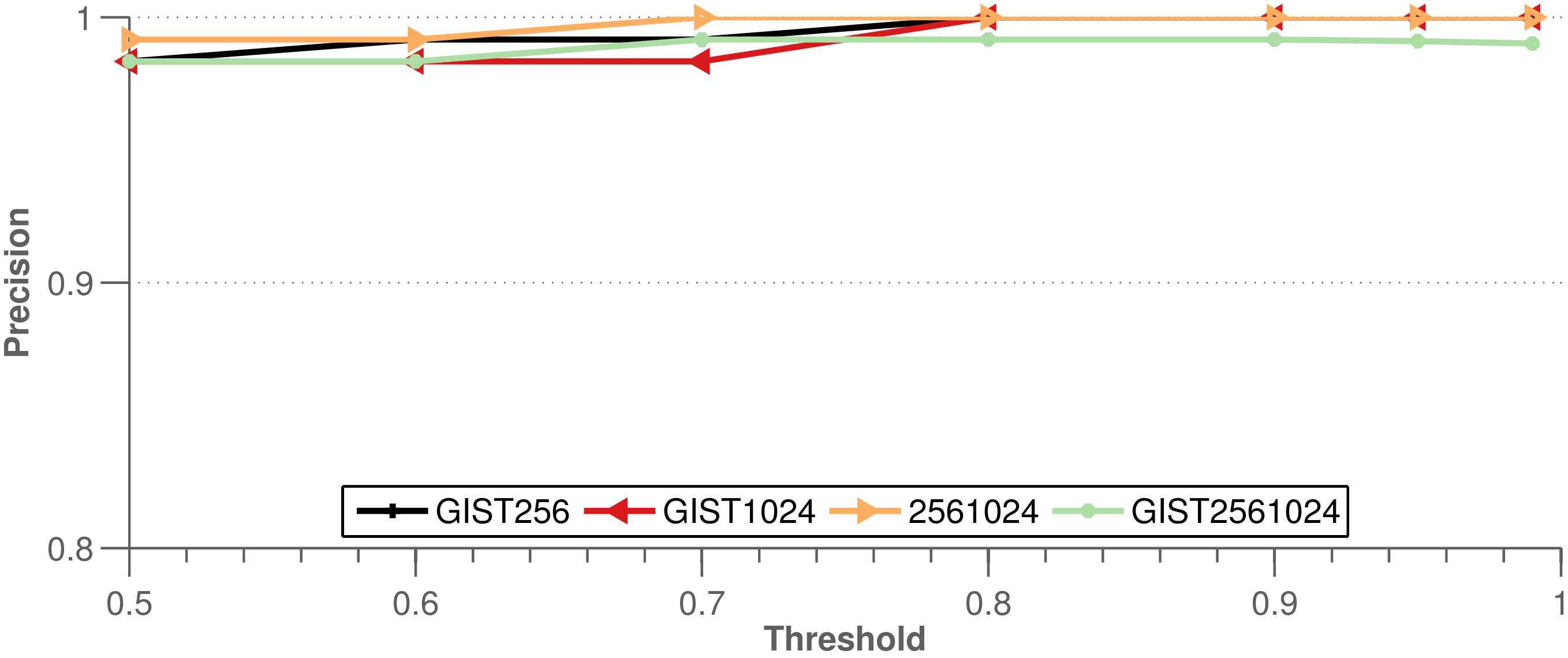} \includegraphics[width=0.5\textwidth]{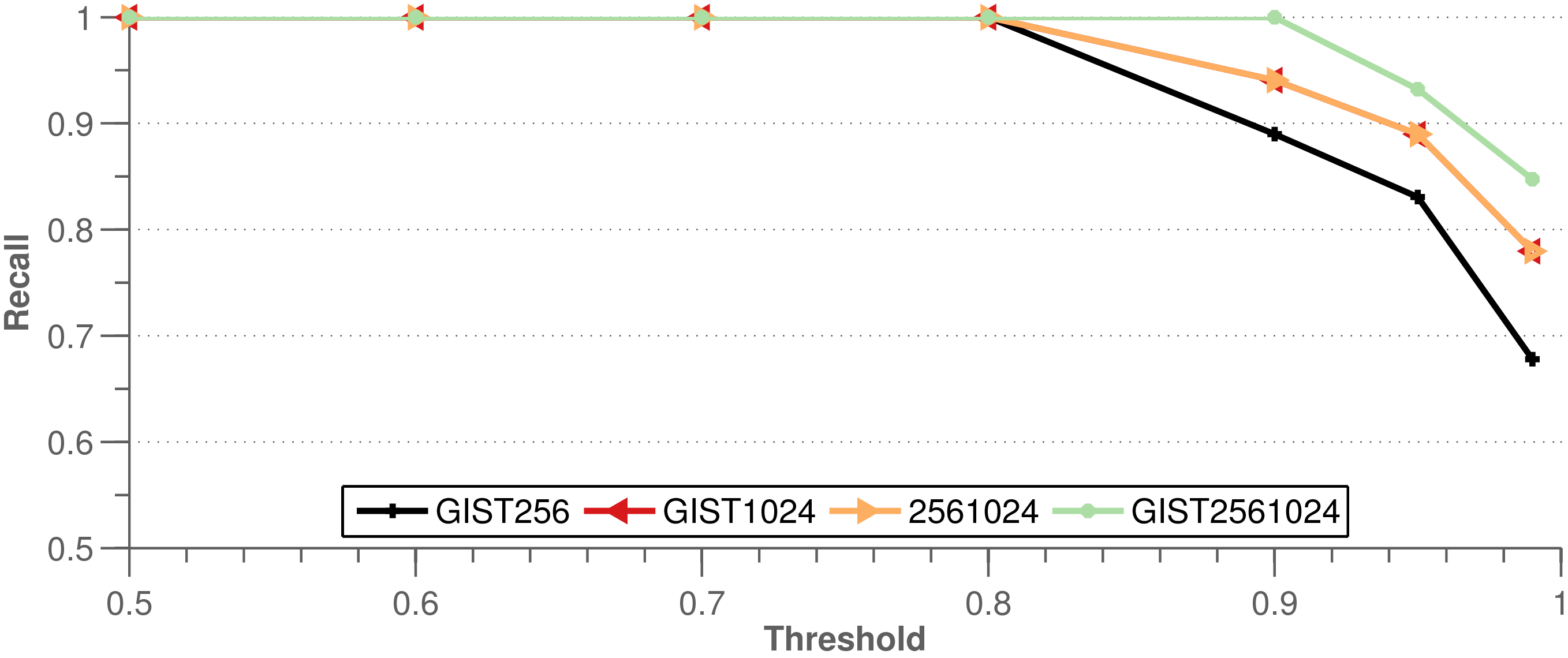} }\\
\subfloat[Sequence 02]{ \includegraphics[width=0.5\textwidth]{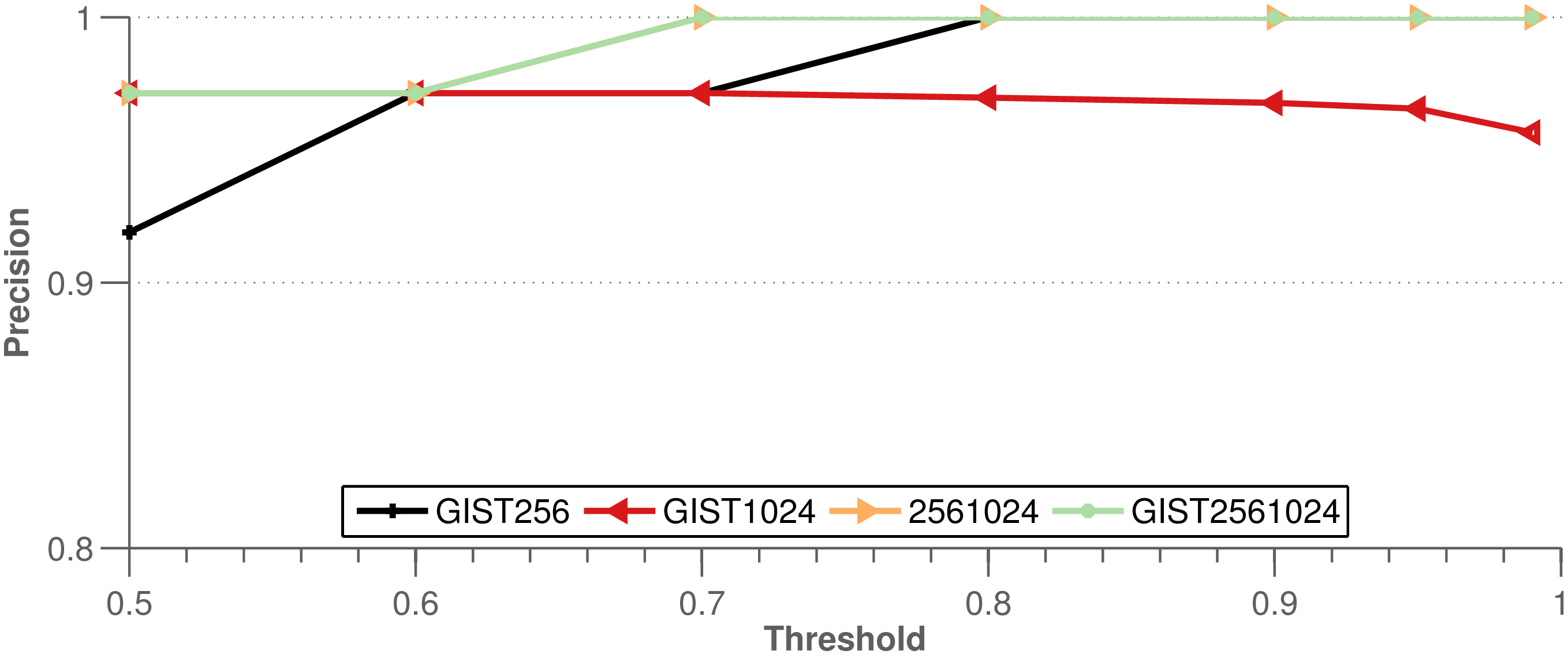} \includegraphics[width=0.5\textwidth]{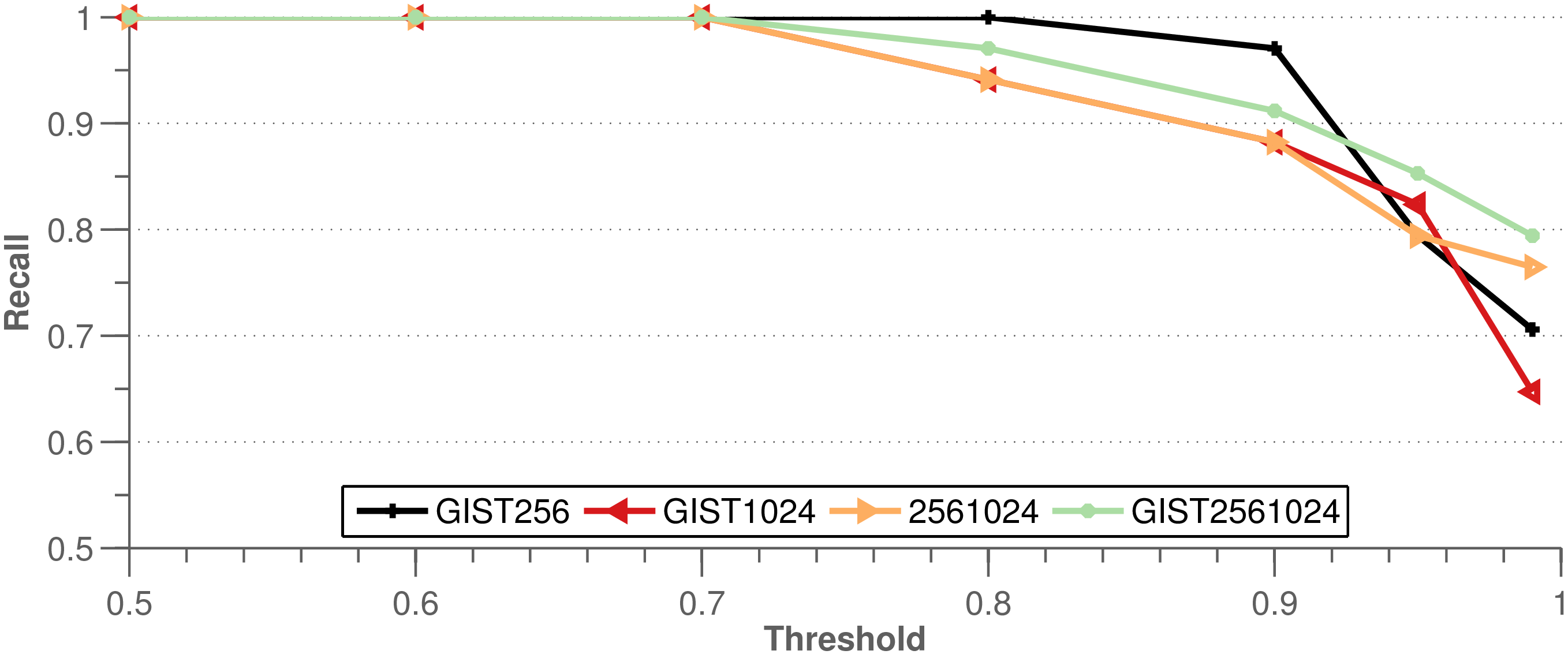} }\\
\subfloat[Sequence 05]{ \includegraphics[width=0.5\textwidth]{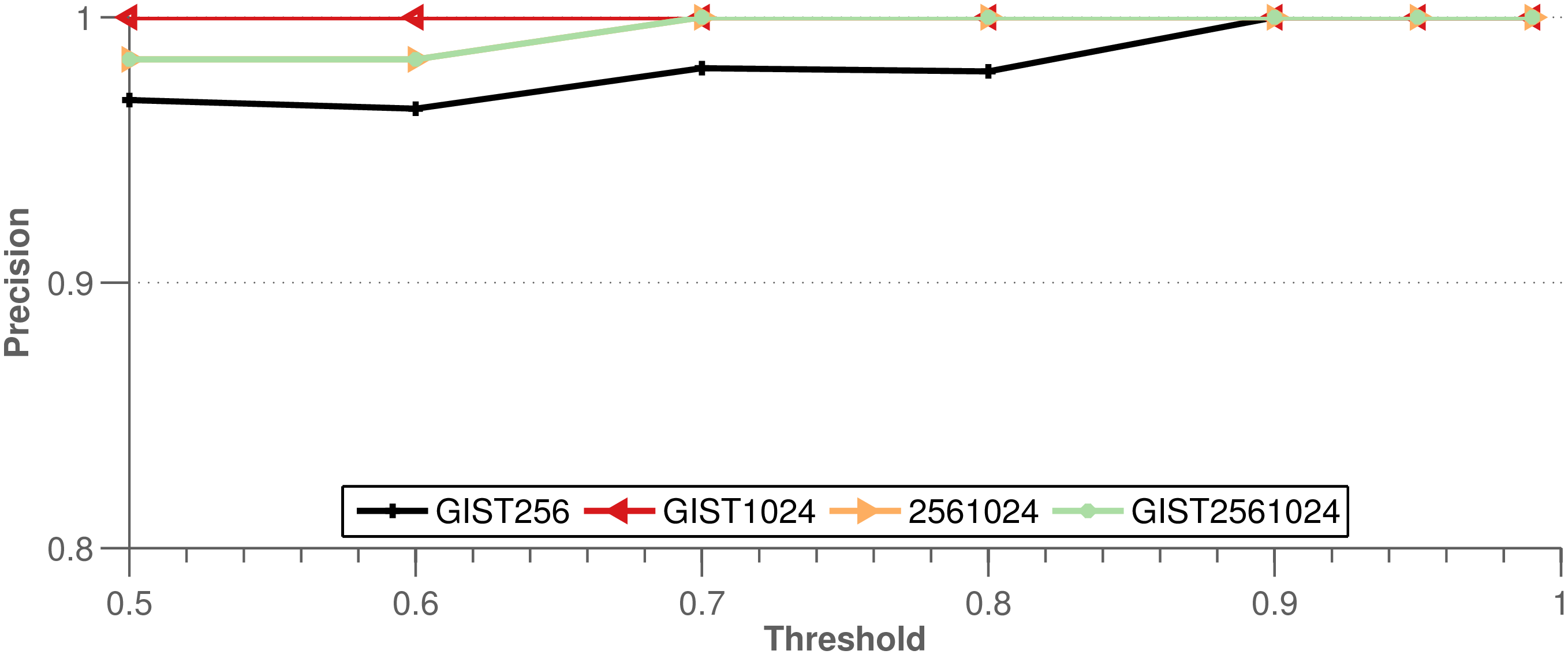} \includegraphics[width=0.5\textwidth]{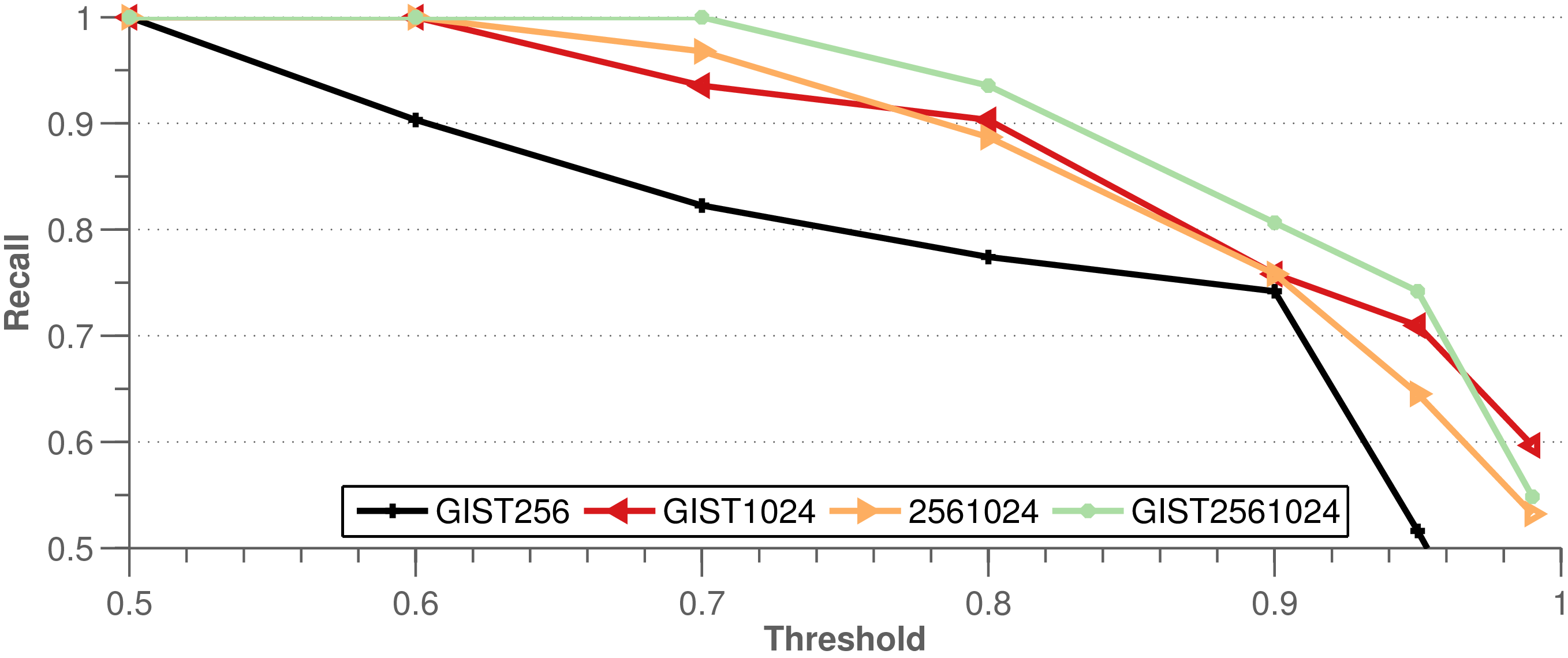} }\\
\subfloat[Sequence 06]{ \includegraphics[width=0.5\textwidth]{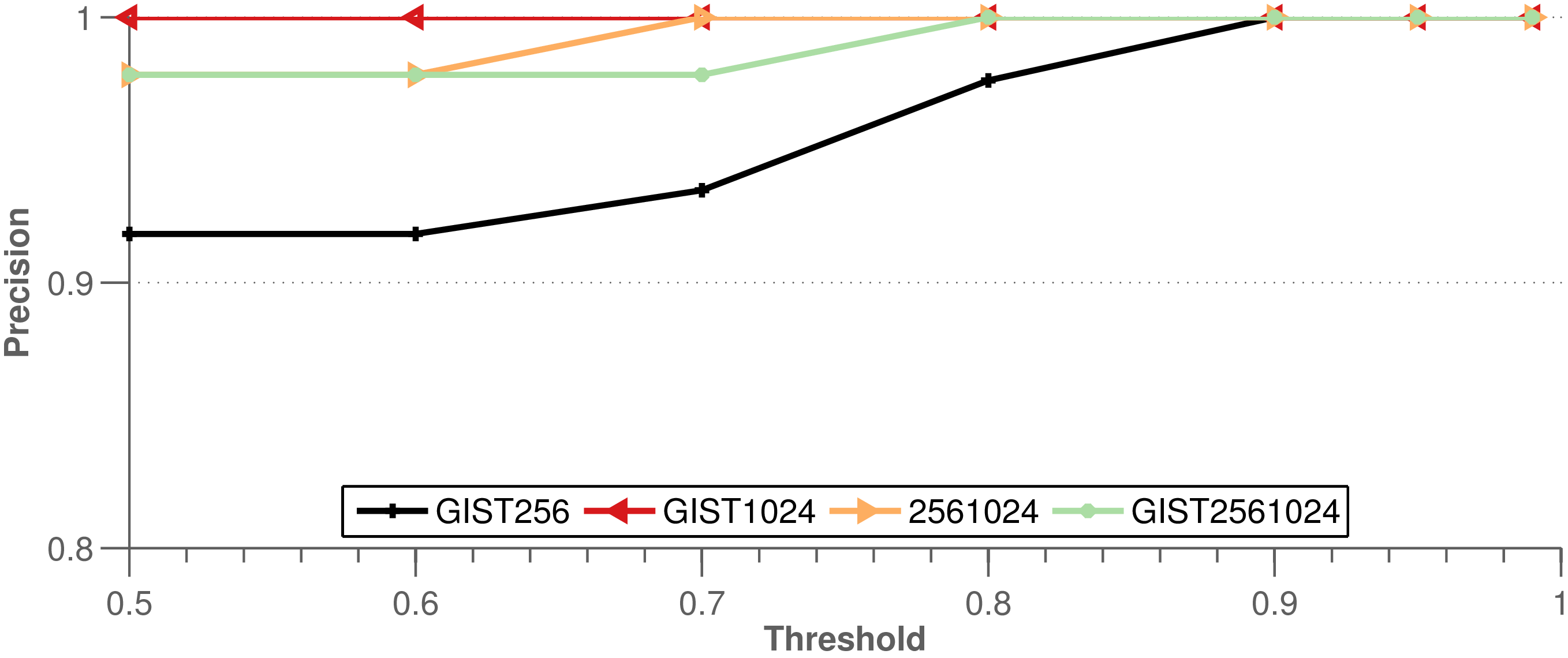} \includegraphics[width=0.5\textwidth]{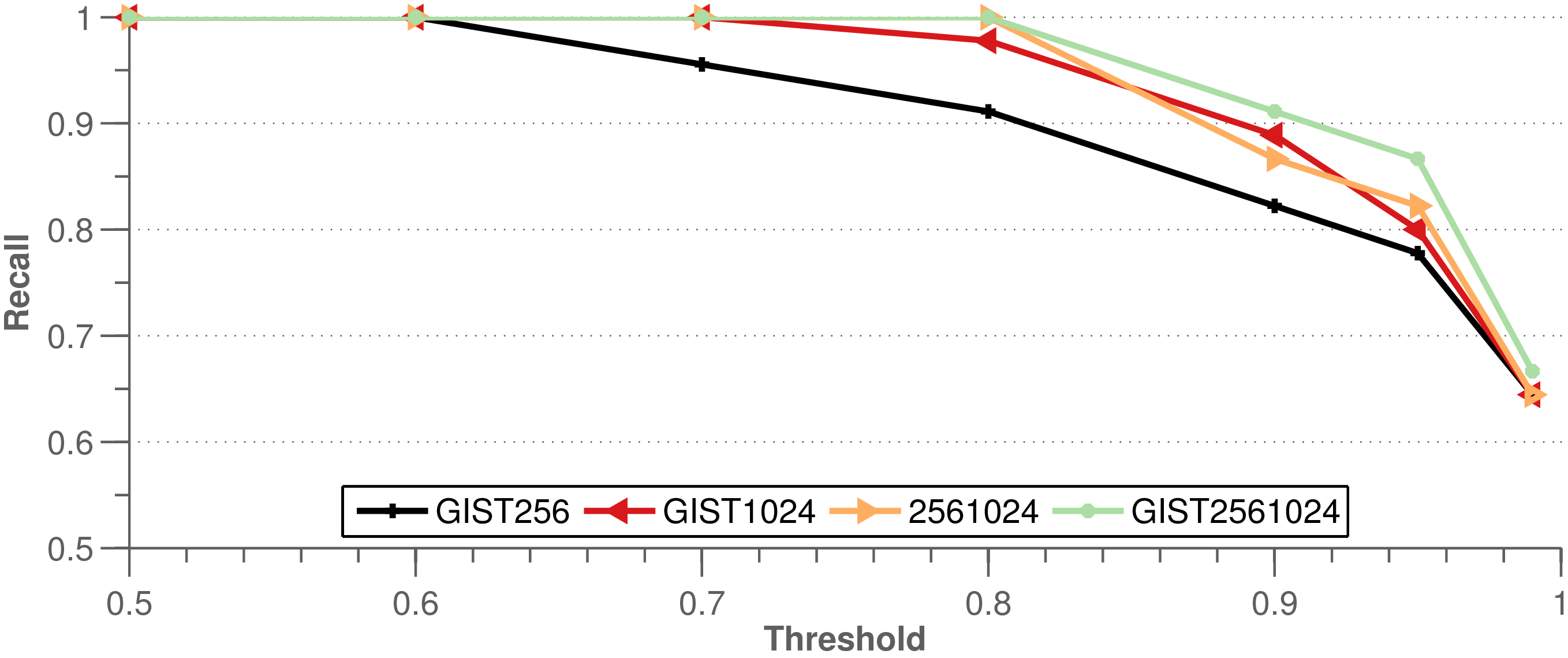} }
\caption{Precision and recall statistics of the proposed loop closing approach with multi-modal features on the KITTI visual odometry benchmark. 
Please note that the Y-axis for the Precision starts at 0.8 to make the details visible.}
\label{fig:KITTI-Multi}
\end{figure}

In order to investigate if this leads to better performance, we use combinations of the GIST and deep features described in Section.~\ref{sec:DeepExpr}.
For the three features: GIST, 256, and 1024, we explore the possible four combinations:
\textbf{1}) S(GIST, 256),
\textbf{2}) S(GIST, 1024),
\textbf{3}) S(256, 1024),
and
\textbf{4}) S(GIST, 256, 1024). 
The results are presented in \refFig{fig:KITTI-Multi}. Comparing it to \refFig{fig:KITTI}, it can be seen that the performance is much better than the single features case, 
the precision is higher with a comparable recall. 

The multi-modal features are more discriminative and can be thought to match images over the intersection of both the descriptor spaces, leading to a better precision. 
This, however, is achieved at the cost of an increased size of the final stacked descriptor. The largest size considered here is that of S(GIST,256,1024) which is $1792$.
However, this is still feasible for runtime operation (see Table \ref{tab:timing}).
This expressive power of the stacked features places images far away from each other in the new combined descriptor space, 
allowing sparser solutions and thus leading to a better recall as well.

\subsection{Repeated Visits}

As highlighted in Section \ref{sec:uniqueness}, the method declares loop closures that are globally unique.
This may lead to missed loops in the worst case, that is, when either two images or the descriptors extracted from them
are exactly the same. 
This is the worst case because in every other case, there would exist a single or a set of images that are able to reconstruct the image. 
Only in the case of the exact same basis vectors, multiple solution with the same value for \refEqu{equ:l1-unconstrained} exist.
\begin{figure}[t!]
\centering
\subfloat[100 images from New College Dataset repeated 60 times]{ \includegraphics[width=0.45\textwidth]{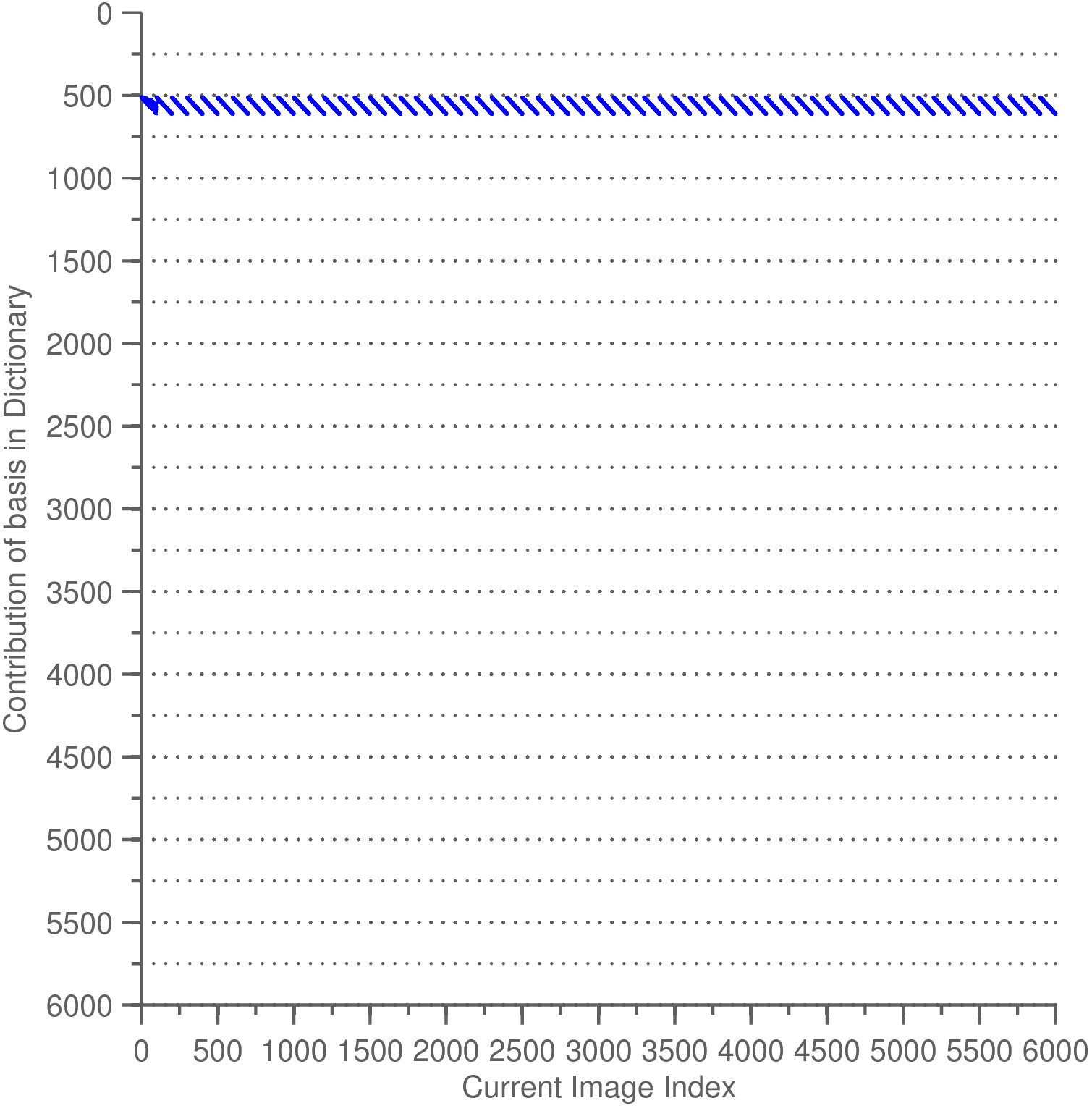} }~~~
\subfloat[Zoomed version of the top-left corner]{ \includegraphics[width=0.45\textwidth]{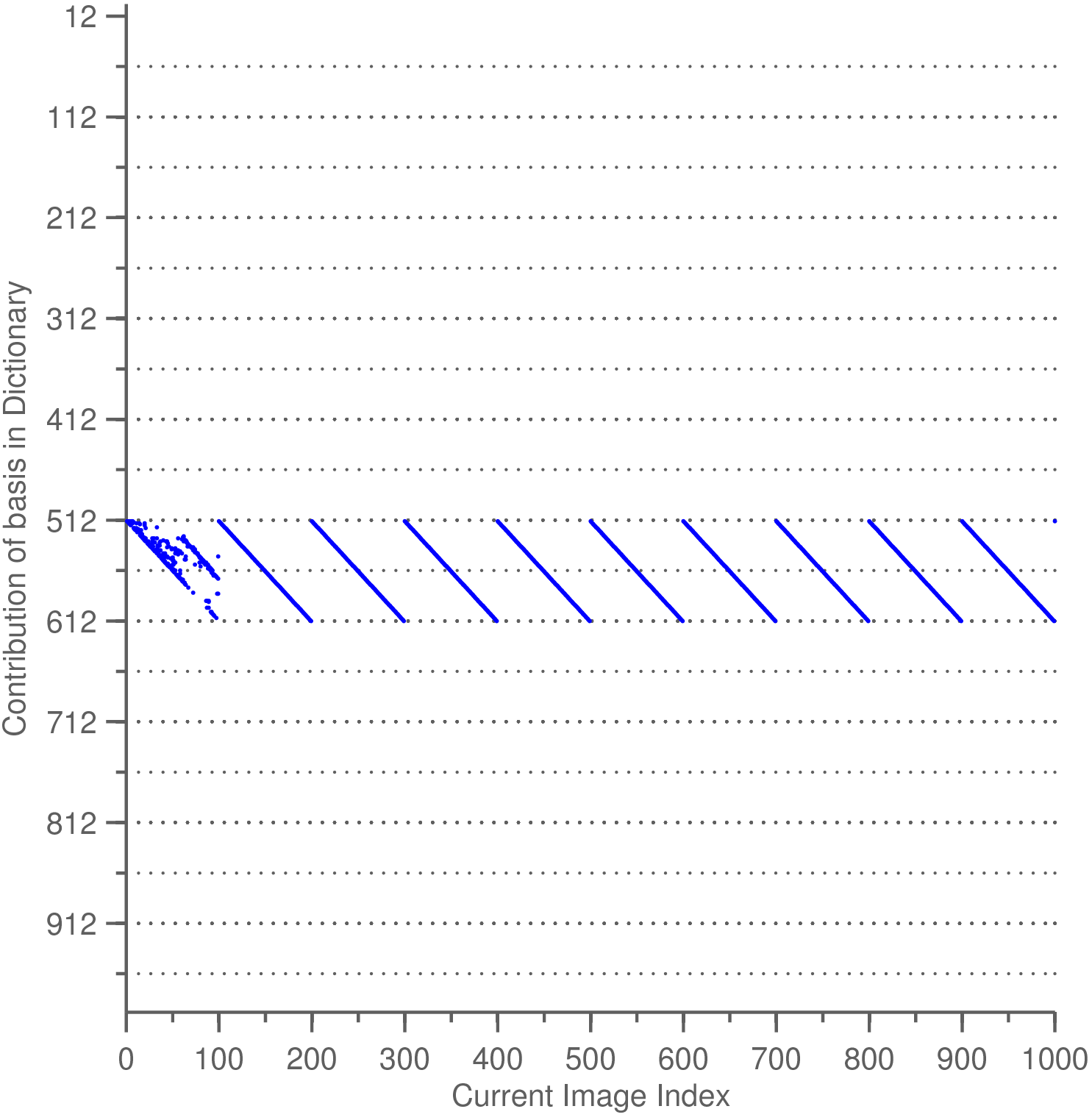} }
\caption{{Revisit scenario using images from the New College Dataset}: GIST descriptor is used as the feature of each image. All the images are presented 60 times to the proposed method for loop closure detection. Plots show the location of non-zero elements returned by the sparse optimization. Diagonal line segments correspond to loop closures.}
\label{fig:RepeatedVisits}
\end{figure}
In order to show how the proposed method behaves in the worst case, 
we take a batch of 100 images from the New College dataset and present the batch 60 times to the loop closing method.
This simulates the situation of 60 repeated visit to the same place leading to the generation of the exact same images. 
Each image is described with a GIST descriptor of length 512.
We use the proposed framework to solve for loop closures and look at the sparse solution [$\isVec{\alpha}_i $ in \refEqu{equ:l1-unconstrained}] for each of the images presented to the method. We stack these $\isVec{\alpha}_i $ as column vectors and the results are shown in \refFig{fig:RepeatedVisits}.

The method is able to correctly associate each loop closure to one of the first 100 images in the dictionary (initial 512 entries correspond to noise bases). At each revisit, we can see diagonal lines associating the current image to one of the corresponding first 100 images. \refFig{fig:RepeatedVisits}(b) shows a zoomed-in version, in which the initial noisy reconstruction during the first 100 images can also been seen, since at that time there are no valid loop closures present in the dictionary.
It can be clearly seen from \refFig{fig:RepeatedVisits} that even in the case of 60 revisit, 
the proposed method is able to associate the current image to the first occurrence of the bases in the dictionary. 
This can be attributed to the greedy nature of the  $\ell_1$-optimizer incorporated in the framework. 
It chooses the first basis that it can find which has the least reconstruction error according to \refEqu{equ:l1-unconstrained}, which in this case corresponds to the first occurrence of the basis in the dictionary. 
The existence of exact basis leads to minimization of the reconstruction error at the first step, hence returning the first occurrence of the corresponding basis.

\blue{

\subsection{Severe Illumination Changes} \label{sec:Norland}

\begin{figure}
\centering
\includegraphics[width=0.4\textwidth]{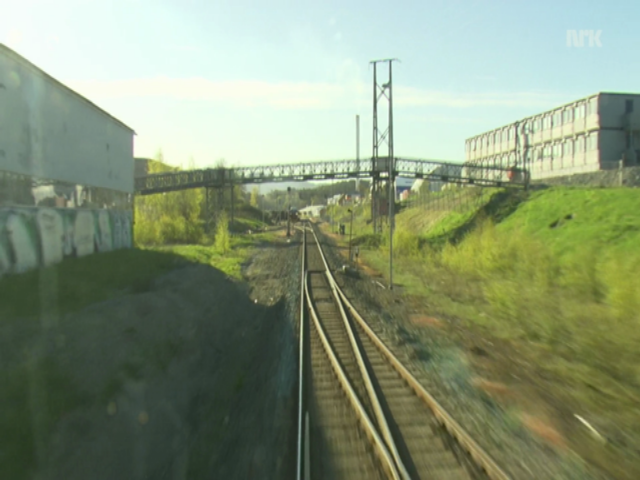}~~~~
\includegraphics[width=0.4\textwidth]{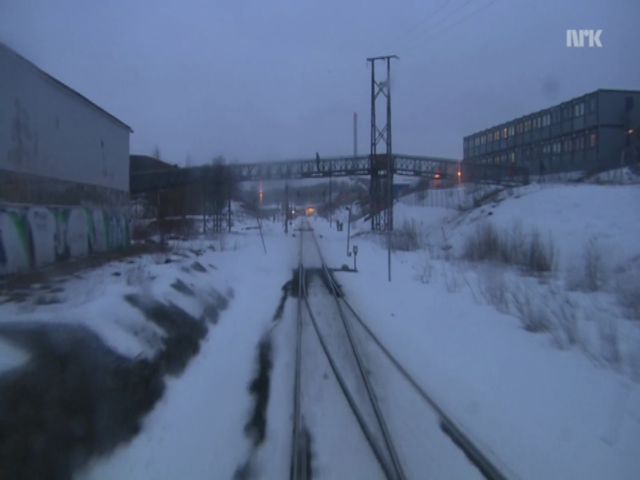}
\caption{Example images from the VPRiCE challenge dataset where images have been acquired over different seasons.}
\label{fig:VPRiceExample}
\end{figure}

One of the challenges that makes place recognition difficult is illumination variation arising during long-term operation such as transitions from day to night or between different seasons.
In order to test the performance of the proposed method under such severe illumination changes,
we use the data from the {Visual Place Recognition in Challenging Environments} (VPRiCE) challenge \cite{vpriceChallenge}. 
The dataset consists of 7778 images from a variety of outdoor environments and under various viewing conditions. The dataset provides both \textit{memory} and \textit{live} images, and the objective is to find a match for each live image in the memory images. 
The first part of the dataset contains images acquired from a camera on-board a train, recorded in spring and winter for the same trajectory of the train, an example of which is shown in \refFig{fig:VPRiceExample}. 
In this test, we use only the images from the train sequences (2289 in memory and 2485 in live).
The following experiments aim to investigate two aspects of the loop-closure problem:
(i) the performance of the proposed method in challenging condition against a baseline of nearest-neighbor (NN)  with exhaustive search,
and (ii) the effect of the $\lambda$ parameter on the sparsity of the solution.

\begin{figure*}[t!]
\centering
\subfloat[Representation : Scaled by $1/8$]
{ 
\includegraphics[width=0.45\textwidth]{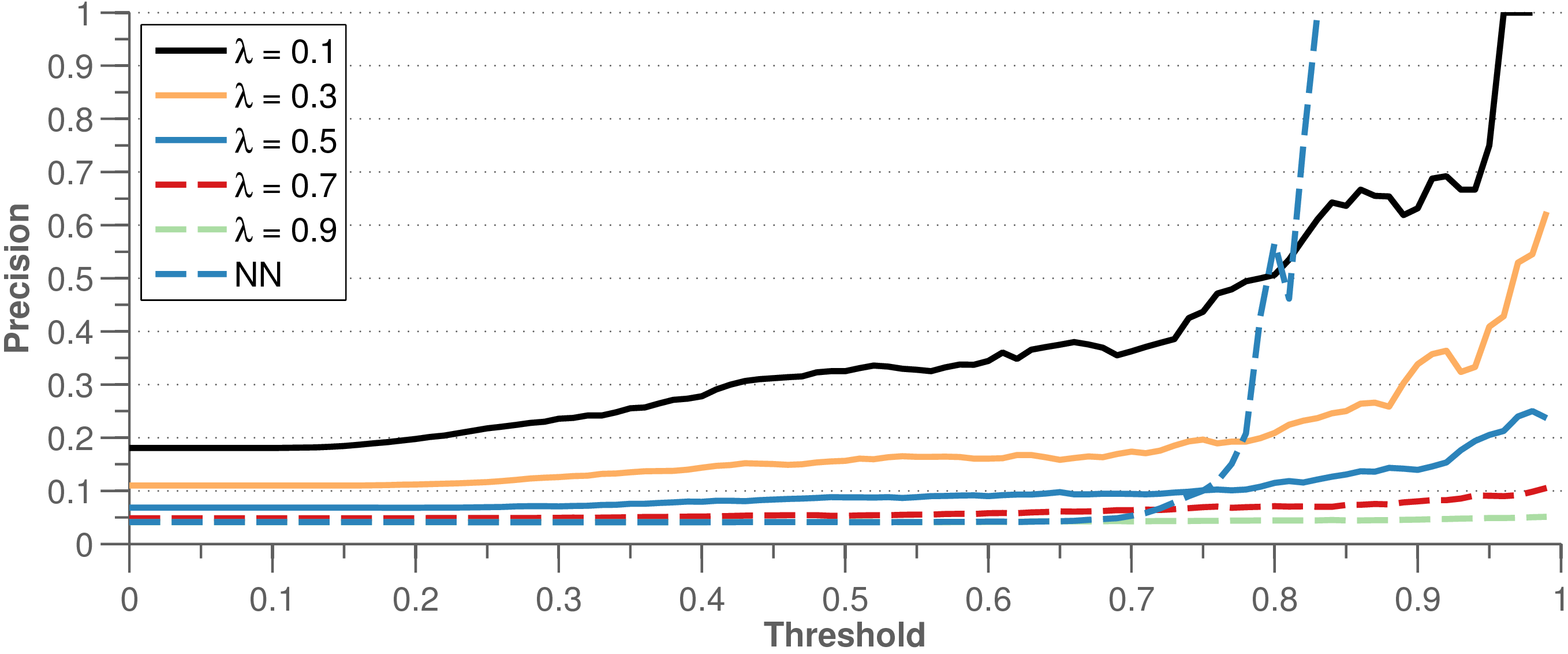} 
\includegraphics[width=0.45\textwidth]{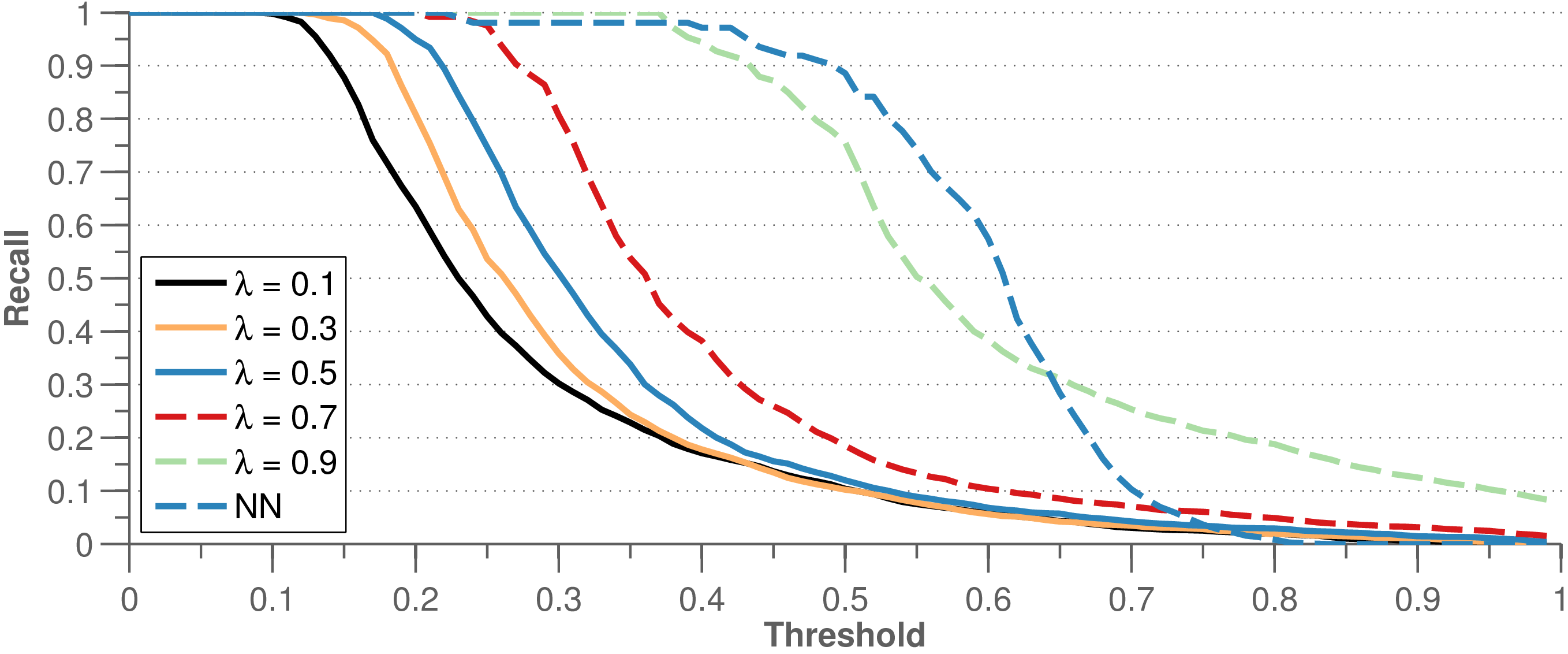} 
}\\
\subfloat[Representation : Scaled by $1/16$]{ 
\includegraphics[width=0.45\textwidth]{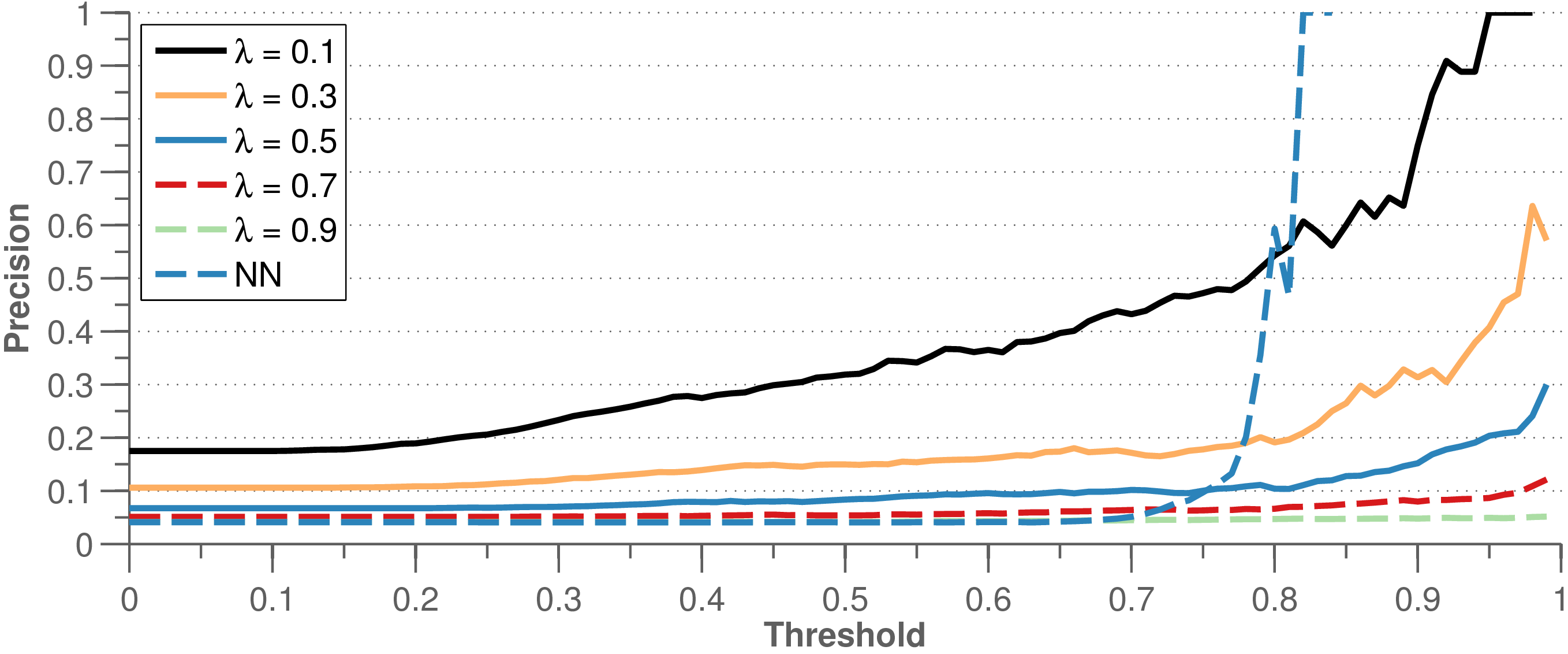} 
\includegraphics[width=0.45\textwidth]{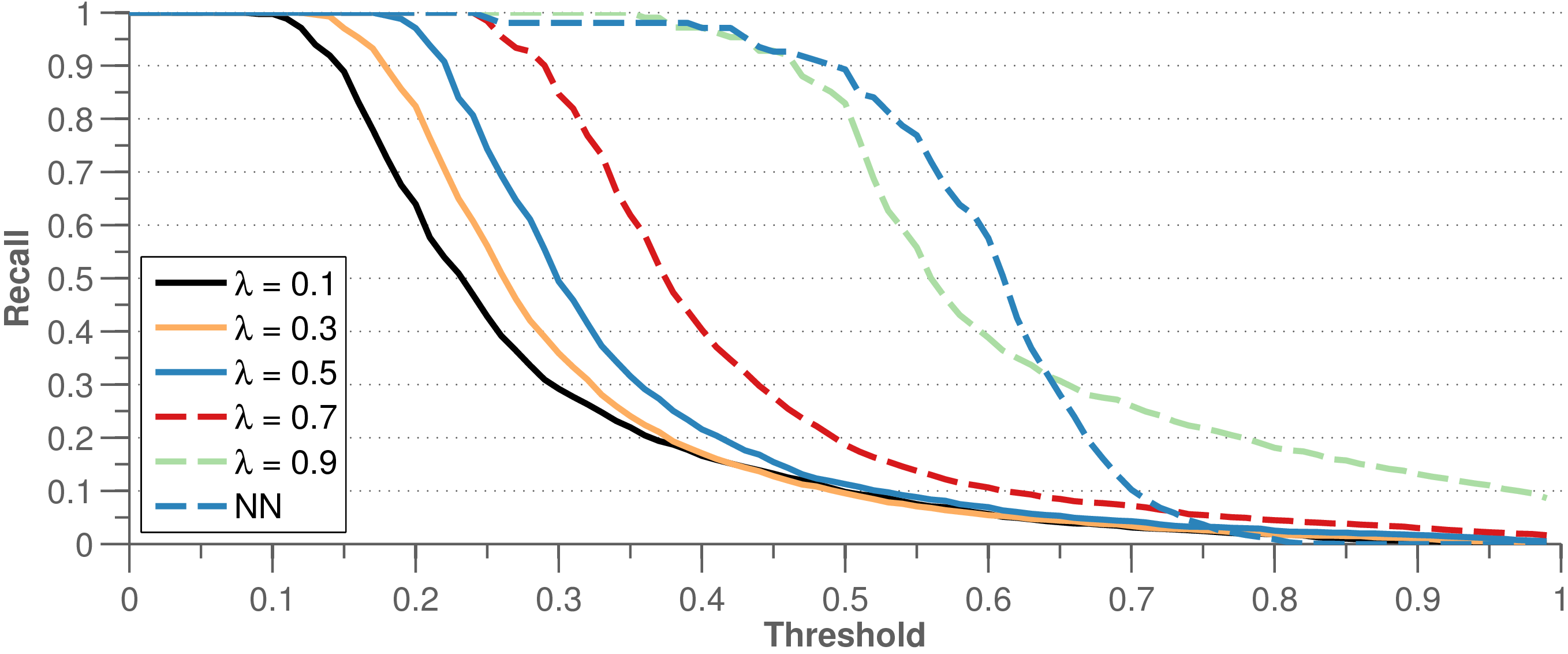} 
}\\
\subfloat[Representation : Scaled by $1/32$]{ 
\includegraphics[width=0.45\textwidth]{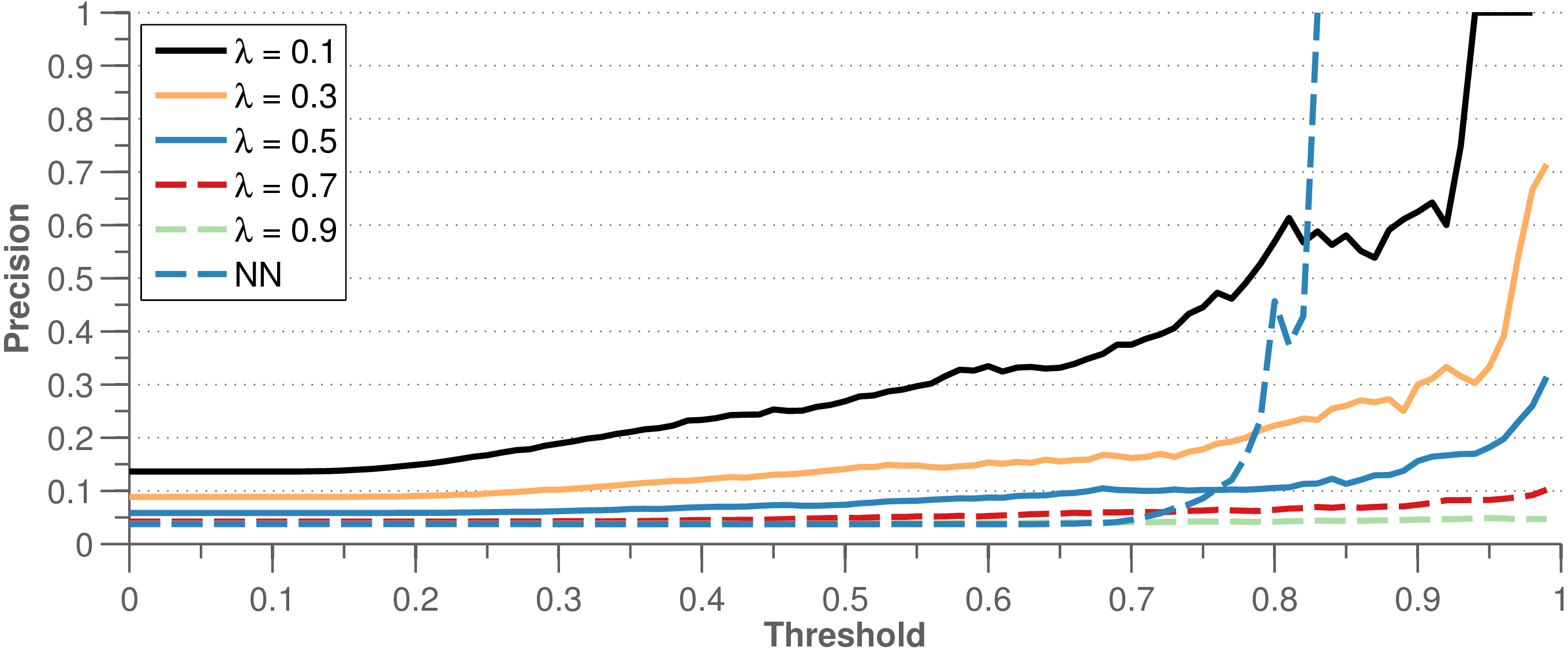} 
\includegraphics[width=0.45\textwidth]{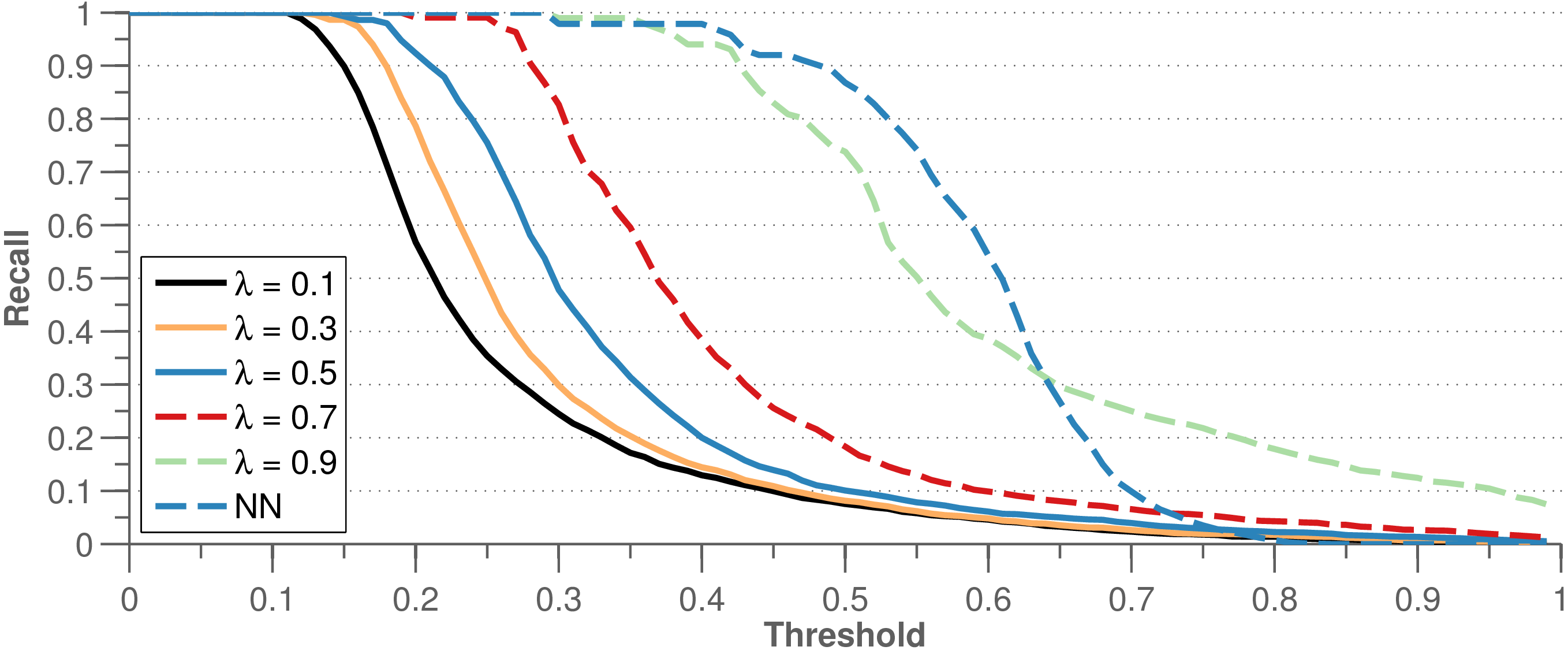} }\\
\subfloat[Representation : GIST]{ 
\includegraphics[width=0.45\textwidth]{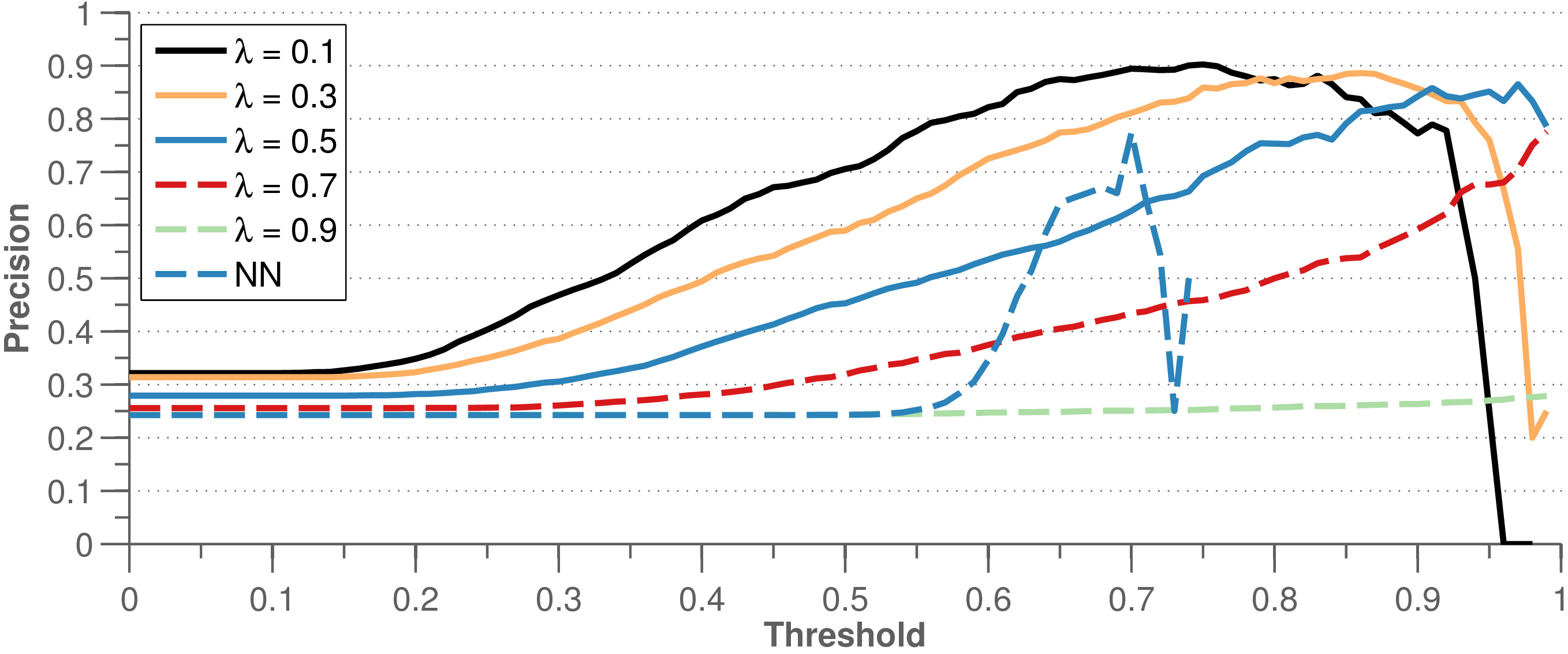} 
\includegraphics[width=0.45\textwidth]{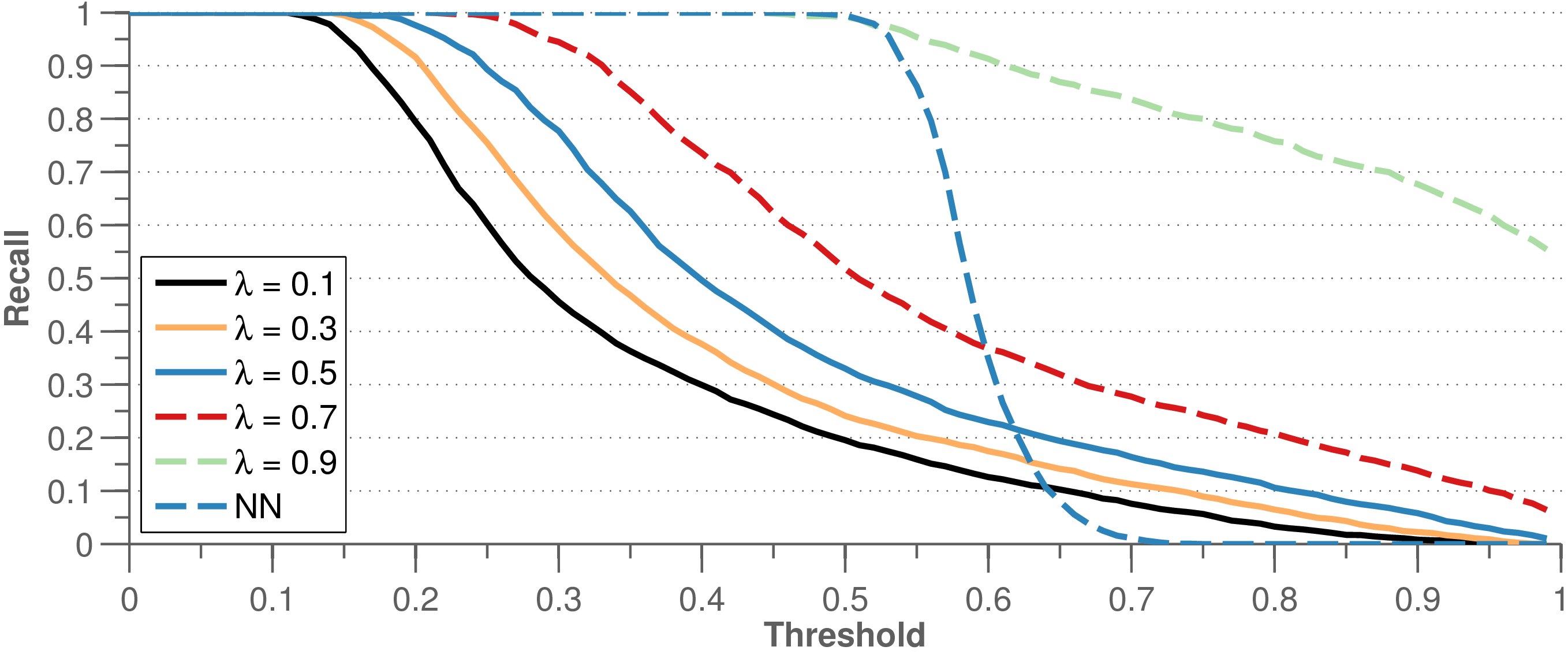} }
\caption{{Performance Comparison on Norland Dataset}: 
For NN, the x-axis corresponds to $1-$distance.
Varying the explanation penalty ($\lambda$) leads to better loop closure detection, but at the same time, the trade-off between precision and recall can be seen from the two plots. Better precision occurs are lower values of $\lambda$ which has the effect of lower recall and vice versa.
}
\label{fig:NorlandResults}
\end{figure*}

To that end, 
we use two representations as mentioned earlier: the down-sampling raw images and the GIST descriptors.
Each frame in \textit{memory} is used to populate the dictionary and then a match is search for each \textit{live} image.  
Note that no temporal or geometric consistency is applied and all decisions are taken just on the current image.
Similarly, for each live image, we find the nearest neighbour match using an {\em exhaustive} search over {\em all} the memory images given the representation.
The results are shown in \refFig{fig:NorlandResults}.
It can be seen that the proposed method provides better precision and recall in almost all the cases, and especially,
the recall degrades more gracefully compared to the NN method.
As opposed to the previous results, the threshold in \refFig{fig:NorlandResults} varies from $0$ to $1$ so that the results from the NN can be shown as well.
\refFig{fig:NorlandResults} also shows that a smaller $\lambda$ (a less sparse solution) leads to a higher precision but lower recall and vice versa,
which agrees with the results in \refFig{fig:varying-tau}.

\begin{figure}
\centering
\includegraphics[width=0.7\textwidth]{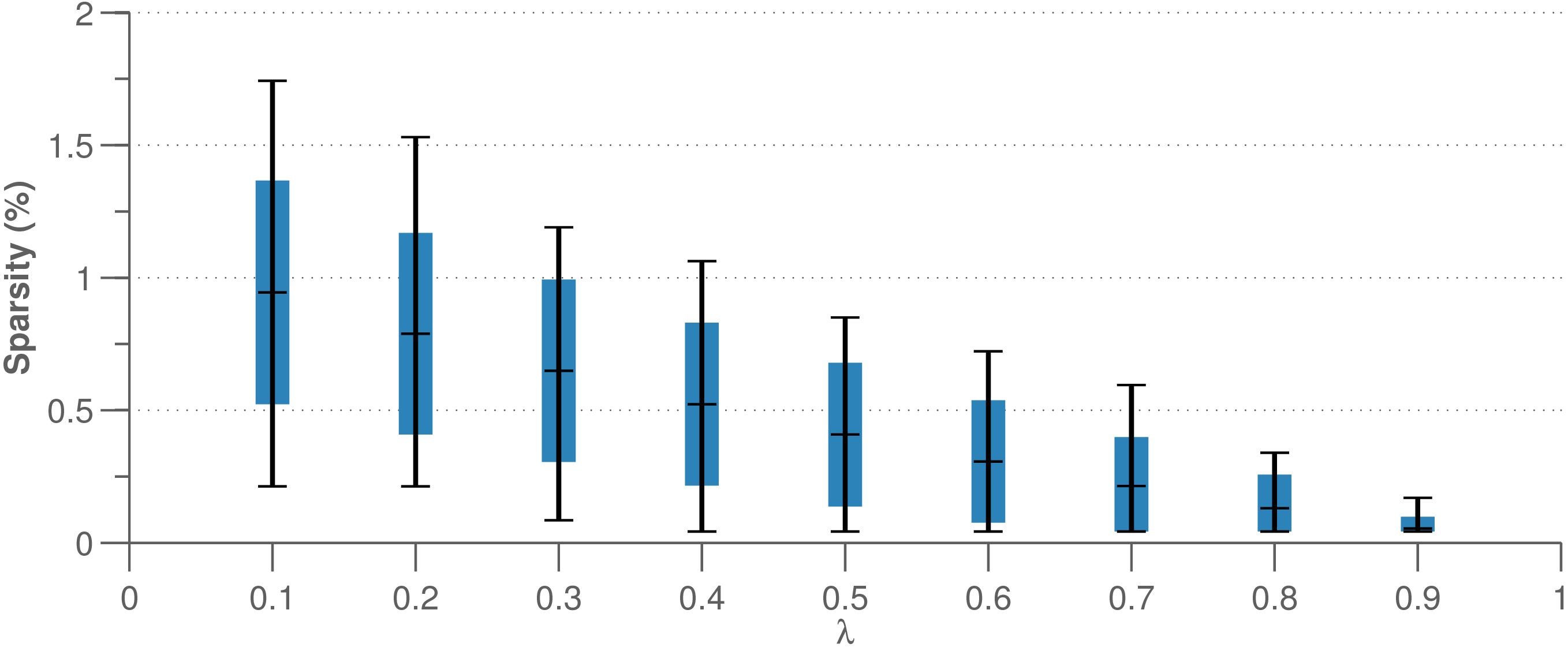}
\caption{Analysis of the sparsity relationship with varying $\lambda$. Black vertical bars represent max, mean and min for each value of $\lambda$,
while the blue bar spans mean with two standard deviations. The dictionary contains over 2000 vectors for this dataset. }
\label{fig:sparsityNNZ}
\end{figure}

Another interesting aspect of the problem is the change in sparsity by varying $\lambda$.
We report percentage of the number of nonzeros (NNZ) for different values of $\lambda$ in \refFig{fig:sparsityNNZ} for the experiments presented in \refFig{fig:NorlandResults}.
As the value of $\lambda$ increases, the solution become more sparse but even for smaller values the sparsity is a very small percentage of the number of vectors in the dictionary.
The NN-based method, on the other hands, has all non-zeros entries corresponding to the number of vectors in the dictionary. These results imply that a sparse solution can be constructed even for  small values of $\lambda$ and the  solutions is still much sparser as compared to the NN solution. 
}

% END \input{experiments.tex}
% BEGIN \input{conclusions.tex}
%!TEX root = ras_2015.tex
\section{Conclusions and Future Work} \label{sec:conclusions}

While the problem of loop closure has been well studied in visual navigation, 
motivated by the sparse nature of the problem 
(i.e., only a small subset of past images actually close the loop with the current image),
in this work, we have for the first time ever posed it as a sparse convex $\ell_1$-minimization problem.
The {\em globally optimal} solution to the formulated convex problem, by construction, is {\em sparse}, 
thus allowing efficient generation of loop-closing hypotheses.
Furthermore, the proposed formulation enjoys a {\em flexible} representation of the basis used in the dictionary,
with {\em no} restriction on how the images should be represented (e.g., what descriptors to use).
Provided any type of image vectors that can be quantified with some metric to measure the similarity, 
the proposed formulation can be used for loop closing.
Extensive experimental results have validated the effectiveness and efficiency of the proposed algorithm,
using either the whole raw images as the simplest possible representation
or the high-dimensional descriptors extracted from the entire images including feature from deep neural networks.
We have also shown empirically how the design parameters effect the performance of our method,
and in general, the proposed approach is able to efficiently detect loop closing for real-time applications.
The quality of loop closure depends on the type of descriptor employed for the task. Raw images do not provide view-point or illumination invariance.
For detecting loop closures in drastically different illumination conditions such as day and night, 
the problem is reduced to finding a suitable descriptor and then the proposed framework can be employed. %for the task of loop closure detection.

%As for the future work,
We currently use a single threshold~$\tau$ to control the loop-closure hypotheses,
which guarantees a globally unique hypothesis.
However, in the case of multiple revisits to the same location, this hard thresholding would prevent detecting 
any loop closures and the revisits would be simply considered as perceptual aliasing,
which is conservative but loses information. 
In the future, we will investigate different ways to address this issue.
For example, as mentioned earlier, we can sum up the contributions of basis vectors 
if a loop has already been detected between them and thus ensure that multiple visits lead to 
more robust detection of loop closures. 
Nevertheless, this has not been a major issue in our tests;
as shown in \refFig{fig:Newcollege} and \refFig{fig:RepeatedVisits}, %for the New College dataset, 
the proposed algorithm is capable of detecting loops at different revisits, even in the worst case scenario.
%
%%%
As briefly mentioned before, 
the number of basis vectors in the dictionary grows continuously and 
can prohibit the real-time performance for large-scale problems.
To mitigate this issue, one possible way would be to update the dictionary dynamically by checking a novelty factor  
in terms of how well the current image can be explained by the existing dictionary,
which is akin to adding ``key frames'' in visual SLAM. 
%
%Alternatively, the linear growth of the size of the dictionary may be prevented by building multiple fixed-size dictionaries,
%in which case, however, the $\ell_1$-minimization needs to be solved for each such dictionary separately, 
%leading to a higher (but feasible) computational cost.

% END \input{conclusions.tex}

%!TEX root = ras_2015.tex
\section*{Acknowledgments}
This work was partially supported by the MINECO-FEDER project DPI2015-68905-P, by the research grant BES-2010-033116, by the travel grant EEBB-I-13-07010,
by the ONR grants N00014-10-1-0936, N00014-11-1-0688 and N00014-13-1-0588, by the NSF awards IIS-1318392 and IIS-15661293, and by the DTRA award HDTRA 1-16-1-0039.
\vspace{2mm}
%% Use plainnat to work nicely with natbib. 
\bibliographystyle{plainnat}

\begin{thebibliography}{46}
\providecommand{\natexlab}[1]{#1}
\providecommand{\url}[1]{\texttt{#1}}
\expandafter\ifx\csname urlstyle\endcsname\relax
  \providecommand{\doi}[1]{doi: #1}\else
  \providecommand{\doi}{doi: \begingroup \urlstyle{rm}\Url}\fi

\bibitem[Amaldi and Kann(1998)]{Amaldi1998237}
E.~Amaldi and V.~Kann.
\newblock On the approximability of minimizing nonzero variables or unsatisfied
  relations in linear systems.
\newblock \emph{Theoretical Computer Science}, 209\penalty0 (12):\penalty0
  237--260, 1998.

\bibitem[Asif(2008)]{asif2008primal}
M.~Asif.
\newblock Primal dual pursuit: A homotopy based algorithm for the {D}antzig
  selector.
\newblock Master's thesis, Dept. of Electrical and Computer Engineering,
  Georgia Institute of Technology, 2008.

\bibitem[Bach et~al.(2011)Bach, Jenatton, Mairal, and
  Obozinski]{bach2011convex}
F.~Bach, R.~Jenatton, J.~Mairal, and G.~Obozinski.
\newblock Convex optimization with sparsity-inducing norms.
\newblock \emph{Optimization for Machine Learning}, pages 19--53, 2011.

\bibitem[Bengio(2012)]{bengio2012deep}
Y.~Bengio.
\newblock Deep learning of representations for unsupervised and transfer
  learning.
\newblock \emph{Unsupervised and Transfer Learning Challenges in Machine
  Learning}, 7:\penalty0 19, 2012.

\bibitem[Boyd and Vandenberghe(2004)]{Boyd2004}
S.~Boyd and L.~Vandenberghe.
\newblock \emph{Convex Optimization}.
\newblock Cambridge University Press, 2004.

\bibitem[Calonder et~al.(2010)Calonder, Lepetit, Strecha, and
  Fua]{calonder2010brief}
M.~Calonder, V.~Lepetit, C.~Strecha, and P.~Fua.
\newblock {BRIEF: Binary Robust Independent Elementary Features}.
\newblock In \emph{European Conference on Computer Vision (ECCV)}, pages
  778--792. Springer, Crete, Greece, 2010.

\bibitem[Cannell and Stilwell(2005)]{Cannell2005OCEANS}
C.~J. Cannell and D.~J. Stilwell.
\newblock A comparison of two approaches for adaptive sampling of environmental
  processes using autonomous underwater vehicles.
\newblock In \emph{{MTS/IEEE OCEANS}}, pages 1514--1521, Washington, DC, Dec.
  19--23, 2005.

\bibitem[Capezio et~al.(2009)Capezio, Mastrogiovanni, Sgorbissa, and
  Zaccaria]{Capezio2009JCIT}
F.~Capezio, F.~Mastrogiovanni, A.~Sgorbissa, and R.~Zaccaria.
\newblock Robot-assisted surveillance in large environments.
\newblock \emph{Journal of Computing and Information Technology}, 17\penalty0
  (1):\penalty0 95--108, 2009.

\bibitem[Casafranca et~al.(2013{\natexlab{a}})Casafranca, Paz, and
  Pinies]{Casafranca13icraWS}
J.~J. Casafranca, L.~M. Paz, and P.~Pinies.
\newblock $\ell_1$ {Factor Graph SLAM: going beyond the} $\ell_2$ {norm}.
\newblock In \emph{Robust and Multimodal Inference in Factor Graphs
  Workshop,IEEE International Conference on Robots and Automation, (ICRA)},
  Karlsruhe, Germany, 2013{\natexlab{a}}.

\bibitem[Casafranca et~al.(2013{\natexlab{b}})Casafranca, Paz, and
  Pinies]{Casafranca13iros}
J.~J. Casafranca, L.~M. Paz, and P.~Pinies.
\newblock A back-end $\ell_1$ norm based solution for factor graph {SLAM}.
\newblock In \emph{IEEE/RSJ International Conference on Intelligent Robots and
  Systems (IROS)}, pages 17--23, Tokyo, Japan, Nov. 3--8, 2013{\natexlab{b}}.

\bibitem[Cheng et~al.(2010)Cheng, Yang, Yan, Fu, and Huang]{cheng2010learning}
B.~Cheng, J.~Yang, S.~Yan, Y.~Fu, and T.~Huang.
\newblock Learning with $\ell_1$-graph for image analysis.
\newblock \emph{IEEE Transactions on Image Processing}, 19\penalty0
  (4):\penalty0 858--866, 2010.

\bibitem[Churchill and Newman(2013)]{churchill2013experience}
W.~Churchill and P.~Newman.
\newblock {Experience-based navigation for long-term localisation}.
\newblock \emph{The International Journal of Robotics Research}, 32\penalty0
  (14):\penalty0 1645--1661, 2013.

\bibitem[Cummins and Newman(2008)]{CumminsNewmanIJRR08}
M.~Cummins and P.~Newman.
\newblock {FAB-MAP: Probabilistic Localization and Mapping in the Space of
  Appearance}.
\newblock \emph{The International Journal of Robotics Research}, 27\penalty0
  (6):\penalty0 647--665, 2008.

\bibitem[Dalal and Triggs(2005)]{dalal2005histograms}
N.~Dalal and B.~Triggs.
\newblock Histograms of oriented gradients for human detection.
\newblock In \emph{IEEE Computer Society Conference on Computer Vision and
  Pattern Recognition (CVPR)}, volume~1, pages 886--893, San Diego, CA, June
  20-26, 2005.

\bibitem[Donoho(2006{\natexlab{a}})]{Donoho2006TSP}
D.~L. Donoho.
\newblock Compressed sensing.
\newblock \emph{IEEE Transactions on Signal Processing}, 52\penalty0
  (4):\penalty0 1289--1306, 2006{\natexlab{a}}.

\bibitem[Donoho(2006{\natexlab{b}})]{donoho2006most}
D.~L. Donoho.
\newblock For most large underdetermined systems of linear equations the
  minimal 𝓁1-norm solution is also the sparsest solution.
\newblock \emph{Communications on pure and applied mathematics}, 59\penalty0
  (6):\penalty0 797--829, 2006{\natexlab{b}}.

\bibitem[Donoho and Tsaig(2006)]{donoho2006fast}
D.~L. Donoho and Y.~Tsaig.
\newblock Fast solution of $\ell_1$-minimization problems when the solution may
  be sparse.
\newblock Technical report, Dept. of Statistics, Stanford University, 2006.

\bibitem[Elad and Aharon(2006)]{elad2006image}
M.~Elad and M.~Aharon.
\newblock Image denoising via sparse and redundant representations over learned
  dictionaries.
\newblock \emph{IEEE Transactions on Image Processing}, 15\penalty0
  (12):\penalty0 3736--3745, 2006.

\bibitem[Elad et~al.(2010)Elad, Figueiredo, and Ma]{elad2010role}
M.~Elad, M.~Figueiredo, and Y.~Ma.
\newblock On the role of sparse and redundant representations in image
  processing.
\newblock \emph{Proceedings of the IEEE}, 98\penalty0 (6):\penalty0 972--982,
  2010.

\bibitem[Everingham et~al.(2012)Everingham, Van~Gool, Williams, Winn, and
  Zisserman]{pascal-voc-2012}
M.~Everingham, L.~Van~Gool, C.~K.~I. Williams, J.~Winn, and A.~Zisserman.
\newblock The {PASCAL} {V}isual {O}bject {C}lasses {C}hallenge 2012 {(VOC2012)}
  {R}esults.
\newblock
  http://www.pascal-network.org/challenges/VOC/voc2012/workshop/index.html,
  2012.

\bibitem[Galvez-Lopez and Tardos(2012)]{GalvezTRO12}
D.~Galvez-Lopez and J.~D. Tardos.
\newblock Bags of binary words for fast place recognition in image sequences.
\newblock \emph{IEEE Transactions on Robotics}, 28\penalty0 (5):\penalty0
  1188--1197, 2012.

\bibitem[Geiger et~al.(2012)Geiger, Lenz, and Urtasun]{KITTI}
A.~Geiger, P.~Lenz, and R.~Urtasun.
\newblock Are we ready for autonomous driving? the {KITTI} vision benchmark
  suite.
\newblock In \emph{Computer Vision and Pattern Recognition (CVPR), 2012 IEEE
  Conference on}, pages 3354--3361. IEEE, 2012.

\bibitem[Kumar B~G et~al.(2016)Kumar B~G, Carneiro, and Reid]{VijayNet}
V.~Kumar B~G, G.~Carneiro, and I.~Reid.
\newblock Learning local image descriptors with deep siamese and triplet
  convolutional networks by minimising global loss functions.
\newblock In \emph{The IEEE Conference on Computer Vision and Pattern
  Recognition (CVPR)}, June 2016.

\bibitem[Latif et~al.(2013)Latif, Cadena, and Neira]{LatifIJRR}
Y.~Latif, C.~Cadena, and J.~Neira.
\newblock Robust loop closing over time for pose graph {SLAM}.
\newblock \emph{The International Journal of Robotics Research}, 32\penalty0
  (14):\penalty0 1611--1626, 2013.

\bibitem[Latif et~al.(2014{\natexlab{a}})Latif, Cadena, and
  Neira]{latif2014robust}
Y.~Latif, C.~Cadena, and J.~Neira.
\newblock Robust graph slam back-ends: A comparative analysis.
\newblock In \emph{Intelligent Robots and Systems (IROS 2014), 2014 IEEE/RSJ
  International Conference on}, pages 2683--2690. IEEE, 2014{\natexlab{a}}.

\bibitem[Latif et~al.(2014{\natexlab{b}})Latif, Huang, Leonard, and
  Neira]{LatifRSS14}
Y.~Latif, G.~Huang, J.~Leonard, and J.~Neira.
\newblock An online sparsity-cognizant loop-closure algorithm for visual
  navigation.
\newblock In \emph{Proceedings of Robotics: Science and Systems}, Berkeley,
  USA, July 2014{\natexlab{b}}.

\bibitem[LeCun and Bengio(1995)]{lecun1995convolutional}
Y.~LeCun and Y.~Bengio.
\newblock Convolutional networks for images, speech, and time series.
\newblock \emph{The handbook of brain theory and neural networks},
  3361\penalty0 (10), 1995.

\bibitem[Lee et~al.(2013)Lee, Zhang, Lim, and Suh]{lee2013place}
J.~H. Lee, G.~Zhang, J.~Lim, and I.~H. Suh.
\newblock Place recognition using straight lines for vision-based {SLAM}.
\newblock In \emph{Robotics and Automation (ICRA), 2013 IEEE International
  Conference on}, pages 3799--3806. IEEE, 2013.

\bibitem[Lee et~al.(2014)Lee, Lee, Zhang, Lim, Chung, and Suh]{lee2014place}
J.~H. Lee, S.~Lee, G.~Zhang, J.~Lim, W.~K. Chung, and I.~H. Suh.
\newblock Outdoor place recognition in urban environments using straight lines.
\newblock In \emph{2014 IEEE International Conference on Robotics and
  Automation (ICRA)}, pages 5550--5557, May 2014.

\bibitem[Lowry et~al.(2016)Lowry, S{\"u}nderhauf, Newman, Leonard, Cox, Corke,
  and Milford]{lowry2016visual}
S.~Lowry, N.~S{\"u}nderhauf, P.~Newman, J.~J. Leonard, D.~Cox, P.~Corke, and
  M.~J. Milford.
\newblock Visual place recognition: A survey.
\newblock \emph{IEEE Transactions on Robotics}, 32\penalty0 (1):\penalty0
  1--19, 2016.

\bibitem[Malioutov et~al.(2005)Malioutov, Cetin, and
  Willsky]{malioutov2005homotopy}
D.~M. Malioutov, M.~Cetin, and A.~S. Willsky.
\newblock Homotopy continuation for sparse signal representation.
\newblock In \emph{IEEE International Conference on Acoustics, Speech, and
  Signal Processing}, 2005.

\bibitem[Milford(2013)]{milford2013vision}
M.~Milford.
\newblock Vision-based place recognition: how low can you go?
\newblock \emph{The International Journal of Robotics Research}, 32\penalty0
  (7):\penalty0 766--789, 2013.

\bibitem[Milford and Wyeth(2012)]{milford-daynight}
M.~Milford and G.~Wyeth.
\newblock {SeqSLAM}: Visual route-based navigation for sunny summer days and
  stormy winter nights.
\newblock In \emph{IEEE International Conference on Robotics and Automation
  (ICRA)}, pages 1643--1649, St. Paul, MN, May 14--18, 2012.

\bibitem[Nister and Stewenius(2006)]{hbow}
D.~Nister and H.~Stewenius.
\newblock Scalable recognition with a vocabulary tree.
\newblock In \emph{Computer Vision and Pattern Recognition, 2006 IEEE Computer
  Society Conference on}, volume~2, pages 2161--2168, 2006.
\newblock \doi{10.1109/CVPR.2006.264}.

\bibitem[Oliva and Torralba(2001)]{oliva2001modeling}
A.~Oliva and A.~Torralba.
\newblock {Modeling the shape of the scene: A holistic representation of the
  spatial envelope}.
\newblock \emph{International Journal of Computer Vision}, 42\penalty0
  (3):\penalty0 145--175, 2001.

\bibitem[Paul and Newman(2013)]{paul2013self}
R.~Paul and P.~Newman.
\newblock {Self-help: Seeking out perplexing images for ever improving
  topological mapping}.
\newblock \emph{The International Journal of Robotics Research}, 32\penalty0
  (14):\penalty0 1742--1766, 2013.

\bibitem[RAWSEEDS(2009)]{rawseeds}
RAWSEEDS.
\newblock {Robotics advancement through Webpublishing of sensorial and
  elaborated extensive data sets (project FP6-IST-045144)}, 2009.

\bibitem[Rosten and Drummond(2005)]{rosten2005fusing}
E.~Rosten and T.~Drummond.
\newblock Fusing points and lines for high performance tracking.
\newblock In \emph{IEEE International Conference on Computer Vision (ICCV)},
  volume~2, pages 1508--1515, Beijing, China, Oct. 17-20, 2005.

\bibitem[Shakeri and Zhang(2015)]{ShakeriFSR14}
M.~Shakeri and H.~Zhang.
\newblock Online loop-closure detection via dynamic sparse representation.
\newblock \emph{Field and Service Robotics (FSR)}, 2015.

\bibitem[Sivic and Zisserman(2003)]{Sivic03}
J.~Sivic and A.~Zisserman.
\newblock {Video Google}: {A} text retrieval approach to object matching in
  videos.
\newblock In \emph{Proceedings of the International Conference on Computer
  Vision}, volume~2, pages 1470--1477, Oct. 2003.

\bibitem[Smith et~al.(2009)Smith, Baldwin, Churchill, Paul, and
  Newman]{smith2009new}
M.~Smith, I.~Baldwin, W.~Churchill, R.~Paul, and P.~Newman.
\newblock The new college vision and laser data set.
\newblock \emph{The International Journal of Robotics Research}, 28\penalty0
  (5):\penalty0 595--599, 2009.

\bibitem[Sugiyama et~al.(2005)Sugiyama, Tsujioka, and Murata]{Sugiyama2005ICCC}
H.~Sugiyama, T.~Tsujioka, and M.~Murata.
\newblock Collaborative movement of rescue robots for reliable and effective
  networking in disaster area.
\newblock In \emph{International Conference on Collaborative Computing:
  Networking, Applications and Worksharing}, San Jose, CA, Dec. 19--21, 2005.

\bibitem[S\"underhauf(2015)]{vpriceChallenge}
N.~S\"underhauf.
\newblock {Visual Place Recogniton in Challenge Enviornments (VPRiCE)
  Challenge}.
\newblock
  https://roboticvision.atlassian.net/wiki/pages/viewpage.action?pageId=14188617,
  2015.

\bibitem[S{\"u}nderhauf et~al.(2015)S{\"u}nderhauf, Dayoub, Shirazi, Upcroft,
  and Milford]{sunderhaufIROS15}
N.~S{\"u}nderhauf, F.~Dayoub, S.~Shirazi, B.~Upcroft, and M.~Milford.
\newblock On the performance of convnet features for place recognition.
\newblock In \emph{Proc. {IEEE/RJS} Int. Conference on Intelligent Robots and
  Systems}, 2015.

\bibitem[Sunderhauf et~al.(2015)Sunderhauf, Shirazi, Jacobson, Dayoub,
  Pepperell, Upcroft, and Milford]{sunderhaufRSS15}
N.~Sunderhauf, S.~Shirazi, A.~Jacobson, F.~Dayoub, E.~Pepperell, B.~Upcroft,
  and M.~Milford.
\newblock Place recognition with convnet landmarks: Viewpoint-robust,
  condition-robust, training-free.
\newblock \emph{Proceedings of Robotics: Science and Systems XII}, 2015.

\bibitem[Zhang et~al.(2016)Zhang, Han, and Wang]{Zhang-RSS-16}
H.~Zhang, F.~Han, and H.~Wang.
\newblock Robust multimodal sequence-based loop closure detection via
  structured sparsity.
\newblock In \emph{Proceedings of Robotics: Science and Systems}, AnnArbor,
  Michigan, June 2016.
\newblock \doi{10.15607/RSS.2016.XII.043}.

\end{thebibliography}
\end{document}